\documentclass[10pt,journal,compsoc]{IEEEtran}
\ifCLASSOPTIONcompsoc
  \usepackage[nocompress]{cite}
\else
  \usepackage{cite}
\fi
\ifCLASSINFOpdf
  \usepackage[pdftex]{graphicx}
\else
\fi
\usepackage{amsmath}
\usepackage{amsthm}
\usepackage{amssymb}
\usepackage{amsfonts}
\usepackage{mathtools}
\usepackage{color}
\usepackage[ruled,vlined,linesnumbered,lined,boxed,commentsnumbered]{algorithm2e}
\usepackage{subcaption}
\usepackage{multirow}
\usepackage{booktabs}
\usepackage{ragged2e}
\usepackage{enumitem}
\usepackage{bigints}
\usepackage{mathrsfs}
\usepackage[mathscr]{euscript}

\usepackage{multirow}
\usepackage{multicol}

\usepackage{balance}

\newtheorem{corollary}{Corollary}

\newtheorem{definition}{Definition}
\newtheorem{assumption}{Assumption}
\newtheorem{proposition}{Proposition}
\newtheorem{property}{Property}

\newcommand\numberthis{\addtocounter{equation}{1}\tag{\theequation}}

\usepackage{multirow}
\usepackage{makecell}
\usepackage{booktabs}
\usepackage{cellspace}
\usepackage{array}

\usepackage{thmtools}
\usepackage{thm-restate}

\usepackage{hyperref}

\usepackage[framemethod=tikz]{mdframed}

\usepackage{siunitx}

\begin{document}
%
\title{Poisoning Attack against Estimating from Pairwise Comparisons}
%
%
%
%
\author
{
    Ke~Ma,~\IEEEmembership{Member,~IEEE,}
    Qianqian~Xu,~\IEEEmembership{Senior Member,~IEEE,}
    Jinshan~Zeng,\\
    Xiaochun~Cao,~\IEEEmembership{Senior Member,~IEEE,}
    and~Qingming~Huang,~\IEEEmembership{Fellow,~IEEE}
    \IEEEcompsocitemizethanks
    {
        \IEEEcompsocthanksitem K. Ma is with the School of Computer Science and Technology, University of Chinese Academy of Sciences, Beijing 100049, China, and with the Artificial Intelligence Research Center, Peng Cheng Laboratory, Shenzhen 518055, China. E-mail: make@ucas.ac.cn\protect\\
        \IEEEcompsocthanksitem Q. Xu is with the Key Laboratory of Intelligent Information Processing, Institute of Computing Technology, Chinese Academy of Sciences, Beijing 100190, China. E-mail: qianqian.xu@vipl.ict.ac.cn, xuqianqian@ict.ac.cn.\protect\\
        \IEEEcompsocthanksitem J. Zeng is with the School of Computer and Information Engineering, Jiangxi Normal University, Nanchang, Jiangxi, 330022, China. E-mail: jinshanzeng@jxnu.edu.cn, jsh.zeng@gmail.com\protect\\   
        \IEEEcompsocthanksitem X. Cao is with the State Key Laboratory of Information Security (SKLOIS), Institute of Information Engineering, Chinese Academy of Sciences, Beijing, 100093, China, and with the School of Cyber Security, University of Chinese Academy of Sciences, Beijing 100049, China. E-mail: caoxiaochun@iie.ac.cn\protect\\
        \IEEEcompsocthanksitem Q. Huang is with the Key Laboratory of Intelligent Information Processing, Institute of Computing Technology, Chinese Academy of Sciences, Beijing 100190, China, and with the School of Computer Science and Technology, University of Chinese Academy of Sciences, Beijing 100049, China, and with the Key Laboratory of Big Data Mining and Knowledge Management, the School of Economics and Management, University of Chinese Academy of Sciences, Beijing 100049, China, and with the Artificial Intelligence Research Center, Peng Cheng Laboratory, Shenzhen 518055, China. E-mail: qmhuang@ucas.ac.cn.\protect\\
    }
    \thanks{}
}

\IEEEtitleabstractindextext
{%
    \begin{abstract}
        \justifying
        As pairwise ranking becomes broadly employed for elections, sports competitions, recommendation, information retrieval and so on, attackers have strong motivation and incentives to manipulate or disrupt the ranking list. They could inject malicious comparisons into the training data to fool the target ranking algorithm. Such a technique is called ``{\textit{poisoning attack}}'' in regression and classification tasks. In this paper, to the best of our knowledge, we initiate the first systematic investigation of data poisoning attack on the pairwise ranking algorithms, which can be generally formalized as the dynamic and static games between the ranker and the attacker, and can be modeled as certain kinds of integer programming problems mathematically. To break the computational hurdle of the underlying integer programming problems, we reformulate them into the distributionally robust optimization (DRO) problems, which are computational tractable. Based on such DRO formulations, we propose two efficient poisoning attack algorithms and establish the associated theoretical guarantees including the existence of Nash equilibrium and the generalization ability bounds. The effectiveness of the suggested poisoning attack strategies is demonstrated by a series of toy simulations and several real data experiments. These experimental results show that the proposed methods can significantly reduce the performance of the ranker in the sense that the correlation between the true ranking list and the aggregated results with toxic data can be decreased dramatically.
    \end{abstract}  
    \begin{IEEEkeywords}
    Adversarial Learning, Poisoning Attack, Pairwise Comparison, Rank Aggregation, Robust Game, Distributionally Robust Optimization.
    \end{IEEEkeywords}
}

\maketitle

\IEEEdisplaynontitleabstractindextext

%
\IEEEpeerreviewmaketitle

\IEEEraisesectionheading{\section{Introduction}\label{sec:introduction}}

\IEEEPARstart{R}{ank} aggregation, in particular estimating a ranking based on comparisons between pairs of objects, arises in a variety of disciplines, including the social choice theory\cite{arrow2012social}, psychology\cite{CRITCHLOW1991294}, statistics\cite{Jiang2011}, machine learning\cite{pmlr-v54-korba17a}, bioinformatics\cite{kolde2012robust} and others. The convenience of these rank aggregation methods relies on their utilization of the ordinal data. Without features, the comparisons only contain the partial ranking lists generated by human beings. For instance, the voters who participated in an election choose one over the other candidates, which generate pairwise comparisons between the candidates. As another example workers in a crowdsourcing platform are often asked to identify the better advertisement of two possible visualization modes. Competitive sports such as tennis or chess also involve a serious of competitions between two players. From a modeling perspective, the rank aggregation approach treats pairwise comparisons as an access to estimate the underlying ``scores'' or ``qualities'' of the items being compared (\textit{e.g.}, preference of candidates, skill levels of tennis players, and advertisement performance). A vast body of prior work has made the significant progress in studying both statistical and computational aspects\cite{DBLP:conf/icml/RajkumarGL015,DBLP:conf/icml/ShahBGW16,DBLP:journals/jmlr/ShahBBPRW16,DBLP:journals/jmlr/NegahbanOTX18,PanMaoMutWaiCou19,DBLP:conf/icml/WauthierJJ13,DBLP:journals/jmlr/ShahW17,DBLP:journals/ior/NegahbanOS17}. 



However, the existing work ignores the security issue. Beyond statistical property and computational complexity, situations become complicated when the pairwise ranking algorithms are utilized in \textbf{\textit{high-stakes}} applications, \textit{e.g.} elections, sports competitions, and recommendation. In pursuit of huge economic benefits, the potential attackers have strong motivations and incentives to manipulate or disrupt the aggregated results. When the victims are ranking algorithms, a profit-oriented adversary could try his/her best to manipulate or disrupt the ranking list which will favor his/her demands-say, the attacker could place the special object at the top of the recommendation list, help the particular candidate to win an election or just defeat the candidate who should have won the election. If the attackers compromise the integrity of ranking results, the fairness and rationality will be lost in these high-stakes applications. Unfortunately, the security risk and serious threat of pairwise ranking problem have not been comprehensively examined yet. Can rank aggregation algorithms with pairwise comparisons be easily manipulated or disrupted? How reliable are their results in the high-stakes applications? 

To the best of our knowledge, the adversarial arsenal for pairwise ranking methods has never been serious studied. On one hand, the pairwise comparisons are the most simple data in the literature as just binary variables can represent them. Due to the absence of features, modifying these binary data is an easy job. On the other hand, any single comparison does not dictate the aggregated result. Even manipulating a small quantity of binary data could not affect the final global ranking. Such a contradiction inspires us to initiate an adversarial investigation of pairwise ranking problem. 

To execute the attack strategy in the scenario, the adversary must analyze the characteristics of pairwise ranking problems. Unlike the supervised learning tasks (\textit{e.g.} regression, classification, multi-arm bandit and reinforcement learning), the rank aggregation does not need the \textbf{\textit{test protocol}}. This means that the \textbf{\textit{evasion attacks}} (\textit{a.k.a} adversarial examples\cite{DBLP:journals/corr/GoodfellowSS14}) are not realistic. Evasion attack causes the fixed model to misbehave by well-crafted test data. But there is no test phrase to implement such a kind of attack. To archive his/her goal, the adversary needs to inject the manipulated data into the training data. Thus, rank aggregation in an adversarial setting is inherently related to the challenging \textbf{\textit{poisoning attacks}}\cite{DBLP:conf/icml/BiggioNL12,8418594}. Next, the adversary should consider the discrete property of the pairwise comparisons. Unlike the data consisting of features in continuous space, the input of pairwise ranking only consists of binary data. The adversary could only add, delete or flip pairwise comparisons to execute the poisoning attacks. Such limitations make the substantial attack operations on pairwise ranking even harder. How to design efficient algorithms that are able to inject toxic data in a discrete domain? It is the distinguishable characteristic of our work which is different with the existing poisoning attack approaches\cite{DBLP:conf/nips/MaZSZ19,DBLP:conf/nips/LiuSZ0H19,DBLP:conf/icml/LiuS19a,DBLP:conf/icml/MahloujifarMM19,DBLP:conf/ijcai/Ma0H19,DBLP:conf/icc/JiangLLRH19,8418594,DBLP:conf/icml/ZhangMS020,DBLP:conf/aaai/Chen020}. 

Given these challenges, we propose a principle framework for adversarial perturbations of pairwise comparisons that aims to break the integrity of rank aggregation result. In particular, we focus on the parametric model solved by maximum likelihood estimation\cite{Jiang2011}. We make the following contributions: 
\begin{itemize}
    \item We propose two game-theoretic frameworks specifically designed for adversary with the full or limited knowledge of the victim algorithm. By introducing the uncertainty set around the original data, the adversary aims to find a toxic distribution which will maximize the risk of estimating the ranking parameters. The dynamic threat model assumes that the adversary is aware of the original pairwise comparisons, the ranking algorithm and the ranking parameter learned from the original data. This model relates to a dynamic distributionally robust game. Besides, we propose a weaker threat model which assumes that the adversary only predominates the original data and the ranking algorithm. It induces a static distributionally robust game where the adversary can only execute the attacks in the ``black-box'' attack style.
    \item Different statistical attacks corresponding to the dynamic and static threat models are formulated into the bi-level optimization problem and distributionally robust optimization problem. In the bi-level optimization problem, we adopt $\chi^2$ divergence to describe the uncertainty set around the original data. The optimal attack strategy can be obtained by the projection onto a simplex. In the distributionally robust optimization problem, the uncertainty set is a Wasserstein ball. Based on the strong duality, the optimal attack behavior is obtained by a least square problem with a special regularization.
    \item We prove the existence of robust optimization equilibrium and establish a minimax framework for pairwise ranking under adversarial setting.
\end{itemize}

To the best of our knowledge, this is the first systematic study of attacking rank aggregation under different adversarial models. The extensively evaluations are conducted on several datasets from different high-stake domains, including election, crowdsourcing, and recommendation. Our experiments demonstrate that the proposed poisoning attack could significantly decrease the correlation between the true ranking list and the aggregated result. 

\noindent\textbf{Notations}

Let $\boldsymbol{V}$ be a finite set. We will adopt the following notation from combinatorics:
\begin{equation*}
    \binom{\boldsymbol{V}}{k}:=\text{set of all}\ k\ \text{element subset of}\ \boldsymbol{V}.
\end{equation*}
In particular $\binom{\boldsymbol{V}}{2}$ would be the set of all unordered pairs of elements of $\boldsymbol{V}$. The sets of ordered pair will be denoted $\boldsymbol{V}\times\boldsymbol{V}$. Ordered and unordered pairs will be delimited by parentheses $(i,\ j)$ and braces $[i,\ j]$ respectively. We will use positive integers to indicate alternatives and voters. Henceforth, $\boldsymbol{V}$ will always be the set $[n]=\{1,\dots, n\}$ and will denote a set of alternatives to be ranked. $\mathcal{U}=\{1,\dots, m\}$ will denote a set of voters. For $i,\ j\in\boldsymbol{V}$, we write $i\succ j$ to mean that alternative $i$ is preferred over alternative $j$. If we wish to emphasize the preference judgment of a particular voter $u\in\mathcal{U}$, we will write $i\succ_u j$. Suppose that $\boldsymbol{\Omega}\subset\mathbb{R}^n$ is the data space, we denote $(\boldsymbol{\Omega}, d(\cdot,\ \cdot))$ as a metric space equipped with some metric $d:\boldsymbol{\Omega}\times\boldsymbol{\Omega}\rightarrow\mathbb{R}$.

\section{Ranking with Pairwise Comparisons}

Given a collection $\boldsymbol{V}$ of $n$ alternatives, we suppose that each $i\in\boldsymbol{V}$ has a certain numeric quality score $\theta^*_i$. We represent the quality scores of $\boldsymbol{V}$ as a vector $\boldsymbol{\theta}^*\in\mathbb{R}^n$. Suppose that a comparison of any pair $[i,\ j]\in\binom{\boldsymbol{V}}{k}$ is generated via the comparison of the corresponding scores $\theta^*_i,\ \theta^*_j$ in the presence of noise. Let $y^*_{ij}$ be the true direction of a pair $[i,\ j]$ as 
\begin{equation}
    y^*_{ij} =
    \begin{cases}
        \ \ \ 1, & \theta^*_i>\theta^*_j,\\
             -1, & \theta^*_i<\theta^*_j.
    \end{cases}
\end{equation}
Let $\mathcal{C}$ be a collection of $N$ pairwise comparisons
\begin{equation}
    \mathcal{C} = \{\boldsymbol{c}=[i,\ j]\ |\ y_{ij}=1,\ i,\ j\in\boldsymbol{V},\ i\neq j\},
\end{equation}
and $y_{ij}$ is the label of pair $[i,\ j]$ which could not be consist with $y^*_{ij}$. It is worth noting that $\mathcal{C}$ is always a multi-set. For any pair $[i,\ j]$, it could be labeled by multiple users. Given a set of voter $\mathcal{U}=\{u_1,\dots,u_m\}$, let $y^u_{ij}$ be the judgment of pair $[i,\ j]$ given by voter $u\in\mathcal{U}$. We can aggregate $y^{u_1}_{ij},\dots,y^{u_m}_{ij}$ into a weight $w^0_{ij}$. Define $w(i,\ j,\ u)$ as the indicator of $y^{u}_{ij}$:
\begin{equation}
    w(i,\ j,\ u)=
    \begin{cases}
    \ 1, & \text{if}\ y^{u}_{ij}=1,\ u\in\mathcal{U}\\ 
    \ 0, & \text{otherwise}
    \end{cases}
\end{equation}
and the weight $w^0_{ij}$ of $y_{ij}$ is 
\begin{equation}
    \label{eq:edge_weight}
    w^0_{ij} = \underset{u\ \in\ \mathcal{U}}{\sum}\ w(i,\ j,\ u).
\end{equation} 
Moreover, we introduce the comparison matrix $\boldsymbol{A}$. If there exists a comparison $\boldsymbol{c}\in\mathcal{C}$, it can be described by its label $y_{ij}$ and a row of $\boldsymbol{A}$ as $\boldsymbol{a}^{\boldsymbol{c}}=\{a^{\boldsymbol{c}}_1,\dots,a^{\boldsymbol{c}}_{|\mathcal{C}|}\}$:
\begin{equation}
    a^{\boldsymbol{c}}_k =
    \begin{cases}
        \ \ \ 1, & k = i,\\
             -1, & k = j,\\
        \ \ \ 0, & \text{otherwise.}
    \end{cases}
\end{equation}
Then the data of pairwise ranking problem can be represented by $\mathcal{C}_{\mathcal{U}}=\{\boldsymbol{A},\ \boldsymbol{y},\ \boldsymbol{w}_0\}$ where $\boldsymbol{w}_0=\{w^0_{ij}\}$, $\boldsymbol{y}=\{y_{ij}\}$ is a $n(n-1)/2$-d \textbf{binary} vector.

In statistical ranking or estimation from pairwise comparison, our goal is to obtain a score vector $\boldsymbol{\hat{\theta}}$ to minimize a loss of a global ranking on the given data $\mathcal{C}_{\mathcal{U}}$. 
\begin{equation}
        \label{opt:pairwise_rank}
        \boldsymbol{\hat{\theta}}\in\underset{\boldsymbol{\theta}\ \in\ \mathbb{R}^n}{\textbf{\textit{arg min}}}\ \ \ell(\boldsymbol{\theta};\ \mathcal{C}_{\mathcal{U}}).
\end{equation}
In particular, let the estimation of $y_{ij}$ be
\begin{equation}
    \label{model:ordinal}
    \hat{y}_{ij} = \textbf{\textit{sgn}}\Big(\langle\boldsymbol{a}^{\boldsymbol{c}},\ \boldsymbol{\theta}\rangle+\varepsilon_{\boldsymbol{c}}\Big),\ \forall\ \boldsymbol{c}\in\mathcal{C},
\end{equation}
where $\textbf{\textit{sgn}}(\cdot)$ is the sign function, $\varepsilon_c$ is the independent and identically distributed (\textit{i.i.d}) noise variable and has a cumulative distribution function (\textit{c.d.f}) $F$. Actually, \eqref{opt:pairwise_rank} minimizes the derivation between the observed label $\boldsymbol{y}$ and its estimation $\hat{\boldsymbol{y}}=\{\hat{y}_{ij}\}$ based on the observing data $\mathcal{C}_{\mathcal{U}}$. In addition, the random variable $\varepsilon_c$ plays the role of a noise parameter, with a higher magnitude of $\varepsilon_c$ leading to more uncertainty in the comparisons and the higher probability of sign inconsistency occurred between $y_{ij}$ and $\theta_i-\theta_j$. The event that object $i$ dominating object $j$ ($y_{ij}=1$) is generally independent of the order of the two items being compared, thus, the following holds:
\begin{equation}
    \Pr(y_{ij}=1)= 1-\Pr(y_{ij}=-1)
\end{equation}
and $F$ is a symmetric \textit{c.d.f} whose continuous inverse is well-defined. Some typical examples of \eqref{model:ordinal} are the uniform model \cite{DBLP:journals/jmlr/ShahBBPRW16}, the Bradley-Terry- Luce (BTL) model \cite{bradley1952rank,luce1959individual}, and the Thurstone model with Gaussian noise (Case V) \cite{thurstone1927law}, which have been extensively studied in literature (\textit{e.g.}, \cite{david1963method,8319957}). In this paper, we focus on the \textbf{Uniform Model}: one can adopt the symmetric \textit{c.d.f} $F(t) = \frac{t+1}{2}$, and the general set-up \eqref{model:ordinal} turns to be a linear model. Furthermore, the loss function in \eqref{opt:pairwise_rank} can be specialized as the weighted sum-of-squares function:     
\begin{equation}
    \label{eq:weight_l2_loss}
    \begin{aligned}
        & \ell\ (\boldsymbol{\theta};\ \mathcal{C}_{\mathcal{U}})&=&\ \ \frac{1}{2|\mathcal{C}_{\mathcal{U}}|}\ \|\ \boldsymbol{y} - \boldsymbol{A}\boldsymbol{\theta}\ \|^2_{2,\ \boldsymbol{w}_0} \\
        & &=&\ \ \frac{1}{2|\mathcal{C}_{\mathcal{U}}|}\ \underset{(i,\ j)}{\sum}w^0_{ij}\ (\ y_{ij}-\theta_i+\theta_j\ )^2.
    \end{aligned} 
\end{equation}

\section{Methodology}

In this section, we systematically introduce the
methodology for poisoning attacks on pairwise ranking. Specifically, we first start by introducing two game-theoretic threat models including the full knowledge and the limited knowledge adversaries. Then we present the corresponding algorithms to generate the optimal strategies of these threat models at different uncertainty budgets. Finally, the existence of equilibrium and the results of generalization analysis are discussed in the end of this section.

\subsection{Poisoning Attack on Pairwise Ranking}

We provide here a detailed adversarial framework for poisoning attacks against pairwise ranking algorithms. The framework consists of defining the adversary’s goal, knowledge of the attacked method, and capability of manipulating the pairwise data, to eventually define the optimal poisoning attack strategies.

\noindent\textbf{The Goal of Adversary. }If an adversary executes the poisoning attack, he/she will provide the ranker with the toxic data. This action will mislead its opponent into picking parameters to generate a different ranking result from $\hat{\boldsymbol{\theta}}$ obtained by the original data $\mathcal{C}_{\mathcal{U}}$ in \eqref{opt:pairwise_rank}. Let $\bar{\boldsymbol{\theta}}$ be the solution of \eqref{opt:pairwise_rank} with the toxic data, it satisfies
\begin{equation}
    d(\pi_{\boldsymbol{\theta}^*},\ \pi_{\hat{\boldsymbol{\theta}}}) \leq d(\pi_{\boldsymbol{\theta}^*},\ \pi_{\bar{\boldsymbol{\theta}}}),
\end{equation}
where $\boldsymbol{\theta}^*$ is the true quality scores of the objects, $\pi_{\boldsymbol{\theta}}$ is the ranking list decided by $\boldsymbol{\theta}$ and $d(\pi_1,\ \pi_2)$ measures the similarity of two ordered lists $\pi_1$ and $\pi_2$. 

\noindent\textbf{The Knowledge of Adversary. }We assume two distinct attack scenarios which are distinguished by the knowledge of adversary, referred to as \textbf{\textit{dynamic}} and \textbf{\textit{static}} attacks in the following. The adversaries in the two scenarios have different knowledge of the victims.
\begin{itemize}
    \item In \textbf{\textit{dynamic}} attacks, the attacker is assumed to know the observed data $\mathcal{C}_{\mathcal{U}}$, the ranking algorithm, and even the ranking parameters $\hat{\boldsymbol{\theta}}$ obtained by the original data $\mathcal{C}_{\mathcal{U}}$ in \eqref{opt:pairwise_rank}. If a dictator wants to sabotage the election which will subvert his/her predominant, he/she would not need to manipulate the results of the election. Making the most competitive opponent lose the advantage in the key districts will achieve the purpose. The dictator could execute the dynamic strategies as the aggregation process is a ``white-box'' to him/her. This adversarial mechanism can be implemented by establishing the \textbf{\textit{hierarchical}} relationship between the ranker and the attacker. The attacker is assumed to anticipate the reactions of the ranker; this allows him/her to choose the best—or optimal—strategy accordingly. Such a \textbf{\textit{hierarchical}} interaction results in the fact that the mathematical program related to the ranking process is part of the adversary's constraints. It is also known as the \textbf{\textit{dynamic}} or Stackelberg (leader-follower) game\cite{Basar-Olsder1999} in the literature: the two agents take their actions in a sequential (or repeated) manner. Moreover, the \textbf{\textit{hierarchical}} relationship is the major feature of bi-level optimization. The bi-level program includes two mathematical programs within a single instance, one of these problems being part of the constraints of the other one.
    \item In \textbf{\textit{static}} attacks, the attacker could not grasp $\hat{\boldsymbol{\theta}}$ but is still aware of the observed data $\mathcal{C}_{\mathcal{U}}$ and the ranking algorithm. This scenario comes from the fact that the ranking aggregation problem does not need the \textbf{\textit{test protocol}}. Once the adversary provides the modified data, the victim would generate the ranking list immediately. There is no chance to monitor the ranker's behavior. In most cases, the adversary can not obtain $\hat{\boldsymbol{\theta}}$. There is no feedback for the adversary to update his/her strategies. A competitor of the e-commerce platform, who wants to disrupt the recommendation results and destroy the user experience, would execute the static strategies. Promoting the rank of specific goods is challenging. Disrupting the normal ranking result is sufficient to archive his/her purpose. The competitor could only execute the static strategies as the aggregation process is a ``gray-box''. The leading e-commerce platform is the only one who could access the ranking parameters. The objective function and the pairwise comparisons for recommendation can be perceivable to the adversary. This adversarial mechanism should be modeled as a \textbf{\textit{static}} game. A static game is one in which a single decision is made by each player, and each player has no knowledge of the decision made by the other players before making their own decision. In other words, decisions or actions are made simultaneously (or the order is irrelevant).
\end{itemize}

\noindent\textbf{The Capability of Adversary. }To modify the original data $\mathcal{C}_{\mathcal{U}}$ in poisoning attacks, the adversary will inject an arbitrary pair $[i,\ j]\in\binom{\boldsymbol{V}}{k}$ with any directions into $\mathcal{C}_{\mathcal{U}}$, delete the existing comparison $\boldsymbol{c}=(i,\ j)$ in $\mathcal{C}_{\mathcal{U}}$ or just flip the label of $\boldsymbol{c}$. The three kinds of operations require some new representations of the observed set. We augment the observed data $\mathcal{C}_{\mathcal{U}}$ with the comparisons which are not labeled by users in $\mathcal{U}$. Let $\mathcal{D} = \boldsymbol{V}\times\boldsymbol{V}$ be the set of all ordered pairs, and $|\mathcal{D}|= N = n(n-1)$. The weights of all possible comparisons are $\boldsymbol{w}'_0$ and there exist $0$ entries in $\boldsymbol{w}'_0$. 

As $\mathcal{D}$ is the complete comparison set, the comparison matrix $\boldsymbol{B}$ will be fixed and we can adopt a $n(n-1)$-d \textbf{single-value} vector to represent the labels, saying that $\boldsymbol{y}'$ is a vector with all entries are $1$. Now all attack operations (adding, deleting and flipping) can be executed by increasing or decreasing the corresponding weight $\boldsymbol{w}'_0$.
\begin{equation}
    \begin{aligned}
    \label{eq:fixed_design_y}
    & \boldsymbol{B} &=&\ \begin{bmatrix}
           \boldsymbol{b}_{1,2}\\ 
           \boldsymbol{b}_{1,3}\\ 
           \vdots\\ 
       \ \ \boldsymbol{b}_{n,n-2}\ \ \\  
       \ \ \boldsymbol{b}_{n,n-1}\ \ 
    \end{bmatrix}\subset\big\{-1,\ 0,\ 1\big\}^{N\times n},\\
    & \boldsymbol{y}' &=&\ \ \big\{\ y_{1,2},\ y_{2,1},\ \dots,\ y_{n-1,n},\ y_{n,n-1}\ \big\}^\top\\
    & &=&\ \ \big\{\ 1,\ 1,\ \dots,\ 1,\ 1\ \big\}^\top. 
    \end{aligned}
\end{equation}

Besides injecting the toxic data, the attacker also needs to disguise himself/herself. It means that the adversary needs to coordinate a poisoned $\boldsymbol{w}=\{w_{ij}\}$ associated with $\boldsymbol{w}'_0$. Intuitively, the adversary could not obtain $\boldsymbol{w}$ through the drastic changes, neither on each $w_{ij}$ nor $\sum_{(i,j)} w_{ij}$. Such limitations lead to the following constraints for the adversary's action. First, the total difference between $\boldsymbol{w}'_0$ and $\boldsymbol{w}$ would be smaller than $b$, namely,
\begin{equation}
    \label{eq:adversary_constraint_1}
    \|\ \boldsymbol{w}\ -\ \boldsymbol{w}'_0\ \|_1\ \leq\ b,\ \ b\ \in\ \mathbb{N}_+.
\end{equation}
Here the positive integer $b$ bounds the total number of malicious samples thereby limiting the capabilities of the attacker. Furthermore, the adversary could not alter the number of votes on each pairwise comparison $\boldsymbol{c}\in\mathcal{D}$ obviously. This constraint on the adversary leads to the following condition: 
\begin{equation}
    \label{eq:adversary_constraint_2}
    \|\ \boldsymbol{w}\ -\ \boldsymbol{w}'_0\ \|_{\infty}\leq\ l,\ l\in\mathbb{N}_+,\ l\leq\textbf{\textit{min}}\{\textbf{\textit{max}}(\boldsymbol{w}'_0),\ b\}.
\end{equation}
The positive integer $l$ leads the conservative perturbations on the observed samples. To summarize, the adversary‘s action set $\boldsymbol{\Omega}_1$ is
\begin{equation}
    \label{eq:adversary_action_space_1}
    \begin{aligned}
        \boldsymbol{\Omega}_1\ \ =\ \ \left\{\ \ \boldsymbol{w}\ \ \left|\  
    \begin{matrix}
        \boldsymbol{w}\ \in\ \mathbb{N}^{N},\ \ l,\ b\ \in\ \mathbb{N},\\
        \ \|\ \boldsymbol{w}\ -\ \boldsymbol{w}'_0\ \|_1\ \leq\ b,\\
        \ \|\ \boldsymbol{w}\ -\ \boldsymbol{w}'_0\ \|_{\infty}\leq\ l,\\
        \ \ l\ \leq\ \textbf{\textit{min}}\{\textbf{\textit{max}}(\boldsymbol{w}'_0),\ b\}\ \ 
    \end{matrix}\right.\right\}.
    \end{aligned}
\end{equation}

Furthermore, the attacker must pay for his/her malicious behaviors. Let $s:\mathbb{N}^N\times\mathbb{N}^N\rightarrow\mathbb{R}$ is a ``cost'' function measured the overhead of the perturbation as changing $\boldsymbol{w}_0$ into $\boldsymbol{w}$. The attacker hopes that the toxic weight $\boldsymbol{w}$ will represent the lowest cost option. Let $\boldsymbol{\Omega}_2$ be the budget set of the adversary
\begin{equation}
    \label{eq:adversary_action_space_2}
    \boldsymbol{\Omega}_2\ \ =\ \ \left\{\boldsymbol{w}\ \Big|\ \boldsymbol{w}\in\underset{}{\textbf{\textit{arg min}}}\ \ s(\boldsymbol{w},\ \boldsymbol{w}'_0)\right\}
\end{equation}
Finally, the action set is $\boldsymbol{\Omega}_0 = \boldsymbol{\Omega}_1\cap\boldsymbol{\Omega}_1$ which figures out the capability of the adversary. 

\vspace{0.3cm}
\noindent\textbf{Poisoning Attack Strategies.} Here we specify the different poisoning strategies for the two attack scenarios.
\begin{itemize}
    \item \textbf{Dynamic attack strategy.} Consider the goal and knowledge of attacker, we formulate the interaction between ranker and the adversary with full knowledge as a dynamic game. In this game, information is assumed to be complete (\textit{i.e.}, the players’ payoff functions, as well as the constraint set $\boldsymbol{\Omega}_0$ and the flexible set of ranking parameter $\boldsymbol{\Theta}$, are common knowledge) and perfect (\textit{i.e.}, the attacker knows the ranker's decision). Having received the ranker's decision $\hat{\boldsymbol{\theta}}$, the attacker chooses a feasible decision $\boldsymbol{w}\in\boldsymbol{\Omega}_0$ that maximizes the ranker's loss function to increase the risk of the ranker's estimation based on $\{\boldsymbol{w},\ \boldsymbol{B},\ \boldsymbol{y}'\}$. Such a dynamic game can be formulated into the following bi-level optimization problem:
    \begin{subequations}
        \label{opt:Stackelberg}
        \begin{align}
            & &\underset{\boldsymbol{w}\ \in\ \boldsymbol{\Omega}_0}{\textbf{\textit{max}}}&\ \ \ell(\boldsymbol{w};\ \boldsymbol{\hat{\theta}},\ \boldsymbol{B},\ \boldsymbol{y}'), \label{opt:upper_Stackelberg}\\
            & &\textbf{\textit{subject to}}&\ \ \boldsymbol{\hat{\theta}}\in\underset{\boldsymbol{\theta}\ \in\ \boldsymbol{\Theta}}{\textbf{\textit{arg min}}}\ \ \ell(\boldsymbol{\theta};\ \boldsymbol{w}'_0,\ \boldsymbol{B},\ \boldsymbol{y}').\ \label{opt:lower_Stackelberg}
        \end{align}  
    \end{subequations}
    The upper level optimization \eqref{opt:upper_Stackelberg} amounts to selecting the toxic data $\boldsymbol{w}$ to maximize the loss function of the ranker, while the lower level optimization \eqref{opt:lower_Stackelberg} corresponds to calculate the ranking parameter $\boldsymbol{\hat{\theta}}$ with original data $\{\boldsymbol{w}_0,\ \boldsymbol{B},\ \boldsymbol{y}'\}$. Once the adversary generates $\boldsymbol{w}$, he/she will deliver the toxic data to the ranker. Then the poisoned parameter $\bar{\boldsymbol{\theta}}$ will be obtained by
    \begin{equation}
        \label{eq:last_step}
        \boldsymbol{\bar{\theta}}=\underset{\boldsymbol{\theta}\ \in\ \boldsymbol{\Theta}}{\textbf{\textit{arg min}}}\ \ \ell(\boldsymbol{\theta};\ \boldsymbol{w},\ \boldsymbol{B},\ \boldsymbol{y}').
    \end{equation}
    \item \textbf{Static attack strategy.} This strategy is represented such a type of adversary whose ability is to inflict the highest possible risk of the ranker when no information about his/her interests is available. It means that the two players make decisions simultaneously, and the attacker does not knows the ranker’s decision. Such a static game can be formulated into the following min-max optimization problem:
    \begin{equation}
    \label{opt:bilevel}
    \begin{aligned}
        \underset{\boldsymbol{\theta}\ \in\ \boldsymbol{\Theta}}{\textbf{\textit{min}}}&\ \underset{\boldsymbol{w}\ \in\ \boldsymbol{\Omega}_0}{\textbf{\textit{max}}}\ \ell(\boldsymbol{\theta},\ \boldsymbol{w};\ \boldsymbol{B},\ \boldsymbol{y}'). 
    \end{aligned}
    \end{equation}
    The poisoned parameter $\bar{\boldsymbol{\theta}}$ will be solved by \eqref{eq:last_step}.
\end{itemize}

  However, solving the dynamic and static attack strategies from \eqref{opt:Stackelberg} and \eqref{opt:bilevel} are challenging. On one hand, the bi-level optimization \eqref{opt:Stackelberg} and the min-max problem \eqref{opt:bilevel}are both mixed-integer programming problem as the variable $\boldsymbol{w}$ is restricted to be positive integers. On the other hand, the feasible set $\boldsymbol{\Omega}_0$ corresponds to a non-linear constraint as it requires to find the perturbation in the neighborhood of $\boldsymbol{w}'_0$ with the lowest cost. It is well-known that linear integer programmings are NP-complete problems \cite{DBLP:conf/coco/Karp72}. Such a non-linear constraint makes these problems even more complex. Obviously, adopting the heuristic methods to solve the optimal attack strategies \eqref{opt:Stackelberg} and \eqref{opt:bilevel} is sub-optimal. In this part, we will develop the other model based on ideas from distributionally robust optimization \cite{DBLP:conf/nips/NamkoongD17,doi:10.1287/moor.2018.0936,DBLP:journals/corr/abs-1712-06050} that provides the tractable convex formulations for solving the optimal strategies in the dynamic and static scenarios.

\subsection{Distributional Perspective and Robust Game}
In the above formulations \eqref{opt:Stackelberg} and \eqref{opt:bilevel}, the attacker modifies the number of votes on each pairwise comparison with constraints $\boldsymbol{\Omega}_0$. This formulation leads to the mixed-integer programming problem. Here we introduce a distributional perspective to establish the tractable optimization problem. Generally speaking, the attacker and the ranker both access the original data $\mathcal{D}$ to play the dynamic or static game. The non-toxic pairwise comparisons $\mathcal{D}=\{\boldsymbol{w}'_0,\ \boldsymbol{B},\ \boldsymbol{y}'\}$ are actually drawn from an empirical distribution $\mathbb{P}_N$
\begin{equation*}
    \label{eq:empricial_distribution}
    \mathbb{P}_N\ = \ \frac{1}{N}\ \underset{\boldsymbol{c}\ \in\ \mathcal{C}}{\sum}\ \delta({w^0_{ij}}',\ \boldsymbol{b}_{i,j},\ {y_{ij}}'),\ \ \boldsymbol{c}=(i,\ j),
\end{equation*}
where $\delta({w^0_{ij}}',\ \boldsymbol{b}_{i,j},\ {y_{ij}}')$ is the Dirac probability measure on $({w^0_{ij}}',\ \boldsymbol{b}_{i,j},\ {y_{ij}}')$. With $\boldsymbol{B}$ and $\boldsymbol{y}'$ as \eqref{eq:fixed_design_y}, the marginal distribution of $\boldsymbol{w}'_0$ plays a vital role in the sequel. With some abuse of symbol, we treat the marginal distribution of $\boldsymbol{w}_0$ as the distribution of the original data and 
\begin{equation*}
    \label{eq:margin_distribution}
    \mathbb{P}_N=\frac{1}{N}\ \underset{\boldsymbol{c}\ \in\ \mathcal{C}}{\sum}\ \delta({w^0_{ij}}'),\ \ \boldsymbol{c}=(i,\ j).
\end{equation*}
The attacker chooses a perturbation function $\psi:\mathbb{N}^{N}\rightarrow\mathbb{N}^{N}$ that changes the weight $\boldsymbol{w}'_0$ to $\boldsymbol{w}\in\boldsymbol{\Omega}_0$. Such a perturbation $\psi$ induces a transition from the empirical distribution $\mathbb{P}_N$ to a poisoned distribution $\mathbb{Q}$. If the attacker selects $\mathbb{Q}$ in a sufficiently small \textit{neighborhood} of $\mathbb{P}_N$, namely, the ``distance'' between the poisoned distribution $\mathbb{Q}$ and the empirical distribution $P$ would be sufficiently small, the attacker could obtain a ``local'' solution and $\mathbb{Q}$ is a ``good'' approximation of $\mathbb{P}_N$ in the sense of such a ``distance''. Therefore, the poisoned sample $\boldsymbol{w}$ would satisfy the constraints \eqref{eq:adversary_constraint_1} and \eqref{eq:adversary_constraint_2}. Here we directly work with the empirical distribution $\mathbb{P}_N$ (or other nominal distribution) and consider $\mathbb{Q}$ is close to the nominal distribution in terms of a certain statistical distance. 

There exists some popular choices of the statistical distance, such as $\phi$-divergences \cite{doi:10.1287/mnsc.1120.1641,Jiang2016,doi:10.1287/educ.2015.0134,wang2016likelihood,DBLP:conf/nips/NamkoongD16,DBLP:conf/nips/NamkoongD17,DBLP:journals/jmlr/DuchiN19}, Prokhorov metric \cite{Erdogan2006}, Wasserstein distances \cite{doi:10.1287/opre.2014.1323,Blanchet_2019,gao2016distributionally,MohajerinEsfahani2018,DBLP:conf/nips/LeeR18} and maximum mean discrepancy \cite{staib2019distributionally}.

For dynamic attack strategy \eqref{opt:Stackelberg}, we adopt the $\phi$-divergence\cite{DBLP:journals/tit/LieseV06} as the discrepancy measure between the empirical distribution $\mathbb{P}_n$ and the toxic distribution $\mathbb{Q}$.
\begin{definition}[$\phi$-divergence and $\chi^2$-divergence]
    Let $\phi: \mathbb{R}_+\rightarrow\mathbb{R}$ be a convex function with $\phi(1) = 0$. Then the $\phi$-divergence between distributions $\mathbb{Q}$ and $\mathbb{P}$ defined on a measurable space $\mathcal{X}$ is
    \begin{equation*}
        \label{eq:phi_divergence}
        \begin{aligned}
            d_{\phi}(\mathbb{Q}\ ||\ \mathbb{P})=\int\phi\left(\frac{d\mathbb{Q}}{d\mathbb{P}}\right)\mathrm{d}\mathbb{P}=\int_{\mathcal{X}}\phi\left(\frac{q(x)}{p(x)}\right)p(x)\mathrm{d}\mu(x),
        \end{aligned}
    \end{equation*}
    where $\mu$ is a $\sigma$-finite measure on $\mathcal{X}$ satisfying $\mathbb{Q}, \mathbb{P}$ are absolutely continuous with respect to $\mu$, and $q=\frac{\mathrm{d}\mathbb{Q}}{\mathrm{d}\mu}$, $p=\frac{\mathrm{d}\mathbb{P}}{\mathrm{d}\mu}$ are the Radon–Nikodym derivative with respect to $\mu$. If $\phi$ is adopted as $\phi(t) = \frac{1}{2}(t-1)^2$, it is known as the $\chi^2$-divergence.
\end{definition}

Suppose that $\boldsymbol{\mathfrak{X}}(\mathbb{P}_N)$ is a set of probability distributions from the empirical distribution with $\chi^2$-divergence. This $\chi^2$ ball with radius $\alpha$ is given by
\begin{equation}
    \boldsymbol{\mathfrak{X}}^{\alpha}(\mathbb{P}_N) = \left\{\ \mathbb{Q}\in\mathcal{P}(\boldsymbol{\Omega}_1)\ \Big\vert\ d_{\chi^2}(\mathbb{Q}\ ||\ \mathbb{P}_N)\leq\alpha\ \right\},
\end{equation}
where $\mathcal{P}(\boldsymbol{\Omega}_1)$ denotes the set of all Borel probability measures on $\boldsymbol{\Omega}_1$. With carefully chosen $\alpha$, the adversary chooses $\boldsymbol{w}$ from the toxic distribution $\mathbb{Q}\in\boldsymbol{\mathfrak{X}}^{\alpha}(\mathbb{P}_N)$. $\boldsymbol{w}$ could satisfy the neighborhood constraints as \eqref{eq:adversary_constraint_1} and \eqref{eq:adversary_constraint_2}. Replacing the minimal `cost' constraint \eqref{eq:adversary_action_space_2} by the neighborhood constraint defined with the $\chi^2$ ball, we formulate the following bi-level optimization to obtain the dynamic attack strategy
\begin{equation}
    \label{opt:chi_2}
    \begin{aligned}
        & &\underset{\mathbb{Q}\ \in\ \boldsymbol{\mathfrak{X}}^{\alpha}(\mathbb{P}_N)}{\textbf{\textit{max}}}&\ \ \mathbb{E}_{\boldsymbol{w}\sim\mathbb{Q}}\big[\ell(\boldsymbol{w};\ \boldsymbol{\hat{\theta}})\big],\\
        & &\textbf{\textit{subject to}}&\ \ \boldsymbol{\hat{\theta}}=\underset{\boldsymbol{\theta}\ \in\ \boldsymbol{\Theta}}{\textbf{\textit{arg min}}}\ \ \ell(\boldsymbol{\theta};\ \boldsymbol{w}'_0).
    \end{aligned}  
\end{equation}
The $\chi^2$-divergence and the ``local'' neighborhood constraint $\mathbb{Q}\in\boldsymbol{\mathfrak{X}}^{\alpha}(\mathbb{P}_N)$ will help us to develop a tractable algorithm for the dynamic attack strategy. 

Different with the dynamic attack strategy, the ranking parameter $\hat{\boldsymbol{\theta}}$ would be unknown for the adversary in the static attack strategy. The $\chi^2$ divergence will not help to simplify the min-max problem \eqref{opt:bilevel}. To sum up, we adopt the $p$-Wasserstein distance \cite{DBLP:journals/corr/abs-1712-06050} as the discrepancy measure between the empirical distribution $\mathbb{P}_n$ and the toxic distribution $\mathbb{Q}$ for the static attack strategy. The $p$-Wasserstein distance will help us to reformulate the min-max problem \eqref{opt:bilevel} into a single regularized problem. 

\begin{definition}[$p$-Wasserstein distance]
    Let $p\in[1,\infty]$. The $p$-Wasserstein distance between distributions $\mathbb{P},\ \mathbb{Q}\in\mathcal{P}(\boldsymbol{\Omega})$ is defined as
    \begin{itemize}
        \item $1\leq p< \infty$
            \begin{align}
                & \mathcal{W}_p\ (\mathbb{P},\ \mathbb{Q})\ \ \ \ \ =\numberthis\label{eq:p_Wasserstein}\\
                & \left(\underset{\gamma\in\Gamma(\mathbb{P},\ \mathbb{Q})}{\textbf{\textit{min}}}\left\{\mathlarger{\int}_{\boldsymbol{\Omega}\times\boldsymbol{\Omega}}\Big[d\Big(\boldsymbol{w},\ \boldsymbol{w}'\Big)\Big]^p\ \gamma\Big(\mathrm{d}\boldsymbol{w},\ \mathrm{d}\boldsymbol{w}'\Big)\right\}\right)^{\frac{1}{p}}\nonumber
            \end{align}
        \vskip 0.25cm
        \item $p=\infty$
        \begin{equation}
            \label{eq:infty_Wasserstein}
            \mathcal{W}_p\ (\mathbb{P},\ \mathbb{Q})\ \ \ =\underset{\gamma\in\Gamma(\mathbb{P},\ \mathbb{Q})}{\textbf{\textit{inf}}}\ \underset{\vphantom{\gamma\in\Gamma(\mathbb{P},\ \mathbb{Q})}\boldsymbol{\Omega}\times\boldsymbol{\Omega}}{\gamma\textnormal{-}\textbf{\textit{ess sup}}}\ d\Big(\boldsymbol{w},\ \boldsymbol{w}'\Big)
        \end{equation}
    \end{itemize}
    where $\Gamma(\mathbb{P},\ \mathbb{Q})$ denotes the set of all Borel probability distributions on $\boldsymbol{\Omega}\times\boldsymbol{\Omega}$ with marginal distributions $\mathbb{P}$ and $\mathbb{Q}$, $d:\boldsymbol{\Omega}\times\boldsymbol{\Omega}\rightarrow\mathbb{R}_+$ is a nonnegative function, and $\gamma\textnormal{-}\textbf{\textit{ess sup}}$ expresses the essential supremum of $d(\cdot,\ \cdot)$ with respect to the measure $\gamma$.
\end{definition}
The Wasserstein distance \eqref{eq:p_Wasserstein} and \eqref{eq:infty_Wasserstein} arise in the problem of optimal transport \cite{monge1781memoire,villani2008optimal}: for any coupling $\gamma\in\Gamma(\mathbb{P},\ \mathbb{Q})$, the conditional distribution $\gamma_{\boldsymbol{w}\vert\boldsymbol{w}'}$ can be viewed as a randomized overhead for ‘transporting’ a unit quantity of some material from a random location $\boldsymbol{w}\sim\mathbb{P}$ to another location $\boldsymbol{w}'\sim\mathbb{Q}$. If the cost of transportation from $\boldsymbol{w}\in\boldsymbol{\Omega}$ to $\boldsymbol{w}'\in\boldsymbol{\Omega}$ is given by $[d(\boldsymbol{w},\boldsymbol{w}')]^p$, $\mathcal{W}_p\ (\mathbb{P},\ \mathbb{Q})$ will be the minimum expected transport cost \cite{peyre2019computational}. 

Suppose that $\boldsymbol{\mathfrak{W}}_p(\mathbb{P}_N)$ is a set of probability distributions constructed from the empirical distribution $\mathbb{P}_N$ with $p$-Wasserstein distance. This Wasserstein ball of radius $\alpha$ is given by
\begin{equation}
    \boldsymbol{\mathfrak{W}}_p^{\alpha}(\mathbb{P}_N) = \left\{\ \mathbb{Q}\in\mathcal{P}(\boldsymbol{\Omega}_1)\ \Big\vert\ \mathcal{W}_p\ (\mathbb{P}_N,\ \mathbb{Q})\leq\alpha\ \right\}.
\end{equation}
With local uncertainty set $\boldsymbol{\mathfrak{W}}^{\alpha}_p(\mathbb{P}_N)$, the min-max optimization \eqref{opt:bilevel} could be expressed as the following distributionally robust optimization (DRO) problem:
\begin{equation}
    \label{opt:robust_game}
    \begin{aligned}
        \underset{\boldsymbol{\theta}\ \in\ \boldsymbol{\Theta}}{\vphantom{\mathbb{Q}^{\top}}\textbf{\textit{min}}}&\ \underset{\mathbb{Q}\ \in\ \boldsymbol{\mathfrak{W}}^{\alpha}_p(\mathbb{P}_N)}{\textbf{\textit{sup}}}\ \ \mathbb{E}_{\boldsymbol{w}\sim\mathbb{Q}}\Big[\ell(\boldsymbol{\theta},\ \boldsymbol{w})\Big],
    \end{aligned}
\end{equation}
where the supremum operation w.r.t. $\mathbb{Q}$ means that all players' optimal decision is based on the worst expected value of $\ell$ from the set of distributions $\boldsymbol{\mathfrak{W}}^{\alpha}_p(\mathbb{P}_N)$. Here we replace the minimal `cost' constraint in \eqref{opt:bilevel} by the neighborhood constraint on the worst-case expectation. With the local constraint $\mathbb{Q}\in\boldsymbol{\mathfrak{W}}^{\alpha}_p(\mathbb{P}_N)$, the Wasserstein distance between the empirical distribution $\mathbb{P}_N$ and the perturbed distribution $\mathbb{Q}$ must be smaller than a given budget $\alpha$ as $\mathcal{W}_p\ (\mathbb{P},\ \mathbb{Q})\leq\alpha$. It means that the attacker has a budget $\alpha$ to implement his/her perturbation on the original data for ranking aggregation. The robust game formulation \eqref{opt:robust_game} would relax the coarse-grid constraint as \eqref{eq:adversary_action_space_1}, and the analysis in the sequel reveals the central role played by this relaxation. 

Actually, the bi-level problem \eqref{opt:chi_2} and the DRO problem \eqref{opt:robust_game} relate to a general robust game \cite{Aghassi2006,liu2018distributionally,DBLP:conf/icores/Loizou16} between the attacker and the ranker as
\begin{equation}
    \underset{\boldsymbol{x}_r\ \in\ \mathcal{X}_r}{\textbf{\textit{min}}}\ \underset{\mathbb{Q}\ \in\ \boldsymbol{\mathfrak{U}}}{\textbf{\textit{sup}}}\ \ \mathbb{E}_{\boldsymbol{\xi}\sim\mathbb{Q}}\Big[f_r(\boldsymbol{x}_r,\ \boldsymbol{x}_{-r},\ \boldsymbol{\xi})\Big],\ \ r = 1,\ 2\ \label{opt:DRG:1}
\end{equation}
where $r$ indicates the role of the agent in the robust game, $\boldsymbol{x}_r$ is the decision variable of the special player $r$, and $\boldsymbol{x}_{-r}$ denotes the decision variables of its rivals, and $\mathcal{X}_r$ is the action set of player $r$. The random variable $\boldsymbol{\xi}$ illustrates the uncertainty or inaccuracy of distributional information to the players, and $\boldsymbol{\mathfrak{U}}$ is the uncertainty set of distribution of random variable $\boldsymbol{\xi}$ for all players (\textit{i.e.}, $\boldsymbol{\mathfrak{X}}^{\alpha}(\mathbb{P}_N)$ and $\boldsymbol{\mathfrak{W}}_p^{\alpha}(\mathbb{P}_N)$). The pay-off function $f_r$ could be different for each player and the corresponding game is a non-zero sum game. Comparing the general case \eqref{opt:DRG:1} with \eqref{opt:chi_2} and \eqref{opt:robust_game}, all players in \eqref{opt:robust_game} focus on the same pay-off function as $f_1 = f_2 = \ell$. Moreover, the decision variable of the ranker $\boldsymbol{\theta}$ equals to $\boldsymbol{x}_1$. The random variable $\boldsymbol{\xi}$ represents the distribution of pairwise comparison as $\boldsymbol{w}$. So the decision variable of the attacker $\boldsymbol{x}_2$ will be the constant (its role has been replaced by $\boldsymbol{\xi}$). The robust game problem is first proposed by Bertsimas and Aghassi in \cite{Aghassi2006}. It expands the boundaries of research of the classical Nash game \cite{vonneumann1947,10.2307/1969529,Nash48} and the Bayesian game \cite{doi:10.1287/mnsc.14.3.159,doi:10.1287/mnsc.14.5.320,doi:10.1287/mnsc.14.7.486}. Different form the Nash and the Bayesian game\cite{Aghassi2006}, the only common knowledge of all participants in robust game is that all players being aware about an uncertainty set like $\mathfrak{X}^{\alpha}(\mathbb{P}_N)$ and $\boldsymbol{\mathfrak{W}}_p^{\alpha}(\mathbb{P}_N)$. All possible parameters of payoff function are related to this set. Here we investigate the existence of the equilibrium for distributionally robust Nash equilibrium of the proposed model \eqref{opt:DRG:1}. First, we give the definition of the distributionally robust Nash equilibrium.

\begin{definition}
    A pair of different players' action $\{\boldsymbol{x}^*_1,\ \boldsymbol{x}^*_2\}$ is called a distributionally robust Nash equilibrium (DRNE) of \eqref{opt:DRG:1} if they satisfy the following 
    \begin{equation}
        \label{eq:drne_equilibrium}
        \boldsymbol{x}^*_r\ \in\ \underset{\boldsymbol{x}_r\ \in\ \mathcal{X}_r}{\vphantom{\mathbb{Q}\in\mathcal{Q}^\top_r}\textbf{\textit{arg min}}}\ \underset{\mathbb{Q}\ \in\ \boldsymbol{\mathfrak{U}}}{\textbf{\textit{sup}}}\ \ \mathbb{E}_{\boldsymbol{\xi}\sim\mathbb{Q}}\Big[f_r(\boldsymbol{x}_r,\ \boldsymbol{x}_{-r},\ \boldsymbol{\xi})\Big],\ r=1,2.
    \end{equation}
\end{definition}

Next, we can prove the existence of DRNE for the general robust game \eqref{opt:DRG:1}.
\begin{restatable}{theorem}{equilibrium}
\label{thm:robust_nash_equilibrium}
Let the pay-off function $f_r,\ r=1,2$ be the weighted sum-of-squared loss $\ell$ \eqref{eq:weight_l2_loss} in \eqref{opt:DRG:1}. If the uncertainty set is $\boldsymbol{\mathfrak{X}}^{\alpha}(\mathbb{P}_N)$ or $\boldsymbol{\mathfrak{W}}_p^{\alpha}(\mathbb{P}_N)$, the general robust game \eqref{opt:DRG:1} has a DRNE.
\end{restatable}
To prove this existence result, we reformulate the problem \eqref{opt:DRG:1} into a single optimization problem and show that the single problem has an optimal solution. The detailed proof is provided in the Appendix \ref{app:theorem_1}. 



\subsection{Optimization}
In this part we show our algorithms for computing the adversarial strategies. Suppose the total number of pairwise comparison without perturbation is $M^0$, and the frequencies of each type of the observed  comparisons are 
\begin{equation}
    \label{eq:fre_p}
    \boldsymbol{p} = \frac{1}{M^0}\cdot\boldsymbol{w}'_0,\ \ M^0 = \sum_{(i,j)}\ {w^0_{ij}}'.
\end{equation}
Let the maximum toxic dosage be $\kappa$. It suggests that the number of toxic pairwise comparisons $M$ satisfies 
\begin{equation}
    M\ =\ \underset{(i,j)}{\sum}\ w_{ij}\ \leq\ (1+\kappa)\cdot M^0.
\end{equation}
We replace the toxic weight $\boldsymbol{w}$ with its frequency $\boldsymbol{q}=\{q_{ij}\}\in\mathbb{R}^N_+$ when analyzing the equilibrium, studying the statistical nature of the worst-case estimator and solving the corresponding optimization problem. We relax the integer programming problem into a general optimization by such a variable substitution. Thus, the pay-off function \eqref{eq:weight_l2_loss} turns to be
\begin{align}
    \ell(\boldsymbol{\theta},\ \boldsymbol{q})=\frac{1}{2N}\ \underset{(i,j)}{\sum}\ q_{ij}\big(y_{ij}-\theta_i+\theta_j\big)^2,\label{eq:new_loss}
\end{align}
and we still adopt $\mathbb{P}_N$ and $\mathbb{Q}$ as the distribution of the empirical data and the toxic data. Furthermore, we can implement the integer attack with the optimal $\boldsymbol{q}$ and $M$. Now we come to solve the bi-level optimization \eqref{opt:chi_2} and the distributionally robust optimization problem \eqref{opt:robust_game} with the variable substitution:
\begin{equation}
    \label{opt:chi_2_2}
    \begin{aligned}
        & &\underset{\mathbb{Q}\ \in\ \boldsymbol{\mathfrak{X}}^{\alpha}(\mathbb{P}_N)}{\textbf{\textit{max}}}&\ \ \mathbb{E}_{\boldsymbol{q}\sim\mathbb{Q}}\big[\ell(\boldsymbol{q};\ \boldsymbol{\hat{\theta}})\big],\\
        & &\textbf{\textit{subject to}}&\ \ \boldsymbol{\hat{\theta}}=\underset{\boldsymbol{\theta}\ \in\ \boldsymbol{\Theta}}{\textbf{\textit{arg min}}}\ \ \ell(\boldsymbol{\theta};\ \boldsymbol{w}'_0),
    \end{aligned}  
\end{equation}
and
\begin{align}
    \label{opt:robust_game_2}
    \underset{\boldsymbol{\theta}\ \in\ \boldsymbol{\Theta}}{\vphantom{\mathbb{Q}\ \in\ \boldsymbol{\mathfrak{W}}^{\alpha}_p\ (\mathbb{P}_N)}\textbf{\textit{min}}}&\ \underset{\mathbb{Q}\ \in\ \boldsymbol{\mathfrak{W}}^{\alpha}_p\ (\mathbb{P}_N)}{\textbf{\textit{sup}}}\ \ \mathbb{E}_{\boldsymbol{q}\sim\mathbb{Q}}\ \Big[\ \ell\big(\boldsymbol{\theta};\ \boldsymbol{q}\big)\ \Big]
\end{align}

For the dynamic attack strategy \eqref{opt:chi_2_2}, a similar formulation has been studied for archiving a better variance-bias trade-off in maximum likelihood estimation\cite{DBLP:conf/nips/NamkoongD17}. Based on the $\chi^2$-divergence, the bi-level problem \eqref{opt:chi_2_2} turns to be a convex problem. We provide a detailed process of solving \eqref{opt:chi_2_2} in the supplementary material. 

The distributionally robust optimization formulation \eqref{opt:robust_game_2} involves optimizing over the uncertainty set $\boldsymbol{\mathfrak{W}}^{\alpha}_2(\mathbb{P}_N)$, which contains countless probability measures. However, recent strong duality results of distributionally robust optimization involving Wasserstein uncertainty set \cite[Theorem 1]{gao2016distributionally} and \cite[Theorem 1]{doi:10.1287/moor.2018.0936}) ensure that the inner supremum in \eqref{opt:robust_game_2} admits an equivalent reformulation which would be a tractable, univariate optimization problem. In the adversarial scenario of pairwise ranking, we have the following result. The DRO problem \eqref{opt:robust_game_2} could be reformulated as a regularized regression problem.

\begin{restatable}{theorem}{mainresult}
    \label{thm:main_result}
    Let $\mathcal{Z}=\big\{\boldsymbol{p},\ \boldsymbol{B},\ \boldsymbol{y}'\big\}$ be the observed data set, where $\boldsymbol{B}$ and $\boldsymbol{y}'$ are defined as \eqref{eq:fixed_design_y}, $\boldsymbol{p}$ is the frequency of each type of pairwise comparison as \eqref{eq:fre_p}. Consider the loss function of $\boldsymbol{z}$, and the distance function between $\boldsymbol{z}_{\boldsymbol{c}}$, $\boldsymbol{z}'_{\boldsymbol{c}}$ are based on the $\ell_2$-norm. In other words, we take $\ell(\boldsymbol{\theta},\ \boldsymbol{z})$ as \eqref{eq:new_loss} and
        \begin{equation}
            \label{eq:metric}
            \begin{aligned}
                & d(\boldsymbol{z}_{\boldsymbol{c}},\ \boldsymbol{z}'_{\boldsymbol{c}})&=&\ \ \big\|\big(p_{ij},\ \boldsymbol{b}_{i,j},\ y'_{ij}\big)-\big(q_{ij},\ \boldsymbol{b}_{i,j},\ y'_{ij}\big)\big\|_2\\
                & &=&\ \ \ \big|\ p_{ij}-q_{ij}\ \big|.
            \end{aligned}
        \end{equation}
        Then, the DRO problem \eqref{opt:robust_game_2} has an equivalent form:
        \begin{equation}
            \label{eq:final_theta}
            \begin{aligned}
                & & & \ \ \underset{\boldsymbol{\theta}\ \in\ \boldsymbol{\Theta}}{\vphantom{\mathbb{Q}\ \in\ \boldsymbol{\mathfrak{W}}^{\alpha}_p\ (\mathbb{P}_N)}\textbf{\textit{min}}}\ \ \underset{\mathbb{Q}\ \in\ \boldsymbol{\mathfrak{W}}^{\alpha}_2(\mathbb{P}_N)}{\textbf{\textit{sup}}}\ \ \mathbb{E}_{\boldsymbol{q}\sim\mathbb{Q}}\Big[\ \ell\big(\boldsymbol{\theta};\ \boldsymbol{q}\big)\ \Big]\\
                & &=&\ \ \underset{\boldsymbol{\theta}\ \in\ \boldsymbol{\Theta}}{\vphantom{\mathbb{Q}\ \in\ \mathfrak{W}^{\alpha}_p(\mathbb{P}_N)}\textbf{\textit{min}}}\ \ \mathcal{L}(\boldsymbol{\theta})\ +\ \mathcal{R}(\boldsymbol{\theta}),
            \end{aligned}
        \end{equation}
        where
        \begin{equation}
            \label{eq:final_loss}
            \mathcal{L}(\boldsymbol{\theta}) = \frac{1}{2N}\underset{(i,\ j)}{\sum}\ p_{ij}(y'_{ij}-\boldsymbol{\theta}^\top\boldsymbol{b}_{i,j})^2,
        \end{equation}
        and
        \begin{equation}
            \label{eq:final_reg}
            \mathcal{R}(\boldsymbol{\theta}) = \sqrt{\frac{\alpha}{4N}\underset{(i,\ j)}{\sum}(y'_{ij}-\boldsymbol{\theta}^\top\boldsymbol{b}_{i,j})^2}.
        \end{equation} 
         
\end{restatable}

We provide a detailed proof in the Appendix \ref{app:theorem_2}. The example of linear regression with Wasserstein distance based uncertainty sets has been considered in \cite{Blanchet_2019}. The representation for regularized linear regression in Theorem \ref{thm:main_result} can be seen as an extension of \cite{Blanchet_2019}. We adopt the weighted sum-of-squared loss and the ``regularization'' \eqref{eq:final_reg} here is not the $\ell_2$-norm of $\boldsymbol{\theta}$. \eqref{eq:final_reg} can be treated as a ``regularization'' which is the square root of the residual between $y'_{ij}$ and its estimation. It represents a `worst' case in pairwise ranking: all possible comparisons appear and they have the same number of votes. In this case, the pairwise ranking algorithm could not generate a reasonable ranking result. The uncertainty budget $\alpha$ play the role as the regularization parameter. As $\alpha$ increase, the ranking scores $\boldsymbol{\theta}$ obtained by \eqref{eq:final_theta} would come closer to the solution of \eqref{eq:final_reg}. The validity of the analysis above will be illustrated in the empirical studies.

With Theorem \ref{thm:main_result}, we will have the following corollary which gives a tractable method to obtain the worst-case distribution $\boldsymbol{q}^*_{\alpha}$. If we have the worst-case solution, we can solve the corresponding dual variable from the optimal vale of the original DRO problem.

\begin{corollary}[]
    \label{coll:1}
    For $\lambda\geq0$ and the weighted least square loss \eqref{eq:new_loss}, we define $\psi:\mathbb{R}^N\rightarrow\mathbb{R}$
    \begin{equation}
        \label{eq:psi_2}
        \begin{aligned}
            & & &\ \ \ \ \psi_{\lambda,\ \ell}(\boldsymbol{p};\ \boldsymbol{\theta})\\
            & & =&\ \ \underset{\boldsymbol{z}'\in\mathbb{R}^{n+2}}{\textbf{\textit{sup}}}\ \frac{1}{N}\underset{(i,j)}{\sum}\Big\{\ell(\boldsymbol{\theta};\ q_{ij})-\lambda(p_{ij}-q_{ij})^2\Big\}   
        \end{aligned}
    \end{equation}
    where
    \begin{equation}
        \label{eq:single_loss}
        \ell(\boldsymbol{\theta};\ q_{ij}) = \frac{q_{ij}}{2}\big(y'_{ij}-\theta_i+\theta_j\big)^2.
    \end{equation}
    Let 
    \begin{equation}
        \label{eq:primal_problem}
        I_{\textit{primal}}=\underset{\mathbb{Q}\ \in\ \boldsymbol{\mathfrak{W}}^{\alpha}_2(\mathbb{P}_N)}{\textbf{\textit{sup}}}\ \mathbb{E}_{\boldsymbol{z}'\sim\mathbb{Q}}\ \Big[\ell\big(\boldsymbol{\theta};\ \boldsymbol{z}'\big)\Big], 
    \end{equation}
    we have
    \begin{equation}
        \label{eq:optimal_value}
       I_{\textit{primal}} = \underset{\lambda\geq 0}{\textbf{\textit{inf}}}\ \Bigg\{\ \lambda\alpha+\mathbb{E}_{\boldsymbol{z}'\sim\mathbb{Q}}\Big[\psi_{\lambda,\ \ell}(\boldsymbol{p};\ \boldsymbol{\theta})\Big]\Bigg\}.
    \end{equation}
    Moreover, let $\boldsymbol{\theta}^*_{\alpha}$ be the optimal solution of the right hand side of \eqref{eq:final_theta} and the dual variable of $\boldsymbol{\theta}^*_{\alpha}$ is $\lambda^*_{\alpha}$ will be a solution of \eqref{eq:optimal_value}:
    \begin{equation}
        \label{eq:opt_dual}
        \lambda^*_{\alpha} = \sqrt{\frac{1}{16N\alpha}\cdot\underset{(i,j)}{\sum}\big(y'_{ij}-(\boldsymbol{\theta}^*_{\alpha})^\top\boldsymbol{b}_{i,j}\big)^2}.
    \end{equation}
    The optimal static attack strategy $\boldsymbol{q}^*_{\alpha}$ is a solution of \eqref{eq:psi_2} corresponding to $\boldsymbol{\theta}^*_{\alpha}$ and $\lambda^*_{\alpha}$:
    \begin{equation*}
        \label{eq:toxic_distribution}
        \boldsymbol{q}^*_{\alpha} = \underset{\boldsymbol{q}\ \in\ \mathbb{R}^N_+}{\textbf{\textit{arg max}}}\ \underset{(i,j)}{\sum}\Bigg\{q_{ij}\Big(y'_{ij}-(\boldsymbol{\theta}^*_{\alpha})^\top\boldsymbol{a}_{\boldsymbol{c}}\Big)^2-\lambda^*_{\alpha}|p_{ij}-q_{ij}|^2\Bigg\}.
    \end{equation*}
\end{corollary}

\begin{algorithm}[h!]
    \SetAlgoLined
    \SetKwInOut{Input}{Input}
    \SetKwInOut{Output}{Output}
    \Input{the original data $\{\boldsymbol{w}'_0,\ \boldsymbol{B},\ \boldsymbol{y}'\}$, the maximum toxic dosage $\kappa$, the uncertainty budget $\alpha$.}
    
    {Initialize} the frequency of weights $\boldsymbol{p}$ by \eqref{eq:fre_p}

    Obtain the worst-case ranking scoring $\boldsymbol{\theta}^*_{\alpha}$ under the uncertainty budget $\alpha$
    \begin{equation*}
        \scriptstyle
        \boldsymbol{\theta}^*_{\alpha}\ \in\ \underset{\boldsymbol{\theta}\in\boldsymbol{\Theta}}{\vphantom{\mathbb{Q}\in\mathfrak{W}^{\alpha}_p(\mathbb{P}_N)}\textbf{\textit{arg min}}}\ \left(\sqrt{\frac{\alpha}{4N}\underset{(i,j)}{\sum}(y'_{ij}-\boldsymbol{\theta}^\top\boldsymbol{b}_{i,j})^2}+\frac{1}{2N}\underset{(i,j)}{\sum}p_{ij}(y'_{ij}-\boldsymbol{\theta}^\top\boldsymbol{b}_{i,j})^2\right).
    \end{equation*}

    Calculate the optimal dual variable through \eqref{eq:opt_dual}
    \begin{equation*}
        \scriptstyle
        \lambda^*_{\alpha}\ =\ \sqrt{\frac{1}{16N\alpha}\underset{(i,j)}{\sum}\big(y'_{ij}-(\boldsymbol{\theta}^*_{\alpha})^\top\boldsymbol{b}_{i,j}\big)^2}.
    \end{equation*} 

    Obtain the toxic distribution $\boldsymbol{q}^*_\alpha$ corresponding to $\boldsymbol{\theta}^*_{\alpha}$ and $\lambda^*_{\alpha}$
    \begin{equation*}
        \scriptstyle
        \boldsymbol{q}^*_\alpha\ \in\ \underset{\boldsymbol{q}\in\mathbb{R}^N_+}{\textbf{\textit{arg max}}}\ \underset{(i,j)}{\sum}\ \Big\{q_{ij}(y'_{ij}-(\boldsymbol{\theta}^*_{\alpha})^\top\boldsymbol{b}_{i,j})^2-\lambda^*_{\alpha}|p_{ij}-q_{ij}|^2\Big\}.
    \end{equation*}

    Assign the toxic weights with $\boldsymbol{q}^*_\alpha$
    \begin{equation*}
        \scriptstyle
        \boldsymbol{w}_\alpha\ =\ M_0(1+\kappa)\cdot\boldsymbol{q}^*_\alpha.
    \end{equation*}

    Round $\boldsymbol{w}_\alpha$ to obtain the $\boldsymbol{w}^*_\alpha$ as integer vector
    \begin{equation*}
        \scriptstyle
        \boldsymbol{w}^*_\alpha\ =\ \textbf{\textit{rounding}}(\boldsymbol{w}_\alpha).
    \end{equation*}

    \Output{the poisoned data $\{\boldsymbol{w}^*_{\alpha},\ \boldsymbol{B},\ \boldsymbol{y}'\}$.}
    \caption{\small Static Poisoning Attack on Pairwise Ranking.}
    \label{alg:main}
\end{algorithm}

Finally, we describe the whole optimization of the static poisoning attack on pairwise ranking with \textbf{Algorithm \ref{alg:main}}. First, the adversary changes the original weight $\boldsymbol{w}'_0$ into the frequency $\boldsymbol{p}$ as the initialization (line 1). By \textbf{Theorem \ref{thm:main_result}}, the attacker could obtain the worst-case estimation $\boldsymbol{\theta}^*_{\alpha}$ through \eqref{eq:final_theta} (line 2). But the attacker cannot adopt $\boldsymbol{\theta}^*_{\alpha}$ as the attack operation. Here we solve the dual variable $\lambda^*_{\alpha}$ (line 4) to find the toxic distribution. Then the toxic distribution $\boldsymbol{q}^*_{\alpha}$ with uncertainty budget $\alpha$ is obtained by \textbf{Corollary \ref{coll:1}} (line 4). With some rounding operation (line 5\ \&\ 6), the adversary prepares the poisoned data $\{\boldsymbol{w}^*_{\alpha},\ \boldsymbol{B},\ \boldsymbol{y}'\}$. Then the poisoned data is provided to the ranker who solved the ranking parameter by \eqref{eq:last_step}. Then the whole poisoning process will be completed.

\subsection{Theoretical Analysis}

In this section, we come back to \eqref{opt:robust_game_2} and give a couple of inequalities relating the local worst-case (or local minimax) risks and the usual statistical risks of the pairwise ranking under adversarial conditions. In the traditional paradigm of statistical learning\cite{Vapnik:1995:NSL:211359}, we have a class of probability measures $\mathcal{P}$ on a measurable instance space $\mathcal{Z}$ and a class $\mathcal{F}$ of measurable functions $\ell:\mathcal{Z}\rightarrow\mathbb{R}_+$. Each $\ell\in\mathcal{F}$ quantifies the loss of a certain decision rule or a hypothesis. With a slight abuse of terminology, we will refer to $\mathcal{F}$ as the hypothesis space. The (expected) risk of a hypothesis $\ell$ on instances generated according to $\mathbb{P}\in\mathcal{P}(\mathcal{Z})$ is given by
\begin{equation}
    R_{\mathbb{P}}(\ell) := \mathbb{E}_{\boldsymbol{z}\sim\mathbb{P}}\big[\ell(\boldsymbol{z})\big] = \mathlarger{\int}_{\mathcal{Z}}\ \ell(\boldsymbol{z})\ \mathbb{P}(\mathrm{d}\boldsymbol{z}).
\end{equation}
Given an $N$-tuple $\{\boldsymbol{z}_1,\dots,\boldsymbol{z}_N\}$ of \textit{i.i.d.} training examples drawn from an unknown distribution $\mathbb{P}\in\mathcal{P}$, the objective is to find a hypothesis $f\in\mathcal{F}$ whose risk $R(\mathbb{P},\ \ell)$ is close to the minimum risk
\begin{equation}
    R^*_{\mathbb{P}}(\mathcal{F}):=\underset{\ell\ \in\ \mathcal{F}}{\textbf{\textit{inf}}}\ R_{\mathbb{P}}(\ell)
\end{equation}
with high probability. Under some suitable regularity assumptions, this objective can be accomplished via Empirical Risk Minimization (ERM):
\begin{equation}
    \label{eq:empirical_risk}
    R_{\mathbb{P}_N}(\ell) := \frac{1}{N}\ \sum^N_{i=1}\ \ell(\boldsymbol{z}_c)
\end{equation}
and the minimum empirical risk is 
\begin{equation}
     R^*_{\mathbb{P}_N}(\mathcal{F}):=\underset{\ell\ \in\ \mathcal{F}}{\textbf{\textit{min}}}\ R_{\mathbb{P}_N}(\ell),
\end{equation}
where $\mathbb{P}_N=\frac{1}{N}\sum^N_{c=1}\delta_{\boldsymbol{z}_c}$ is the empirical distribution of the training examples. Meanwhile, the minimax risk\cite{DBLP:conf/nips/LeeR18} can be defined as 
\begin{equation}
    \hat{R}_{\mathbb{P}_N}(\mathcal{F}):=\ \underset{\ell\ \in\ \mathcal{F}\vphantom{\mathbb{Q}\ \in\ \mathfrak{W}(\mathbb{P}_N)}}{\textbf{\textit{min}}\vphantom{\textbf{\textit{sup}}}}\ \ \underset{\mathbb{Q}\ \in\ \boldsymbol{\mathfrak{W}}(\mathbb{P}_N)}{\textbf{\textit{sup}}}\ R_{\mathbb{Q}}(\ell)
\end{equation}

We assume that the instance space $\mathcal{Z}$ is a Polish space (\textit{i.e.}, a complete separable metric space) with metric $d_{\mathcal{Z}}$. We denote by $\mathcal{P}(\mathcal{Z})$ the space of all Borel probability measures on $\mathcal{Z}$, and by $\mathcal{P}_m(\mathcal{Z})$ with $m\geq 1$ the space of all $\mathbb{P}\in \mathcal{P}(\mathcal{Z})$ with finite $m\text{-\textit{th}}$ moments. The metric structure of $\mathcal{Z}$ can be used to define a family of metrics on the spaces $\mathcal{P}_m(\mathcal{Z})$. We then define the local worst-case risk of $\ell$ at $\mathcal{P}$,
\begin{equation}
    R_{\mathbb{P},\alpha,p}(\ell) := \underset{\mathbb{Q}\ \in\ \boldsymbol{\mathfrak{W}}^{\alpha}_p(\mathbb{P}_N)}{\textbf{\textit{sup}}}\ R_{\mathbb{Q}}(\ell)
\end{equation}
and the local minimax risk of $\mathcal{P}$,
\begin{equation}
    \label{eq:local_minimax_risk}
    R^*_{\mathbb{P},\alpha,p}(\mathcal{F}) := \underset{\ell\ \in\ \mathcal{F}}{\textbf{\textit{inf}}}\ R_{\mathbb{P},\alpha,p}(\ell).
\end{equation}




Next, we analyze the performance of the local minimax ERM procedure of the pairwise ranking, namely, 
\begin{equation}
    \label{eq:minimax_erm}
    \hat{\ell}\in\underset{\ell\ \in\ \mathcal{F}}{\textbf{\textit{arg min}}}\ R_{\mathbb{P}_N,\alpha,2}(\ell).
\end{equation}


\begin{restatable}{theorem}{worstprob}
    \label{thm:worst_prob}
    Consider the setting of pairwise ranking problem with the sum-of-squared loss, for any $t>0$, it holds
    \begin{equation}
        \label{eq:union_bound}
        \Pr\Big(\ \exists\ \ell\in\mathcal{F}: R_{\mathbb{P}, \alpha, 2}(\ell)>\varsigma_1\ \Big)\ \leq\ e^{-2t^2}
    \end{equation}
    and
    \begin{equation}
        \label{eq:empirical_bound_1}
        \ \ \Pr\Big(\exists\ \ell\in\mathcal{F}: R_{\mathbb{P}_N, \alpha, 2}(\ell)>\varsigma_2\Big)\ \leq\ 2e^{-2t^2}
    \end{equation}
    where
    \begin{equation}
        \begin{aligned}
            \varsigma_1=\underset{\lambda \geq 0}{\textbf{\textit{min}}}\ \ \Bigg\{\ \lambda\alpha^2\ +\ \mathbb{E}_{\boldsymbol{z}\sim\mathbb{Q}}\ \big[\ \psi_{\lambda,\ell}\ (\boldsymbol{z})\ \big]\Bigg\}+\frac{24\mathcal{J}(\mathcal{F})+t}{\sqrt{N\ }}.
        \end{aligned}
    \end{equation}
    and
    \begin{equation}
        \begin{aligned}
            & \varsigma_2=\underset{\lambda \geq 0}{\textbf{\textit{min}}}\ \ \Bigg\{\ (\lambda+1)\alpha^2\ +\ \mathbb{E}_{\boldsymbol{z}\sim\mathbb{Q}}\ \big[\ \psi_{\lambda,\ell}\ (\boldsymbol{z})\ \big]\\
            & {\color{white}\underset{\lambda \geq 0}{\textbf{\textit{min}}}\ \ \Bigg\{}\ \ \ \ \ \ +\frac{\sqrt{\textbf{\textit{log}}(\lambda+1)}}{\sqrt{N\ }}\ \Bigg\}+\frac{24\mathcal{J}(\mathcal{F})+t}{\sqrt{N\ }},
        \end{aligned}
    \end{equation}
    where $\mathcal{J}(\mathcal{F})$ is the Dudley’s entropy integral \cite{DUDLEY1967290}, which is served as the complexity measure of the hypothesis class $\mathcal{F}$.
\end{restatable}

Theorem \ref{thm:worst_prob} is a type of data-dependent generalization bounds which is proposed for margin cost function class \cite{koltchinskii2002,DBLP:conf/nips/LeeR18}. By the strong duality results, we can establish this result from the dual representation of the Wasserstein DRO problem. The detailed proof is provided in the Appendix \ref{app:thm:worst_prob}. Here we note that the hypothesis selected by the minimax ERM procedure \eqref{eq:minimax_erm} are uniform smoothness with respect to the underlying metric $d_{\mathcal{Z}}(\cdot,\cdot)$. Further, we have the following result. Proofs are relegated to the Appendices \ref{app:thm:excess_risk}.

\begin{restatable}{theorem}{excessrisk}
    \label{thm:excess_risk}
    Consider the setting of pairwise ranking problem with the sum-of-squared loss, the following holds with probability as least $1-\eta$
    \begin{equation}
        \begin{aligned}
             & & &\ \ \ R_{\mathbb{P},\alpha,2}(\hat{\ell})-R^*_{\mathbb{P},\alpha,2}(\mathcal{F})\\
             & &\leq&\ \ \frac{48\mathcal{J}(\mathcal{F})}{\sqrt{N}}+\frac{48L\big[\text{diam}(\mathcal{Z})\big]^2}{\alpha\sqrt{N}} + 3\sqrt{\frac{\textbf{\textit{log}}(\frac{2}{\eta})}{2N}},
        \end{aligned}
    \end{equation}
    where $\text{diam}(\mathcal{Z})$ is the diameter of $\mathcal{Z})$
    \begin{equation}
            \text{diam}(\mathcal{Z}) = \underset{\boldsymbol{z},\boldsymbol{z}'\in\mathcal{Z}}{\textbf{\textit{sup}}} d_{\mathcal{Z}}(\boldsymbol{z},\ \boldsymbol{z}').
    \end{equation}
\end{restatable}

\section{Experiments}

In this section, four examples are exhibited with both simulated and real-world data to illustrate the validity of the proposed poisoning attack on pairwise ranking. The first example is with simulated data while the latter three exploit real-world datasets involved crowdsourcing, election and recommendation.

\subsection{Simulated Study}

\textbf{Settings. }We first validate our poisoning attack framework on simulated data. We create a random total ordering on set $\boldsymbol{V}$ with $n$ candidates as the ground-truth ranking and generate the comparison matrix $\boldsymbol{B}$ and the labels $\boldsymbol{y}'$ as \eqref{eq:fixed_design_y}. Next, we generate the ground-truth weight of each comparisons $\boldsymbol{w}_0$. Notice that the original data $\{\boldsymbol{w}_0,\ \boldsymbol{B},\ \boldsymbol{y}'\}$ consists of some noisy comparisons. In the simulation study, we can specify the percentage of noisy comparisons, denoted as $\varrho$. We validate the proposed attack framework when $n$, $\boldsymbol{w}_0$ and $\varrho$ vary. Moreover, the maximum toxic dosage $\kappa$ and the uncertainty budget $\alpha$ are the hyper-parameters of the Algorithm \ref{alg:main}. Since the annotations of pairwise data are usually collected via crowdsourcing platforms where the attacker could produce hundreds of zombie accounts easily to inject the poisoned pairwise comparisons, we also vary $\kappa$ and $\alpha$ in our experiments. At last, there exists a rounding operator in the Algorithm \ref{alg:main} and we explore the results of different rounding functions, \textit{e.g.} ceiling, floor, and the nearest integer of each element in $\boldsymbol{w}_{\alpha}$.
\\
\\
\noindent\textbf{Competitors. }To the best of our knowledge, the proposed method is the first poisoning attack on pairwise ranking. To see whether our proposed method could provide efficient perturbation data for misleading the pairwise ranking algorithm, we implement the random perturbation attack (referred to as `Random') and the Stackelberg or dynamic game attack (referred to as `Dynamic') as the competitors.  
\begin{itemize}[leftmargin=*]
    \vspace{0.05cm}
    \item The \textbf{\textit{random perturbation attack}} modifies $\boldsymbol{w}_0$ as $\boldsymbol{w}_{\text{random}}$ to manipulate the ranking result. The random perturbation attack generates $\boldsymbol{w}_{\text{random}}$ and obeys the constraints \eqref{eq:adversary_constraint_1} and \eqref{eq:adversary_constraint_2} to hide his/her behaviors. We vary $b$ and $l$ to explore the ability of random attack. The random perturbation data is noted as $\mathcal{Z}_{\text{random}} = \{\boldsymbol{A},\ \boldsymbol{y},\ \boldsymbol{w}_{\text{random}}\}$. We assume this attacker is also lack of prior knowledge on the true ranking. So the random perturbation attack also adopts the fixed label set $\boldsymbol{y}$.
    \vspace{0.1cm}
    \item The \textbf{\textit{Stackelberg (dynamic) game attack}} comes from \eqref{opt:Stackelberg}. To execute this type of poisoning attack, the adversary would have the full knowledge of original training data $\boldsymbol{w}_0$ and the corresponding relative ranking score $\boldsymbol{\theta}_{\text{original}}$. With these advantages, the adversary can adjust his/her strategies to provide the optimal malicious action with the bi-level optimization like \eqref{opt:Stackelberg}. Without a doubt, the adversary endues with the privilege by such a hierarchical relation. For the fair competition, we only perform one round of the leader-follower game as the other competitors. Notice that this kind to attack is also proposed by this paper. Due to the length limitation, we provide the details of this attack in the supplementary materials.
\end{itemize}
It is worth noting that the poisoning attack with dynamic game is not a practical attack method. \eqref{opt:Stackelberg} is a bi-level optimization and the maximization process needs the solution of the minimization problem. In other words, the attacker must obtain the relative ranking score $\boldsymbol{\hat{\theta}}$ estimated from the original training data without perturbation. This operation is much harder than injecting some modified training samples into the victim's training set. Only the so-called ``\textbf{\textit{white-box}}'' setting would satisfy its necessary requirements. As the `Dynamic' method needs more exorbitant conditions, the `Dynamic' method only reflects the vulnerability of ranking aggregation algorithms but can not show the superiority of the `Static' method.

\vspace{0.25cm}
\noindent\textbf{Evaluation Metrics. }We adopt the following measures for evaluating the ranking results aggregated by the different sets of pairwise comparisons.
\begin{itemize}[leftmargin=*]
    \vspace{0.05cm}
    \item \textbf{\textit{Kendall $\tau$ Distance (Kendall-$\tau$).}} The Kendall rank correlation coefficient evaluates the degree of similarity between two sets of ranks given the same objects. This coefficient depends upon the number of inversions of pairs of objects which would be needed to transform one rank order into the other. Let $V=[n]$ be a set of $n$ candidates and $\pi_1,\ \pi_2$ are two total orders or permutations on $V$, the Kendall $\tau$ distance is defined to be 
    \begin{equation}
        d_K(\pi_1,\ \pi_2) = \frac{2}{n(n-1)}\cdot\vartheta,
    \end{equation}
    where
    \begin{equation}
        \vartheta = \sum_{i=1}^{n-1}\sum_{j=i+1}^n\vartheta(\pi_1(i),\ \pi_1(j),\ \pi_2(i),\ \pi_2(j))
    \end{equation}
    is the number of different pairs between these two ordered sets $\pi_1, \pi_2$ as
    \begin{equation}
        \begin{aligned}
            & & &\ \ \vartheta(\pi_1(i),\ \pi_1(j),\ \pi_2(i),\ \pi_2(j))\\
            & &=& 
            \left\{\begin{matrix}
            \ \ 1, & \text{if\ } (\pi_1(i)-\pi_1(j))(\pi_2(i)-\pi_2(j))>0,\\ 
            -1,    & \text{if\ } (\pi_1(i)-\pi_1(j))(\pi_2(i)-\pi_2(j))<0,\\ 
            \ \ 0, & \text{otherwise, }
            \end{matrix}\right.            
        \end{aligned}
    \end{equation}
    and $\pi_1(i)$ represents the ranking score of the $i^{th}$ object in ranking list $\pi_1$. Kendall $\tau$ distance counts the number of pairwise mismatches between two rank orders. Then this metric considers all candidates of $V$. However, Kendall-$\tau$ ignores the importance of the top objects in a ranking list.
    \\
    \item \textbf{\textit{Reciprocal Rank (R-Rank).}} The reciprocal rank is a statistic measure for evaluating any process that produces an order list of possible responses to a sample of queries, ordered by the probability of correctness or the ranking scores. The reciprocal rank of a rank order is the multiplicative inverse of the rank of the first correct object:
    \begin{equation}
         RR = \frac{1}{\text{rank}_i},
    \end{equation}
    where ${\text{rank}_i}$ refers to the rank position of the first candidates of the ground-truth ranking in the other list.
    \\
    \item \textbf{\textit{Precision at $K$ (P$@K$).}} Precision at $K$ is the proportion of the top-$K$ objects in the other rank order that are consistent with the true ranking. In this case, the precision and recall will be the same. So we do not report the recall and F score for our poisoning attack method.  
    \\
    \item \textbf{\textit{Average Precision at $K$ (AP$@K$).}} Average precision at K is a weighted average of the precision. If the top objects in the new ranking list are consistent with the true ranking, they will contribute more than the tail objects in this metric. 
    \\
    \item \textbf{\textit{Normalized Discounted Cumulative Gain at K (NDCG$@K$).}} Using a graded relevance scale of objects in ranking result, discounted cumulative gain (DCG) measures the usefulness, or gain, of the objects based on its position in the order list when recovering to the true ranking. The gain is accumulated from the top to the bottom, with the gain of each result discounted at lower ranks. Compared to DCG, NDCG will be normalized by the ideal DCG.
\end{itemize}

\vspace{0.25cm}

\begin{table*}[]
\centering
\caption{Comparative results of different attack methods on simulated data.}
\label{table:simulation}
\tiny
\parbox{0.45\linewidth}
{
    \centering
    \begin{tabular}{@{}cclcc@{}c@{}c@{}c@{}}
    \toprule
    Method                   & Budget    & Kendall-$\tau$          & Tendency (ideal)  & R-Rank\ \ \ & P$@3$\ \ \ & AP$@3$\ \ \ & NDCG$@3$\ \ \\ \midrule
    Original                 & -         & \ 1.0000                & -                 & 1.0000 & 1.0000 & 1.0000  & 1.0000 \\ \midrule
    Random                   & 0.05/0.05 & \ 0.9556                & -                 & 1.0000 & 1.0000 & 1.0000  & 1.0000 \\ \midrule
    \multirow{7}{*}{Static}  & $10^{-6}$ & \ 1.0000                & \multirow{7}{*}{\includegraphics[height=12mm]{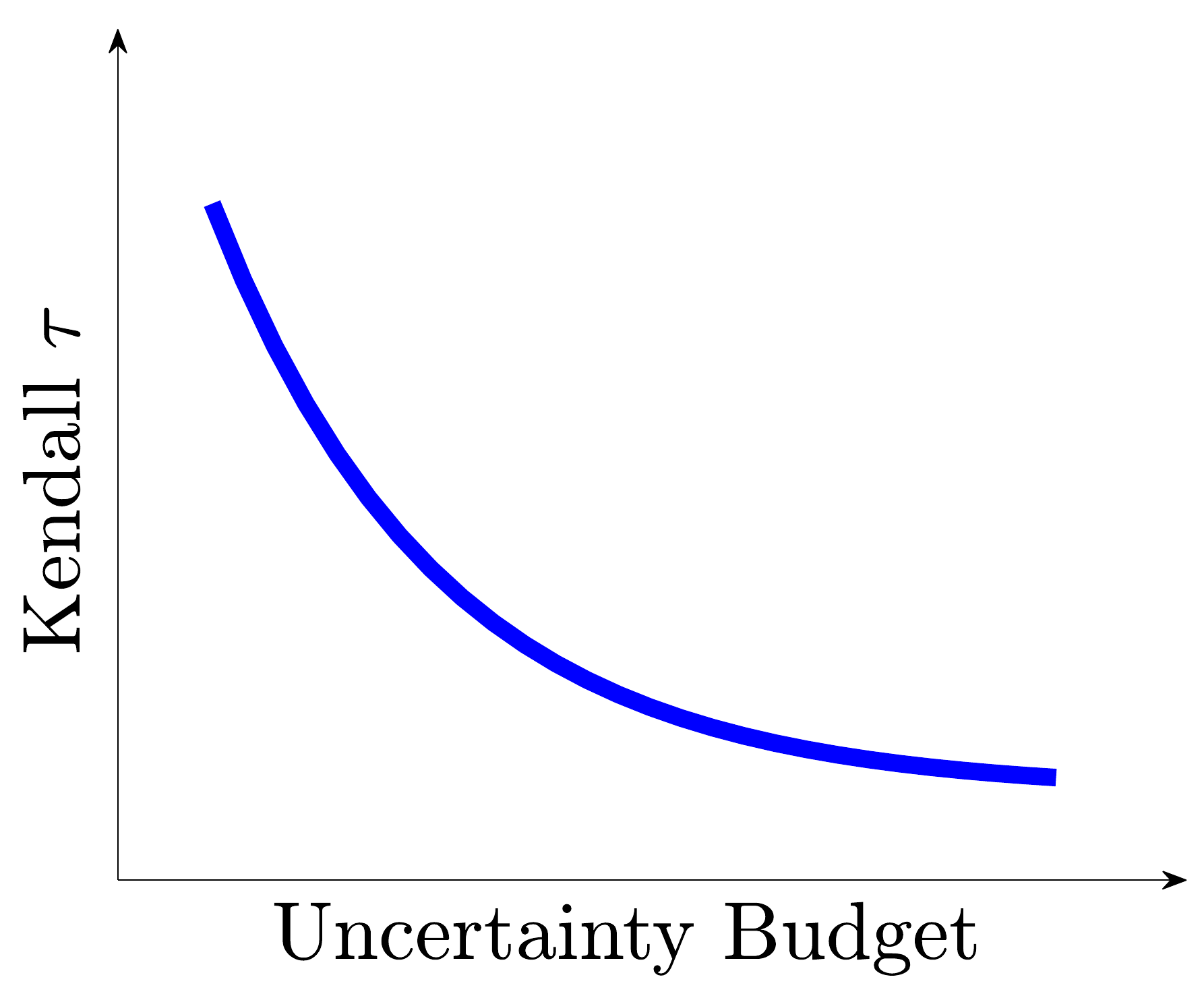}} & 1.0000 & 1.0000 & 1.0000 & 1.0000      \\ 
                             & $10^{-5}$ & \ 1.0000 ($-$)          &                   & 1.0000 & 1.0000 & 1.0000  & 1.0000 \\ 
                             & $10^{-4}$ & \ 1.0000 ($-$)          &                   & 1.0000 & 1.0000 & 1.0000  & 1.0000 \\ 
                             & $10^{-3}$ & -0.6889  ($\downarrow$) &                   & 0.1111 & 0.0000 & 0.0000  & 0.0000 \\ 
                             & $10^{-2}$ & -1.0000  ($\downarrow$) &                   & 0.1000 & 0.0000 & 0.0000  & 0.0000 \\  
                             & $10^{-1}$ & -0.8222                 &                   & 0.1250 & 0.0000 & 0.0000  & 0.0000 \\  
                             & $1$       & -0.9111                 &                   & 0.1111 & 0.0000 & 0.0000  & 0.0000 \\  
                             \midrule 
    \multirow{7}{*}{Dynamic} & $10^{-6}$ & -0.7333                 & \multirow{7}{*}{\includegraphics[height=12mm]{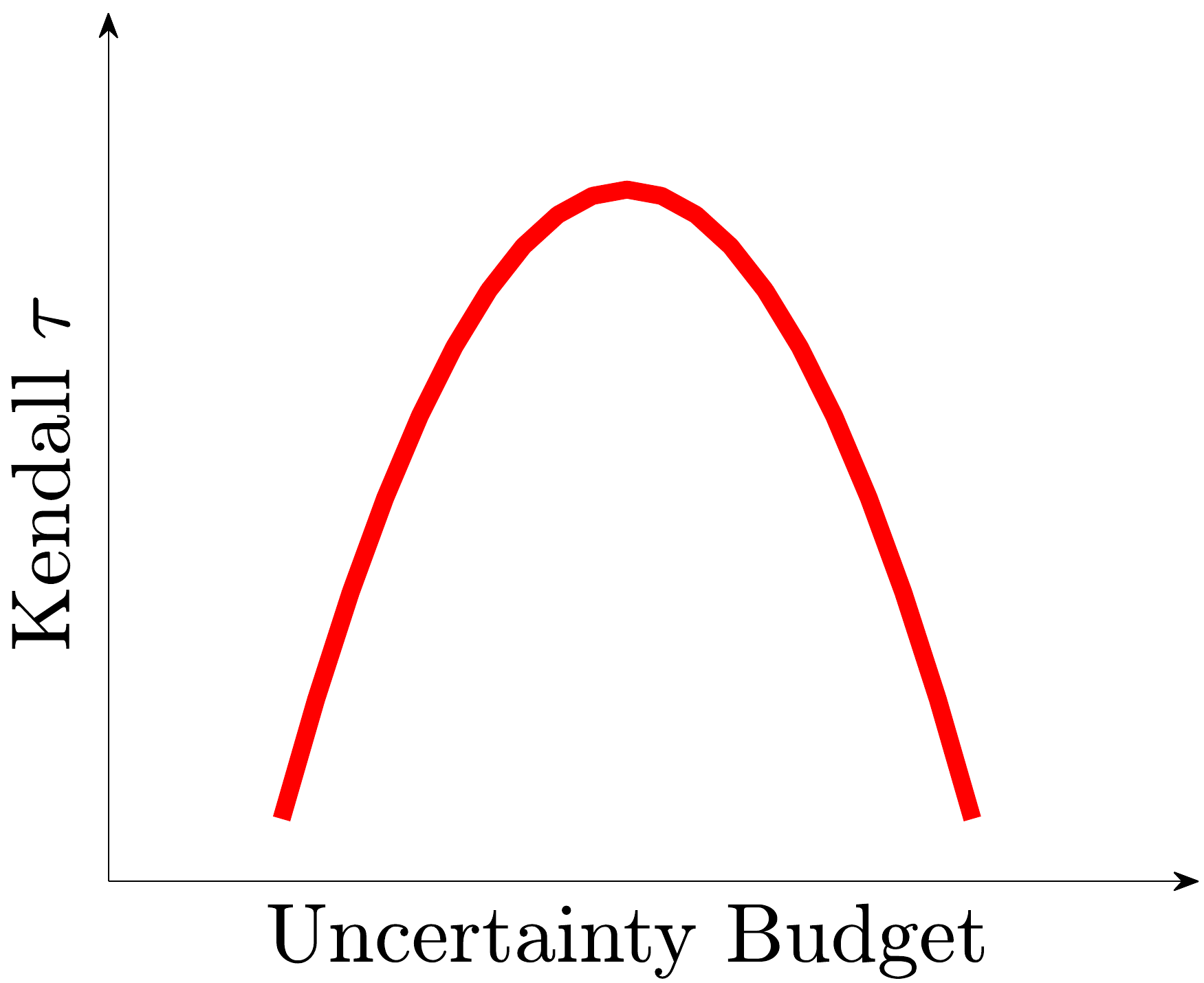}} & 0.1429 & 0.0000 & 0.0000  & 0.0000      \\ 
                             & $10^{-5}$ & -0.7333  ($-$)          &                   & 0.1429 & 0.0000 & 0.0000 & 0.0000 \\ 
                             & $10^{-4}$ & -0.7333  ($-$)          &                   & 0.1429 & 0.0000 & 0.0000 & 0.0000 \\ 
                             & $10^{-3}$ & -0.7333  ($-$)          &                   & 0.1429 & 0.0000 & 0.0000 & 0.0000 \\ 
                             & $10^{-2}$ & \ 0.5111 ($\uparrow$)   &                   & 1.0000 & 0.3333 & 0.3333 & 0.4040 \\  
                             & $10^{-1}$ & \ 0.3778 ($\downarrow$) &                   & 0.5000 & 0.0000 & 0.0000 & 0.0000 \\  
                             & $1$       & -0.4222  ($\downarrow$) &                   & 1.0000 & 0.3333 & 0.3333 & 0.4040 \\  
                             \bottomrule
    \end{tabular}
    \label{table:1}
}
\hspace{2em}
\parbox{0.45\linewidth}
{
    \centering
    \begin{tabular}{@{}cclcc@{}c@{}c@{}c@{}}
    \toprule
    Method                   & Budget    & Kendall-$\tau$          & Tendency (ideal)  & R-Rank\ \ \ & P$@5$\ \ \ & AP$@5$\ \ \ & NDCG$@5$\\ \midrule
    Original                 & -         & \ 1.0000                & -                 & 1.0000 & 1.0000 & 1.0000 & 1.0000  \\ 
    \midrule
    Random                   & 0.05/0.05 & \ 0.9684                & -                 & 1.0000 & 1.0000 & 1.0000 & 1.0000  \\ 
    \midrule
    \multirow{7}{*}{Static}  & $10^{-6}$ & \ 1.0000                & \multirow{7}{*}{\includegraphics[height=12mm]{Trend_1_crop.pdf}} & 1.0000 & 1.0000 & 1.0000 & 1.0000   \\ 
                             & $10^{-5}$ & \ 0.9684 ($\downarrow$) &                   & 1.0000 & 1.0000 & 1.0000 & 1.0000   \\ 
                             & $10^{-4}$ & -0.4737  ($\downarrow$) &                   & 0.0588 & 0.0000 & 0.0000 & 0.0000   \\ 
                             & $10^{-3}$ & -1.0000  ($\downarrow$) &                   & 0.0500 & 0.1000 & 0.0333 & 0.1127   \\ 
                             & $10^{-2}$ & -0.4842                 &                   & 0.0526 & 0.0000 & 0.0000 & 0.0000   \\  
                             & $10^{-1}$ & -0.7474                 &                   & 0.0500 & 0.0000 & 0.0000 & 0.0000   \\
                             & $1$       & -0.7579                 &                   & 0.0500 & 0.0000 & 0.0000 & 0.0000   \\
                             \midrule 
    \multirow{7}{*}{Dynamic} & $10^{-6}$ & -0.7579                 & \multirow{7}{*}{\includegraphics[height=12mm]{Trend_2_crop.pdf}} & 0.0556 & 0.0000     & 0.0000    & 0.0000      \\ 
                             & $10^{-5}$ & -0.7579 ($-$)           &                   & 0.0556 & 0.0000 & 0.0000 & 0.0000   \\ 
                             & $10^{-4}$ & -0.7279 ($-$)           &                   & 0.0556 & 0.0000 & 0.0000 & 0.0000   \\ 
                             & $10^{-3}$ & -0.6842  ($\uparrow$)   &                   & 0.0556 & 0.0000 & 0.0000 & 0.0000   \\
                             & $10^{-2}$ & \ 1.0000 ($\downarrow$) &                   & 1.0000 & 1.0000 & 1.0000 & 1.0000   \\
                             & $10^{-1}$ & \ 0.4526 ($\downarrow$) &                   & 0.2500 & 0.2000 & 0.0400 & 0.1546   \\ 
                             & $1$       & -0.7053  ($\downarrow$) &                   & 1.0000 & 0.2000 & 0.2000 & 0.2738   \\  \bottomrule
    \end{tabular}
    \label{table:2}
}
\bigskip

\parbox{0.45\linewidth}
{
    \centering
    \begin{tabular}{@{}cclc@{}c@{}c@{}c@{}c@{}}
    \toprule
    Method                   & Budget    & Kendall-$\tau$          & Tendency (ideal)  & R-Rank\ \ \ & P$@10$\ \ \ & AP$@10$\ \ \ & NDCG$@10$ \\ \midrule
    Original                 & -         & \ 1.0000                & -                 & 1.0000 & 1.0000 & 1.0000 & 1.0000   \\ \midrule
    Random                   & 0.05/0.05 & \ 0.9396                & -                 & 0.5000 & 0.1000 & 0.0250 & 0.1012   \\ \midrule
    \multirow{7}{*}{Static}  & $10^{-6}$ & \ 0.9886                & \multirow{7}{*}{\includegraphics[height=12mm]{Trend_1_crop.pdf}} & 1.0000 & 0.8000 & 0.6709 & 0.8056   \\ 
                             & $10^{-5}$ & \ 0.6327 ($\downarrow$) &                   & 1.0000 & 0.3000 & 0.2333 & 0.3715   \\ 
                             & $10^{-4}$ & -0.9200  ($\downarrow$) &                   & 0.0227 & 0.0000 & 0.0000 & 0.0000   \\ 
                             & $10^{-3}$ & -1.0000  ($\downarrow$) &                   & 0.0200 & 0.0000 & 0.0000 & 0.0000   \\ 
                             & $10^{-2}$ & -0.6637                 &                   & 0.0294 & 0.0000 & 0.0000 & 0.0000   \\ 
                             & $10^{-1}$ & -0.7224                 &                   & 0.0250 & 0.0000 & 0.0000 & 0.0000   \\ 
                             & $1$       & -0.7741                 &                   & 0.0200 & 0.0000 & 0.0000 & 0.0000   \\  \midrule 
    \multirow{7}{*}{Dynamic} & $10^{-6}$ & -0.7486                 & \multirow{7}{*}{\includegraphics[height=12mm]{Trend_2_crop.pdf}} & 0.0238 & 0.0000 & 0.0000 & 0.0000   \\ 
                             & $10^{-5}$ & -0.7486  ($-$)          &                   & 0.0238 & 0.0000 & 0.0000 & 0.0000   \\ 
                             & $10^{-4}$ & -0.6669  ($\uparrow$)   &                   & 0.0238 & 0.0000 & 0.0000 & 0.0000   \\ 
                             & $10^{-3}$ & \ 0.8824 ($\uparrow$)   &                   & 0.5000 & 0.1000 & 0.0200 & 0.0932   \\ 
                             & $10^{-2}$ & \ 1.0000 ($\uparrow$)   &                   & 1.0000 & 1.0000 & 1.0000 & 1.0000   \\ 
                             & $10^{-1}$ & -0.0580  ($\downarrow$) &                   & 0.1429 & 0.0000 & 0.0000 & 0.0000   \\ 
                             & $1$       & -0.8808  ($\downarrow$) &                   & 0.3333 & 0.0000 & 0.0000 & 0.0000   \\  \bottomrule
    \end{tabular}
    \label{table:3}
}
\hspace{2em}
\parbox{0.45\linewidth}
{
    \centering
    \begin{tabular}{@{}cclc@{}c@{}c@{}c@{}c@{}}
    \toprule
    Method                   & Budget    & Kendall-$\tau$          & Tendency (ideal)  & R-Rank\ \ \ & P$@10$\ \ & AP$@10$\ \ \ & NDCG$@10$ \\ \midrule
    Original                 & -         & \ 0.9996                & -                 & 1.0000 & 1.0000 & 1.0000 & 1.0000   \\ \midrule
    Random                   & 0.05/0.05 & \ 0.9543                & -                 & 0.5000 & 0.2000 & 0.0422 & 0.1688   \\ \midrule
    \multirow{7}{*}{Static}  & $10^{-6}$ & \ 0.9762                & \multirow{7}{*}{\includegraphics[height=12mm]{Trend_1_crop.pdf}} & 1.0000 & 0.2000 & 0.1000    & 0.2380      \\ 
                             & $10^{-5}$ & -0.8242  ($\downarrow$) &                   & 0.0119 & 0.0000 & 0.0000 & 0.0000   \\ 
                             & $10^{-4}$ & -0.9996  ($\downarrow$) &                   & 0.0100 & 0.0000 & 0.0000 & 0.0000   \\ 
                             & $10^{-3}$ & -0.6776                 &                   & 0.0102 & 0.0000 & 0.0000 & 0.0000   \\ 
                             & $10^{-2}$ & -0.6933                 &                   & 0.0133 & 0.0000 & 0.0000 & 0.0000   \\ 
                             & $10^{-1}$ & -0.7459                 &                   & 0.0102 & 0.0000 & 0.0000 & 0.0000   \\ 
                             & $1$       & -0.8307                 &                   & 0.0103 & 0.0000 & 0.0000 & 0.0000   \\  \midrule 
    \multirow{7}{*}{Dynamic} & $10^{-6}$ & -0.7693                 & \multirow{7}{*}{\includegraphics[height=12mm]{Trend_2_crop.pdf}} & 0.0120 & 0.0000 & 0.0000 & 0.0000   \\ 
                             & $10^{-5}$ & -0.7568  ($\uparrow$)   &                   & 0.0120 & 0.0000 & 0.0000 & 0.0000   \\ 
                             & $10^{-4}$ & -0.7095  ($\uparrow$)   &                   & 0.0120 & 0.0000 & 0.0000 & 0.0000   \\ 
                             & $10^{-3}$ & \ 0.9996 ($\uparrow$)   &                   & 1.0000 & 1.0000 & 1.0000 & 1.0000   \\ 
                             & $10^{-2}$ & \ 0.4853 ($\downarrow$) &                   & 0.0333 & 0.0000 & 0.0000 & 0.0000   \\ 
                             & $10^{-1}$ & -0.6402  ($\downarrow$) &                   & 0.0286 & 0.0000 & 0.0000 & 0.0000   \\ 
                             & $1$       & -0.9402  ($\downarrow$) &                   & 0.0222 & 0.0000 & 0.0000 & 0.0000   \\  \bottomrule
    \end{tabular}
    \label{table:4}
}
\end{table*}

\begin{figure}
    \centering
    \begin{subfigure}[b]{\linewidth}
        \centering
        \includegraphics[width=0.75\textwidth]{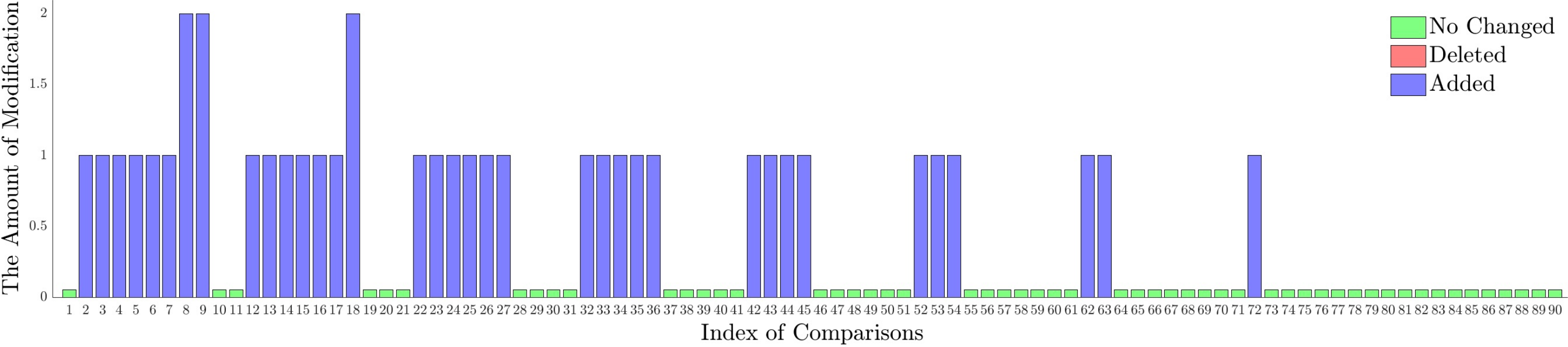}
        \caption{$\alpha=10^{-6}$}
        \label{fig:data:1}
    \end{subfigure}
    \begin{subfigure}[b]{\linewidth}
        \centering
        \includegraphics[width=0.75\textwidth]{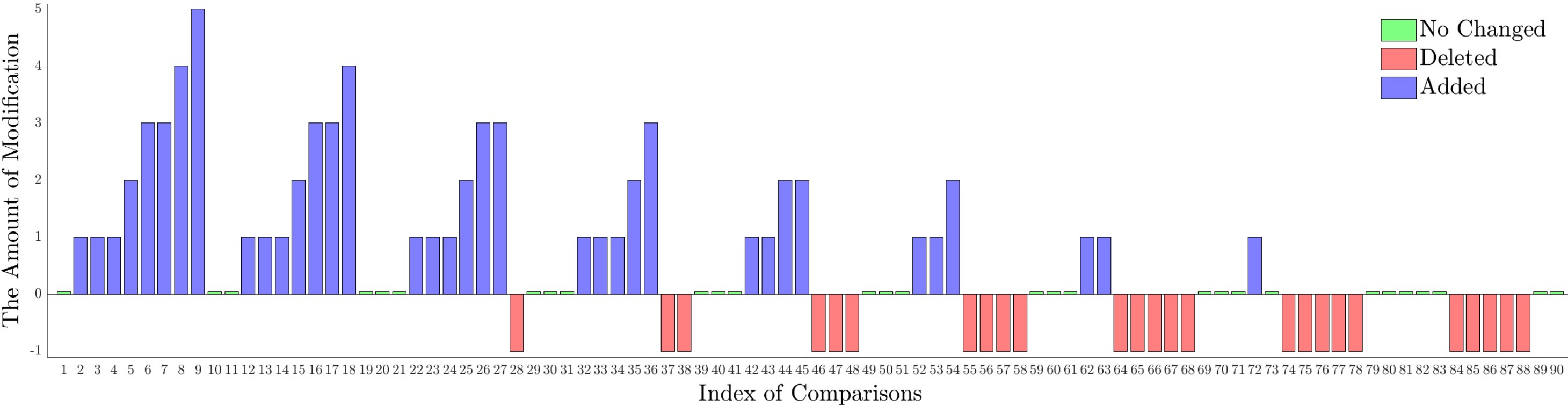}
        \caption{$\alpha=10^{-3}$}
        \label{fig:data:2}
    \end{subfigure}
    \begin{subfigure}[b]{\linewidth}
        \centering
        \includegraphics[width=0.75\textwidth]{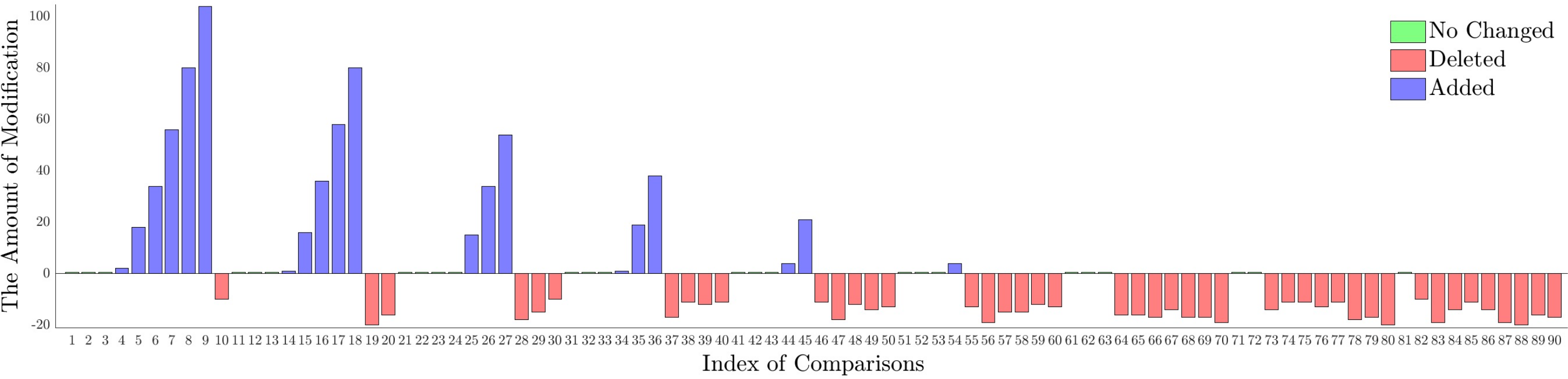}
        \caption{$\alpha=1$}
        \label{fig:data:3}
    \end{subfigure}
    \caption{The amount of changed pairwise comparisons by the poisoning attack with static game. The x-axis is the index of pairwise comparisons and the y-axis is the amount of change. Note that the ranges of y-axis in each sub-figure are different.}
    \label{figure:1}
\end{figure}

\begin{figure}
    \centering
    \includegraphics[width=0.75\linewidth]{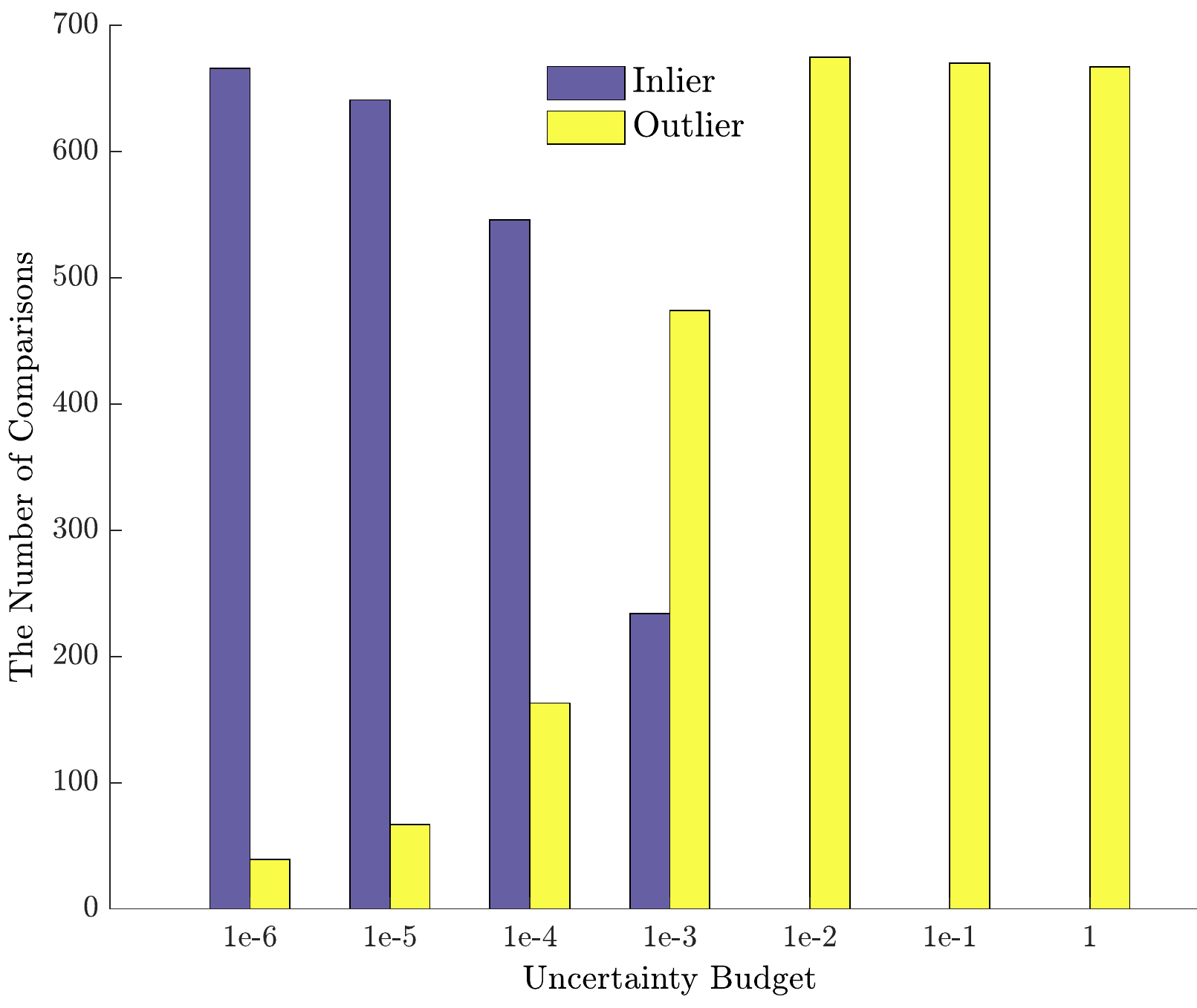}
    \caption{The number of correct pairwise comparisons and comparisons which conflict with the ground-truth ranking in the poisoned training set by `Static' method.}
    \label{figure:2}
\end{figure}
    
\noindent\textbf{Comparative Results. }We display the comparative results of different attack methods in Table \ref{table:1}. There the number of candidates  ranges from $10$ to $100$ ( $n=10, 20, 50, 100$ ). The percentage of noisy comparisons is $\varrho=0$ in the four cases. We let the maximum toxic dosage to be 0 as $\kappa=0$ to verify the effectiveness of the worst-case distribution in the Wasserstein ball with uncertainty budget $\alpha$. We show the attack effect of `Static' and `Dynamic' methods with different budgets. The performance of `Random' are affected by two parameters: the percentage of the new comparisons injected into the original training set, and the percentage of the existed comparisons deleted from the original training set. Here we set these two parameters be $s_1 = s_2 = 0.05$. We obtain the following observations from Table \ref{table:1}. The `Static' method can decrease the Kendall-$\tau$ when the uncertainty budget $\alpha$ increases. Looking back on the \textbf{Algorithm \ref{alg:main}}, the uncertainty budget $\alpha$ is the weight of the second term in \eqref{eq:final_theta} and the two parts of \eqref{eq:final_theta} have the same monotonic respect to $\boldsymbol{\theta}$. With the increasing of $\alpha$, the impact of the second term \eqref{eq:final_reg} to the solution \eqref{eq:final_theta} becomes gradually. The solution of \eqref{eq:final_reg} means that the algorithm will adopt all possible pairwise comparisons with same number of voting to aggregate the final ordered list. There is no doubt that this case would be far away from the ground-truth ranking. If $\alpha$ approaches $\infty$, we would obtain this confusing solution. This explains the behaviors of the `Static' methods when the Kendall-$\tau$ is larger than $0$. In Figure \ref{figure:1}, we see that the `Static' method does two things to perturb the training set: adding pairwise comparisons which conflict with the ground-truth ranking and removing the pairwise comparisons which is consistent with the ground-truth ranking. The total amount of change enlarge when the uncertainty budget $\alpha$ increase. If the Kendall-$\tau$ is smaller than $0$, it means that the poisoned training dataset would support an opposite ranking list. In Figure \ref{figure:2}, each group corresponds to a poisoned data set by `Static' method with a certain uncertainty budget. When the Kendall-$\tau$ is smaller than $0$ ($\alpha\geq10^{-3}$), we observe that the number of comparisons which conflict with the ground-truth ranking is larger that the number of comparisons which is consistent with the ground-truth ranking. Such training data could generate an arbitrarily ordered list. If it happens, the Kendall-$\tau$ could not monotonically decrease when we increase the uncertainty budget continuously. Moreover, the uncertainty budget $\alpha$ plays a totally different role in the `Dynamic' method. The existing work \cite{DBLP:conf/nips/NamkoongD17,DBLP:journals/jmlr/DuchiN19} reveal that such kind of min-max problem is a new type of regularization. This regularization also carries out the `bias-variance' trade-off like the classical approaches like Tikhonov regularization. In this case, the uncertainty budget $\alpha$ can be explained as a regularization coefficient. The Kendall-$\tau$ of `Dynamic' method presents a `U'-type curve in our experiments. 

\begin{figure}
    \centering
    \includegraphics[width=0.8\linewidth]{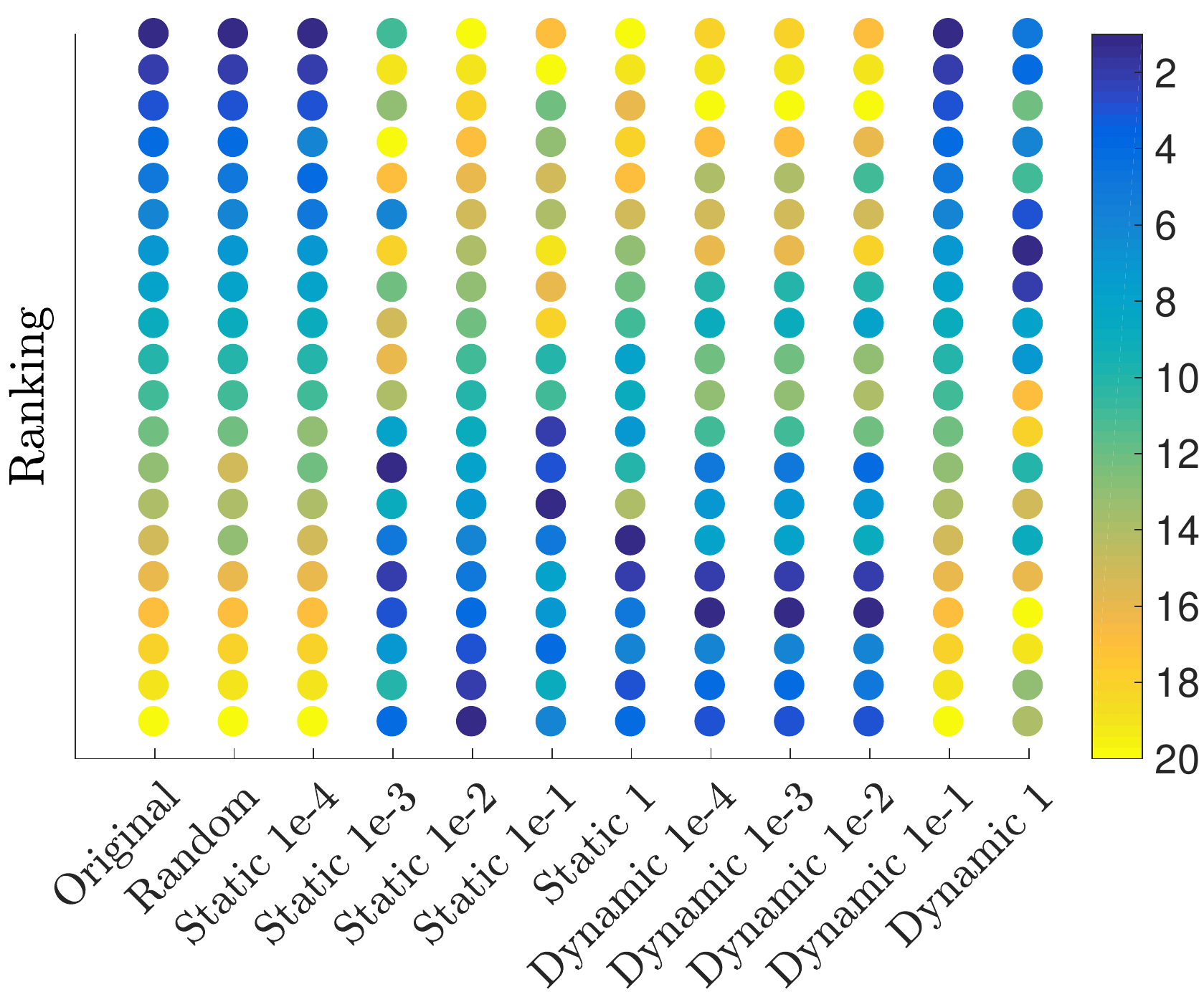}
    \caption{The ranking generated from the original data (Original), random attack data (Random), static poisoning attack data (Static) and dynamic poisoning attack data (Dynamic).}
    \label{figure:3}
\end{figure}

\vspace{0.25cm}
\noindent\textbf{Visualization. }We visualize the ranking list in Figure \ref{figure:3}. The visualization shows the same phenomenons as the numeric results in Table \ref{table:simulation}. As the target ranking aggregation algorithm does not emphasize the top-K results and the adversary has no prior knowledge of the ranking results, the untrustworthy results of `Static' method only depend on the original data and the uncertainty budget. So the proposed method is the non-target attack for pairwise ranking algorithm. Manipulating the ranking list with specific goals, \textit{a.k.a} the target attack, is the future work.

\begin{figure}
    \centering
    \includegraphics[width=\linewidth]{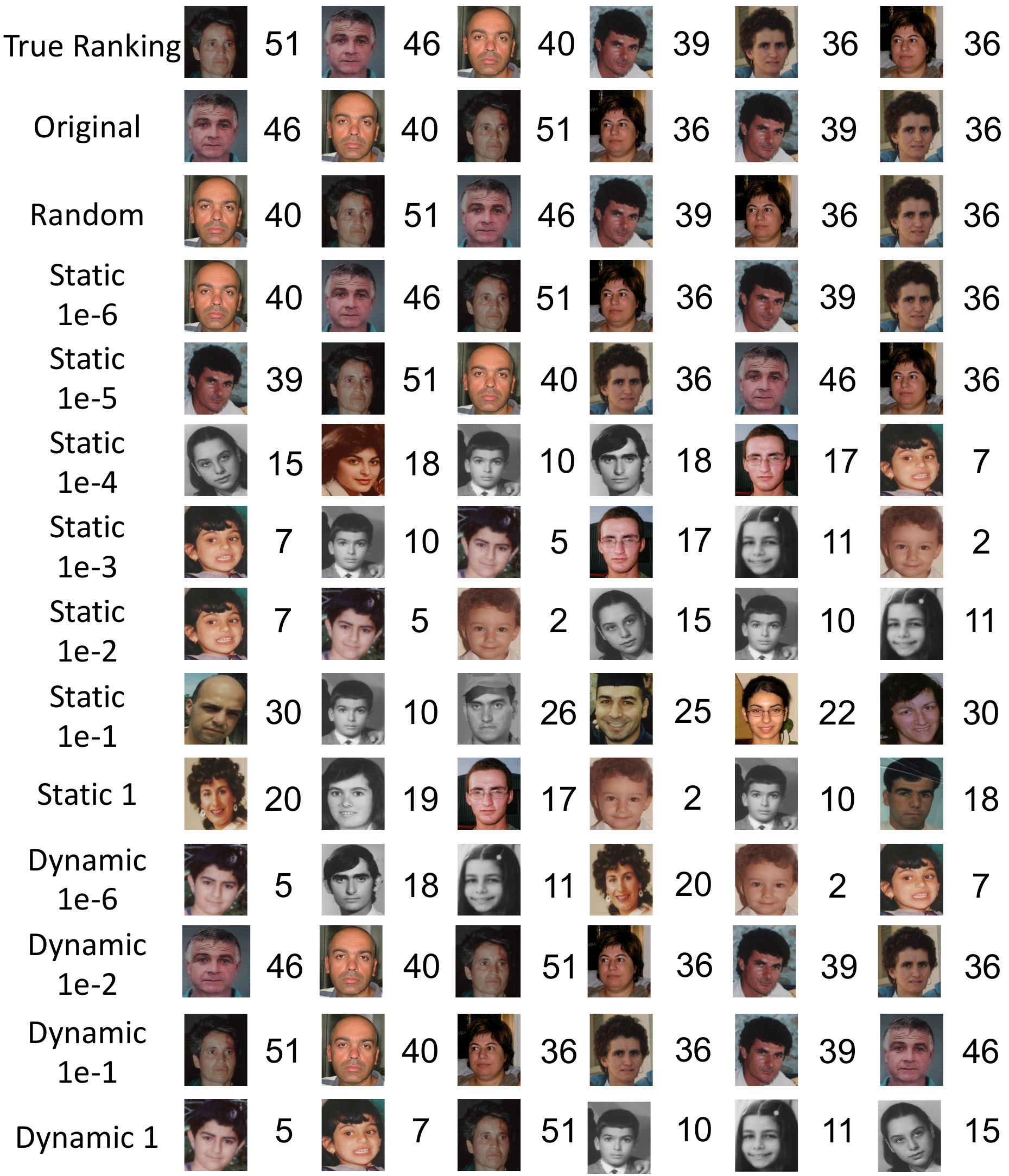}
    \caption{The ranking generated from the original data (Original), random perturbation data (Random), poisoned data (Static and Dynamic) on Human Age dataset. When the Kendall-$\tau$ is smaller than $0$ ($\alpha\geq10^{-4}$), we observe that the aggregated results would put the younger people at the top of the lists. Moreover, the same phenomenons in the simulation are still observed. The training data with more than $50\%$ outliers could generate an arbitrarily ordered list. If it happens, the Kendall-$\tau$ could not monotonically decrease when we increase the uncertainty budget continuously for the static attack strategies.}
    \label{figure:4}
\end{figure}

\subsection{Human Age}

\textbf{Description. }$30$ images from human age dataset FGNET are annotated by a group of volunteer users on ChinaCrowds platform. The ground-truth age ranking is known to us. The annotator is presented with two images and given a binary choice of which one is older. Totally, we obtain $8,017$ pairwise comparisons from $94$ annotators.

\vspace{0.25cm}
\noindent\textbf{Comparative Results. } Notice that the real-world data has a high percentage of outliers (about $20\%$ comparisons conflict with the correct age ranking). We observe similar phenomenons as the simulation experiments. When the uncertainty budget increase, the `Static' method would inject more comparisons which conflict with the true age ranking and delete the original comparisons which indicate the true ordered list. Once the `wrong' samples overwhelm the `correct' samples, the ranking aggregation algorithm would like to generate a reversed list. As there are only the `wrong' samples in the toxic training set by `Static' method, the final result could be arbitrary.

\begin{table}[]
\tiny
\centering
\caption{Comparative results of different attack methods on human age data.}
\begin{tabular}{@{}ccl@{}c@{}c@{}c@{}c@{}c@{}}
\toprule
Method                   & Budget    & Kendall-$\tau$          & Tendency          & R-Rank\ \ \ & P$@10$\ \ \ & AP$@10$\ \ \ & NDCG$@10$\ \ \\ \midrule
Original                 & -         & \ 0.6872                & -                 & 0.3333 & 0.0000 & 0.0000  & 0.0000 \\ \midrule
Random                   & 0.05/0.05 & \ 0.7425                & -                 & 0.3333 & 0.0000 & 0.0000  & 0.0000 \\ \midrule
\multirow{7}{*}{Static}  & $10^{-6}$ & \ 0.7149                & \multirow{7}{*}{\includegraphics[height=17mm]{Trend_1_crop.pdf}} & 0.3333 & 0.1000 & 0.0500 & 0.1308     \\ 
                         & $10^{-5}$ & \ 0.7793                &                   & 0.5000 & 0.3000 & 0.1095  & 0.2842 \\ 
                         & $10^{-4}$ & -0.3655                 &                   & 0.0500 & 0.0000 & 0.0000  & 0.0000 \\ 
                         & $10^{-3}$ & -0.5402                 &                   & 0.0345 & 0.0000 & 0.0000  & 0.0000 \\ 
                         & $10^{-2}$ & -0.6552                 &                   & 0.0345 & 0.0000 & 0.0000  & 0.0000 \\  
                         & $10^{-1}$ & -0.1724                 &                   & 0.0385 & 0.0000 & 0.0000  & 0.0000 \\  
                         & $1$       & -0.4851                 &                   & 0.0500 & 0.0000 & 0.0000  & 0.0000 \\  
                         \midrule 
\multirow{7}{*}{Dynamic} & $10^{-6}$ & -0.5494                 & \multirow{7}{*}{\includegraphics[height=17mm]{Trend_2_crop.pdf}} & 0.0345 & 0.0000 & 0.0000  & 0.0000    \\ 
                         & $10^{-5}$ & -0.5494                 &                   & 0.0345 & 0.0000 & 0.0000 & 0.0000 \\ 
                         & $10^{-4}$ & -0.5494                 &                   & 0.0345 & 0.0000 & 0.0000 & 0.0000 \\ 
                         & $10^{-3}$ & -0.5448                 &                   & 0.0345 & 0.0000 & 0.0000 & 0.0000 \\ 
                         & $10^{-2}$ & \ 0.6782                &                   & 0.3333 & 0.0000 & 0.0000 & 0.0000 \\  
                         & $10^{-1}$ & -0.2000                 &                   & 1.0000 & 0.1000 & 0.1000 & 0.1651 \\  
                         & $1$       & -0.7517                 &                   & 0.3333 & 0.0000 & 0.0000 & 0.0000 \\  
                         \bottomrule
\end{tabular}
\label{table:5}
\end{table}

\subsection{Dublin Election}

\textbf{Description. }The Dublin election data set\footnote{\url{http://www.preflib.org/data/election/irish/}} contains a complete record of votes for elections held in county Meath, Dublin, Ireland on 2002. This set contains $64,081$ votes over $14$ candidates. These votes could be a complete or partial list over the candidate set. The ground-truth ranking of $14$ candidates are based on their obtained first preference votes\footnote{\url{https://electionsireland.org/result.cfm?election=2002&cons=178&sort=first}}. The five candidates who receive the most first preference votes will be the winner of the election. We are interested in the top-$5$ performance of the pairwise rank aggregation method. Then these votes are converted into the pairwise comparisons. The total number of the comparisons is $652,817$. 

\vspace{0.25cm}
\noindent\textbf{Comparative Results. }In this experiment, we evaluate the ability of poisoning attack in election. The election result is not obtained by pairwise ranking aggregation. However, the ordered list aggregated from induced comparisons still shows positive correlation with the actual election result. Different from the manipulation or strategic voting setting in election, the adversary could control the whole votes but with some constraints. As a consequence, the poisoning attack could break the barrier of computational complexity \cite{DBLP:journals/amai/Walsh11,DBLP:conf/atal/VaishM0B16}. The proposed method focuses on the `non-target' attack on pairwise ranking aggregation. The `Static' method could perturb the ranking list generated by the original algorithm with a sufficient uncertainty budget. But the adversary is not able to manipulate the order with her/his preference as she/he can not decide the winner of election. We call the problem as the `target' attack, where the adversary manipulates the order with her/his preference. Our future work will study the `target' poisoning attack on pairwise ranking. Moreover, the `Dynamic' method does not completely destroy the election result. It indicates that the inaccurate supervision would mislead the adversary and the corresponding Nash equilibrium could show partiality for the ranking aggregation algorithm. 

\begin{table}[]
\tiny
\centering
\caption{Comparative results of different attack methods on Dublin election data.}
\begin{tabular}{@{}ccl@{}c@{}c@{}c@{}c@{}c@{}}
\toprule
Method                   & Budget    & Kendall-$\tau$          & Tendency          & R-Rank\ \ \ & P$@5$\ \ \ & AP$@5$\ \ \ & NDCG$@5$\ \ \\ \midrule
Original                 & -         & \ 0.4725                & -                 & 0.0769 & 0.4000 & 0.2333  & 0.4038 \\ \midrule
Random                   & 0.05/0.05 & \ 0.4736                & -                 & 0.0769 & 0.4000 & 0.2333  & 0.4038 \\ \midrule
\multirow{7}{*}{Static}  & $10^{-6}$ & \ 0.4725                & \multirow{7}{*}{\includegraphics[height=17mm]{Trend_1_crop.pdf}} & 0.0769 & 0.4000 & 0.2333 & 0.4038     \\ 
                         & $10^{-5}$ & \ 0.4725                &                   & 0.0769 & 0.4000 & 0.2333  & 0.4038 \\ 
                         & $10^{-4}$ & \ 0.5824                &                   & 0.0769 & 0.0000 & 0.0000  & 0.0000 \\ 
                         & $10^{-3}$ & -0.3846                 &                   & 0.1250 & 0.0000 & 0.0000  & 0.0000 \\ 
                         & $10^{-2}$ & -0.4725                 &                   & 0.1250 & 0.0000 & 0.0000  & 0.0000 \\  
                         & $10^{-1}$ & -0.4725                 &                   & 0.1250 & 0.0000 & 0.0000  & 0.0000 \\  
                         & $1$       & -0.0330                 &                   & 0.1250 & 0.0000 & 0.0000  & 0.0000 \\  
                         \midrule 
\multirow{7}{*}{Dynamic} & $10^{-6}$ & \ 0.4286                & \multirow{7}{*}{\includegraphics[height=17mm]{Trend_2_crop.pdf}} & 0.0769 & 0.0000 & 0.0000  & 0.0000    \\ 
                         & $10^{-5}$ & \ 0.5385                &                   & 0.0769 & 0.0000 & 0.0000 & 0.0000 \\ 
                         & $10^{-4}$ & \ 0.5385                &                   & 0.0769 & 0.0000 & 0.0000 & 0.0000 \\ 
                         & $10^{-3}$ & \ 0.4725                &                   & 0.0769 & 0.4000 & 0.2333 & 0.4038 \\ 
                         & $10^{-2}$ & \ 0.5385                &                   & 0.0769 & 0.6000 & 0.3533 & 0.5584 \\  
                         & $10^{-1}$ & \ 0.5385                 &                   & 1.0000 & 0.2000 & 0.2000 & 0.2738 \\  
                         & $1$       & \ 0.1648                 &                   & 0.3333 & 0.0000 & 0.0000 & 0.0000 \\  
                         \bottomrule
\end{tabular}
\label{table:6}
\end{table}

\begin{table}[]
\tiny
\centering
\caption{Comparative results of different attack methods on Sushi election data.}
\begin{tabular}{@{}ccll@{}c@{}c@{}c@{}c@{}}
\toprule
Method                   & Budget    & Kendall-$\tau$          & Tendency          & R-Rank &\ \ P$@3$\ \ &\ \ AP$@3$\ \ & NDCG$@3$\ \ \\ \midrule
Original                 & -         & \ 1.0000                & -                 & 1.0000 & 1.0000 & 1.0000  & 1.0000  \\ \midrule
Random                   & 0.05/0.05 & \ 1.0000                & -                 & 1.0000 & 1.0000 & 1.0000  & 1.0000  \\ \midrule
\multirow{7}{*}{Static}  & 1e-6      & \ 1.0000                & \multirow{7}{*}{\includegraphics[height=15mm]{Trend_1_crop.pdf}} & 1.0000 & 1.0000 & 1.0000 & 1.0000     \\ 
                         & 1e-5      & \ 1.0000                &                   & 1.0000 & 1.0000 & 1.0000  & 1.0000 \\ 
                         & 1e-4      & \ 1.0000                &                   & 1.0000 & 1.0000 & 1.0000  & 1.0000 \\ 
                         & 1e-3      & -0.9556                 &                   & 0.2500 & 0.0000 & 0.0000  & 0.0000 \\ 
                         & 1e-2      & -1.0000                 &                   & 0.2500 & 0.0000 & 0.0000  & 0.0000 \\  
                         & 1e-1      & -1.0000                 &                   & 0.2500 & 0.0000 & 0.0000  & 0.0000 \\  
                         & 1         & -0.7333                 &                   & 0.2500 & 0.0000 & 0.0000  & 0.0000 \\  
                         \midrule 
\multirow{7}{*}{Dynamic} & 1e-6      & \ 0.4222                & \multirow{7}{*}{\includegraphics[height=15mm]{Trend_2_crop.pdf}} & 0.1000 & 0.0000 & 0.0000  & 0.0000    \\ 
                         & 1e-5      & \ 0.4222                &                   & 0.1000 & 0.0000 & 0.0000 & 0.0000 \\ 
                         & 1e-4      & \ 0.4667                &                   & 0.1000 & 0.0000 & 0.0000 & 0.0000 \\ 
                         & 1e-3      & \ 0.7333                &                   & 0.1250 & 0.3333 & 0.1667 & 0.3202 \\ 
                         & 1e-2      & \ 1.0000                &                   & 1.0000 & 1.0000 & 1.0000 & 1.0000 \\  
                         & 1e-1      & \ 0.7778                &                   & 0.2500 & 0.0000 & 0.0000 & 0.0000 \\  
                         & 1         & \ 0.4222                &                   & 0.2500 & 0.0000 & 0.0000 & 0.0000 \\  
                         \bottomrule
\end{tabular}
\label{table:7}
\end{table}

\subsection{Sushi Preference}
\textbf{Description. }This dataset contains the results of a series of surveys which involves 5000 individuals for their preferences about various kinds of sushi. The original survey provides 10 complete strict rank orders of 10 different kinds of sushi as 1) ebi (shrimp), 2) anago (sea eel), 3) maguro (tuna), 4) ika (squid), 5) uni (sea urchin), 6) sake (salmon roe), 7) tamago (egg), 8) toro (fatty tuna), 9) tekka-maki (tuna roll), and 10) kappa-maki (cucumber roll). The complete strict rank orders are converted into the pairwise graph by \cite{MaWa13a}. We adopt the whole $221,670$ comparisons and the Hodgerank\cite{Jiang2011} method to aggregate a ranking list as the ground-truth. Then $20$ percent of pairwise comparisons are chosen to consist of the observation set. The different attack approaches can manipulate the subset of data and induce the pairwise ranking algorithm to generate a different order list.

\vspace{0.25cm}
\noindent\textbf{Comparative Results. } This experiment is a classic setting in recommendation and computational advertisement. With the selected subset, the ranking aggregation method can produce a same ranking list as adopting with the whole preference data. The random attack would not change this list in this experiment. In addition, the `Dynamic' method is trapped with the inaccurate supervision and only shows a moderate destructive effect. The `Static' method could generate a promise perturbation to mislead the ranking aggregation method as the Kendall-$\tau$ would be $-1$.

\subsection{Computational Complexity Analysis}
The computational complexity of the dynamic strategy depends on the number of turns of \eqref{opt:chi_2}. Given $n$ candidates, the complexity of the ranker is $\mathcal{O}(n^6)$ for solving a least square problem and the complexity of the  adversary is $\mathcal{O}(n^2\log(n^2)+\log\frac{1}{\epsilon}\cdot\log(n^2))$ where $\epsilon$ is the solution accuracy, $n^2\log(n^2)$ is for sorting and the last part corresponds to the projection onto the $\ell_2$ ball. The computational complexity of the static strategy depends on the subroutines of Line 2 and Line 4 in Algorithm 1. We solve the subroutine of Line 2 by gradient descent and evaluating the gradient needs $\mathcal{O}(n^4)$ each time. The complexity of Line 4 is $\mathcal{O}(n^3+n^2\log(n^2)+n^2)$ where $n^3$ is for the closed form, $n^2\log(n^2)$ is for the sorting and $n^2$ for the projection onto the simplex. We also display the computational complexity comparisons on the synthetic and the real-world datasets in Table \ref{table:8} and \ref{table:9}. The results are mean of 100 trials with different pairwise comparisons or initialization. All computation is done using MATLAB$^\circledR$ R2016b, on a Laptop PC with MacOS$^\circledR$ Big Sur, with 3.1GHz Intel$^{\circledR}$ Core i7 CPU, and 16GB 2133MHz DDR3 memory. 

\begin{table}[]
\centering
\caption{Computational complexity (ms) comparisons on the synthetic dataset. The results are the mean of 100 trials with different pairwise comparisons.}
\label{table:8}
\begin{tabular}{cccccc}
\toprule
\multirow{2}{*}{Method}   & \multirow{2}{*}{Budget} & \multicolumn{4}{c}{No. of Candidates}      \\ \cline{3-6} 
                          &                         & $10$     & $20$     & $50$     & $100$     \\ \hline
\multirow{7}{*}{Static}   & $1e^{-6}$               & $0.0736$ & $0.2123$ & $9.7374$ & $198.7755$\\
                          & $1e^{-5}$               & $0.0727$ & $0.2274$ & $9.6190$ & $200.4348$\\
                          & $1e^{-4}$               & $0.0692$ & $0.1976$ & $9.6027$ & $200.7480$\\
                          & $1e^{-3}$               & $0.0715$ & $0.2133$ & $9.5939$ & $197.5005$\\
                          & $1e^{-2}$               & $0.0712$ & $0.2124$ & $9.3595$ & $196.2806$\\
                          & $1e^{-1}$               & $0.0662$ & $0.2026$ & $9.7653$ & $197.9314$\\
                          & 1                       & $0.0680$ & $0.2410$ & $9.9514$ & $197.4316$\\ \hline
\multirow{7}{*}{Dynamic}  & $1e^{-6}$               & $0.0279$ & $0.0764$ & $2.7743$ & $55.9370$ \\
                          & $1e^{-5}$               & $0.0238$ & $0.0726$ & $2.7711$ & $54.5632$ \\
                          & $1e^{-4}$               & $0.0258$ & $0.0796$ & $2.7001$ & $55.5834$ \\
                          & $1e^{-3}$               & $0.0351$ & $0.0758$ & $2.7028$ & $57.0046$ \\
                          & $1e^{-2}$               & $0.0243$ & $0.0749$ & $2.7095$ & $56.0811$ \\
                          & $1e^{-1}$               & $0.0261$ & $0.0792$ & $2.7665$ & $55.8359$ \\
                          & $1$                     & $0.0252$ & $0.0730$ & $2.7860$ & $55.9948$ \\ \bottomrule
\end{tabular}
\end{table}

\begin{table}[]
\centering
\caption{Computational complexity (ms) comparisons on the real-world datasets. The results are the mean of 100 trials with different initialization.}
\label{table:9}
\begin{tabular}{ccccc}
\toprule
\multirow{2}{*}{Method}   & \multirow{2}{*}{Budget} & \multicolumn{3}{c}{Dataset}  \\ \cline{3-5} 
                          &                         & Age     & Dublin  & Sushi    \\ \hline
\multirow{7}{*}{Static}   & $1e^{-6}$               & $1.2012$& $0.1018$& $0.0548$ \\
                          & $1e^{-5}$               & $1.1174$& $0.0798$& $0.0645$ \\
                          & $1e^{-4}$               & $1.6814$& $0.0862$& $0.0483$ \\
                          & $1e^{-3}$               & $1.3130$& $0.1256$& $0.0455$ \\
                          & $1e^{-2}$               & $1.1457$& $0.0809$& $0.0535$ \\
                          & $1e^{-1}$               & $0.9435$& $0.0779$& $0.0473$ \\
                          & 1                       & $1.0184$& $0.0811$& $0.0403$ \\ \hline
\multirow{7}{*}{Dynamic}  & $1e^{-6}$               & $0.2664$& $0.0356$& $0.0205$ \\
                          & $1e^{-5}$               & $0.2424$& $0.0244$& $0.0240$ \\
                          & $1e^{-4}$               & $0.3147$& $0.0277$& $0.0190$ \\
                          & $1e^{-3}$               & $0.2779$& $0.0369$& $0.0180$ \\
                          & $1e^{-2}$               & $0.2461$& $0.0253$& $0.0207$ \\
                          & $1e^{-1}$               & $0.2235$& $0.0247$& $0.0182$ \\
                          & $1$                     & $0.2382$& $0.0245$& $0.0154$ \\ \bottomrule
\end{tabular}
\end{table}

\vspace{0.5cm}

\section{Conclusion}
We initiate the first study of data poisoning attacks in the context of pairwise ranking. We formulate the attack problem as a robust game between two players, the ranker and the adversary. The attacker’s strategies are modeled as the distributionally robust optimization problems and some theoretical results are established, including the existence of distributionally robust Nash equilibrium and the generalization bounds. Our empirical studies show that our attack strategies significantly break the performance of pairwise ranking in the sense that the correlation between the true ranking list and the aggregated result with toxic data can be decreased dramatically. 

There are many avenues for further investigation – such as, providing the finite-sample and asymptotic results characterizing the theoretical performance of the estimator with adversarial learning, extending our attacks to more pairwise ranking algorithms such as spectral ranking, and trying to attack the ranking algorithms with defense paradigm. We believe that a very interesting open question is to expand our understanding to better understand the role and capabilities of adversaries in pairwise ranking.


%







\vspace{0.75cm}

\newpage
\bibliographystyle{plain}
\bibliography{sample}

{
    \newpage
    \onecolumn
    \appendices
    
    \section{Proof of Theorem \ref{thm:robust_nash_equilibrium}.}
    \label{app:theorem_1}
    \begin{property}
        \label{pro:equilibrium}
        Let the pay-off function $f_r,\ r=1,2$ be the weighted sum-of-squared loss $\ell$ \eqref{eq:weight_l2_loss} in \eqref{opt:DRG:1}. If the uncertainty set is $\boldsymbol{\mathfrak{X}}^{\alpha}(\mathbb{P}_N)$ or $\boldsymbol{\mathfrak{W}}_p^{\alpha}(\mathbb{P}_N)$, we have
        \begin{enumerate}
            \item $f_r$ is a continuous function, and for any fixed $\{\boldsymbol{x}_{-r},\ \boldsymbol{\xi}\}$, $f_r(\boldsymbol{x}_r,\ \boldsymbol{x}_{-r},\ \boldsymbol{\xi})$ is convex over $\mathcal{X}_r$.
            \vspace{0.15cm}
            \item $\mathcal{X} = \mathcal{X}_1\times \mathcal{X}_2$ is a compact set. 
            \vspace{0.15cm}
            \item $\mathbb{E}_{\boldsymbol{\xi}\sim\mathbb{Q}}[f_r(\boldsymbol{x}_r,\ \boldsymbol{x}_{-r},\ \boldsymbol{\xi})]$ is finite-valued, $\forall\ \boldsymbol{x}\in\mathcal{X}$, $\mathbb{Q}\in\boldsymbol{\mathfrak{U}}$.
            \vspace{0.15cm}
            \item $\boldsymbol{\mathfrak{U}}$ is a weakly compact set.
        \end{enumerate}
    \end{property}

    \begin{proposition}
        \label{prop:reformulation}
        Let $\boldsymbol{x}=\{\boldsymbol{x}_1,\ \boldsymbol{x}_2\}$, $\boldsymbol{v}=\{\boldsymbol{v}_1,\ \boldsymbol{v}_2\}$, we define $\phi:\mathcal{X}\times\mathcal{X}\rightarrow\mathbb{R}_+$ as 
        \begin{align}
            \label{eq:single_level}
            \phi(\boldsymbol{v},\ \boldsymbol{x})\ =\underset{\mathbb{Q}\ \in\ \boldsymbol{\mathfrak{U}}}{\textbf{\textit{sup}}}\ \mathbb{E}_{\boldsymbol{\xi}\sim\mathbb{Q}}\ \Big[\ f_1(\boldsymbol{v}_1,\ \boldsymbol{x}_{2},\ \boldsymbol{\xi})\ \Big]\numberthis+\underset{\mathbb{Q}\ \in\ \boldsymbol{\mathfrak{U}}}{\textbf{\textit{sup}}}\ \mathbb{E}_{\boldsymbol{\xi}\sim\mathbb{Q}}\ \Big[\ f_2(\boldsymbol{x}_1,\ \boldsymbol{v}_{2},\ \boldsymbol{\xi})\ \Big]\nonumber
        \end{align}
        With Property \ref{pro:equilibrium}, $ \boldsymbol{x}^*=\{\boldsymbol{x}^*_1,\ \boldsymbol{x}^*_2\}$ is a distributional robust Nash equilibrium of \eqref{eq:drne_equilibrium} if and only if
        \begin{equation}
            \{\boldsymbol{x}^*_1,\ \boldsymbol{x}^*_2\}\ \in\ \underset{\boldsymbol{v}\ \in\ \mathcal{X}}{\textbf{\textit{arg\ min}}}\ \phi(\boldsymbol{v},\ \boldsymbol{x}^*).
        \end{equation}
    \end{proposition}
    \begin{proof}
        The reformulation $\phi$ is well known for deterministic Nash equilibrium, see for example \cite{10.2307/1911749}. The ``if'' part follows from the fact that if $\{\boldsymbol{x}^*_1,\ \boldsymbol{x}^*_2\}$ is not an equilibrium of \eqref{opt:DRG:1}, there exists some $\boldsymbol{\bar{x}}_r$, $r=1, 2$, such that
        \begin{align}
            \underset{\mathbb{Q}\ \in\ \boldsymbol{\mathfrak{U}}}{\textbf{\textit{sup}}}\ \mathbb{E}_{\boldsymbol{\xi}\sim\mathbb{Q}}\Big[\ f_r(\boldsymbol{\bar{x}}_r,\ \boldsymbol{x}^*_{-r},\ \boldsymbol{\xi})\ \Big]<\underset{\mathbb{Q}\ \in\ \boldsymbol{\mathfrak{U}}}{\textbf{\textit{sup}}}\ \mathbb{E}_{\boldsymbol{\xi}\sim\mathbb{Q}}\Big[\ f_r(\boldsymbol{x}^*_r,\ \boldsymbol{x}^*_{-r},\ \boldsymbol{\xi})\ \Big]\nonumber
        \end{align}
        Let $\boldsymbol{\bar{x}}=\{\boldsymbol{\bar{x}}_r,\ \boldsymbol{x}^*_{-r}\}$, we have $\phi(\boldsymbol{\bar{x}},\ \boldsymbol{x}^*)<\phi( \boldsymbol{x}^*,\ \boldsymbol{x}^*)$. This is a contradiction. 

        The ``only if'' part is obvious as     
        \begin{align}
            \underset{\mathbb{Q}\ \in\ \boldsymbol{\mathfrak{U}}}{\textbf{\textit{sup}}}\ \mathbb{E}_{\boldsymbol{\xi}\sim\mathbb{Q}}\Big[\ f_r(\boldsymbol{x}_r,\ \boldsymbol{x}^*_{-r},\ \boldsymbol{\xi})\ \Big]>\underset{\mathbb{Q}\ \in\ \boldsymbol{\mathfrak{U}}}{\textbf{\textit{sup}}}\ \mathbb{E}_{\boldsymbol{\xi}\sim\mathbb{Q}}\Big[\ f_r(\boldsymbol{x}^*_r,\ \boldsymbol{x}^*_{-r},\ \boldsymbol{\xi})\ \Big]\nonumber
        \end{align}
        Summing up each $r$ on both sides, the inequality shows that $\{\boldsymbol{x}^*_1,\ \boldsymbol{x}^*_2\}$ is a global minimizer.
    \end{proof} 

    Based on the Proposition \ref{prop:reformulation}, we have the following existence result for distributional robust Nash equilibrium of \ref{opt:DRG:1}. 

    \equilibrium*
    \begin{proof}
        Based on the Proposition \ref{prop:reformulation}, each $\mathbb{E}_{\boldsymbol{\xi}\sim\mathbb{Q}}[f_r(\boldsymbol{x}_r,\ \boldsymbol{x}_{-r},\ \boldsymbol{\xi})]$ is continuous and convex for any $\mathbb{Q}\in\boldsymbol{\mathfrak{U}}$. The supremum preserves the convexity of $f_r$ and, under weakly compactness of $\boldsymbol{\mathfrak{U}}$, the continuity of $f_r$ will hold. Therefore $\phi(\boldsymbol{v}, \boldsymbol{x})$ is continuous and convex w.r.t. $\boldsymbol{v}$ on $\mathcal{X}$ for any fixed $\boldsymbol{x}\in\mathcal{X}$.   

        The existence of an optimal solution to
        \begin{equation}
            \underset{\boldsymbol{v}\ \in\ \mathcal{X}}{\textbf{\textit{min}}}\ \phi(\boldsymbol{v},\ \boldsymbol{x})
        \end{equation}
        follows from compactness of $\mathcal{X}$ under the third condition in Assumption \ref{pro:equilibrium}. To complete the proof, we are left to show the existence of $\boldsymbol{x}^*\in\mathcal{X}$ such that
        \begin{equation}
            \boldsymbol{x}^*\ \in\ \underset{\boldsymbol{v}\ \in\ \mathcal{X}}{\textbf{\textit{arg\ min}}}\ \phi(\boldsymbol{v},\ \boldsymbol{x}^*).
        \end{equation}
        Let $\Phi(\boldsymbol{x})$ be the set of optimal solutions to $\underset{}{\textbf{\textit{min}}}\ \phi(\boldsymbol{v},\ \boldsymbol{x})$ for each fixed $\boldsymbol{x}\in\mathcal{X}$. Then $\Phi(\boldsymbol{x})\subset\mathcal{X}$ holds. By the convexity of $\phi$, $\Phi(\boldsymbol{x})$ is a convex set. Obviously, $\Phi(\boldsymbol{x})$ is closed, namely, there exists a sequence $\{\boldsymbol{x}_k\}$ with $\textbf{\textit{lim}}_{k\rightarrow\infty}\boldsymbol{x}_k=\boldsymbol{\bar{x}}$ and $\boldsymbol{v}_k\in\Phi(\boldsymbol{x}_k)$, if $\textbf{\textit{lim}}_{k\rightarrow\infty}\boldsymbol{v}_k=\boldsymbol{\bar{v}}$, we have $\boldsymbol{\bar{v}}\in\Phi(\boldsymbol{\bar{x}})$. Further, following by {Theorem 4.2.1} in \cite{bank1982non}, $\Phi$ is upper semi-continuous on $\mathcal{X}$. By Kakutani’s fixed point theorem \cite{kakutani1941}, , there exists $\boldsymbol{x}^*\in\mathcal{X}$ such that $\boldsymbol{x}^*\in\Phi(\boldsymbol{x}^*)$.
    \end{proof}

    \section{Proof of Theorem \ref{thm:main_result}.}
    \label{app:theorem_2}
    The following proposition shows the strong duality result for Wasserstein DRO \cite{doi:10.1287/moor.2018.0936}, which ensures that the inner supremum in \eqref{opt:robust_game_2} admits a reformulation which is a simple, univariate optimization problem. Note that there exists the other strong duality result of Wasserstein DRO \cite{gao2016distributionally}.
    \begin{proposition}
        \label{prop:duality}
        Let $d:\mathbb{R}^{n+2}\times\mathbb{R}^{n+2}\rightarrow[0,\infty]$ be a lower semi-continuous cost function satisfying $d(\boldsymbol{z},\ \boldsymbol{z}')=0$ whenever $\boldsymbol{z} = \boldsymbol{z}',\ \boldsymbol{z} = (p,\ \boldsymbol{b},\ y),\ \boldsymbol{z}' = (p',\ \boldsymbol{b}',\ y')$. For $\lambda\geq 0$ and loss function $\ell$ \eqref{eq:new_loss} that is upper semi-continuous in $(p,\ \boldsymbol{b},\ y)$ for each $\boldsymbol{\theta}$, define
        \begin{equation}
            \label{eq:psi_1}
            \psi_{\lambda,\ell}(\boldsymbol{z};\boldsymbol{\theta}) :=\underset{\boldsymbol{z}'\in\mathbb{R}^{n+2}}{\textbf{\textit{sup}}}\ \ \underset{(i,j)}{\sum}\ \Bigg\{\ \ell(\boldsymbol{\theta};\ \boldsymbol{z}')-\lambda d(\boldsymbol{z},\ \boldsymbol{z}')\ \Bigg\}.
        \end{equation}
        Then
        \begin{equation}
            \label{eq:duality}
            \underset{\mathbb{Q}\ \in\ \boldsymbol{\mathfrak{W}}^{\alpha}_p(\mathbb{P}_N)}{\textbf{\textit{sup}}}\ \ \mathbb{E}_{\boldsymbol{z}'\sim\mathbb{Q}}\Big[\ \ell\big(\boldsymbol{\theta},\ \boldsymbol{z}'\big)\ \Big] = \underset{\lambda\geq 0}{\textbf{\textit{min}}}\left\{\ \lambda\alpha+\frac{1}{N}\sum_{z}\psi_{\lambda,\ell}(\boldsymbol{z};\boldsymbol{\theta})\ \right\}
        \end{equation}
    \end{proposition}
    \mainresult*
    \begin{proof}
        Let $\Delta_{ij} = q_{ij}-p_{ij}$. The $\psi_{\lambda,\ell}$ function \eqref{eq:psi_2} has a new formulation as
        \begin{equation}
            \label{eq:psi_3}
            \begin{aligned}
                & & &\ \ \ \ \psi_{\lambda,\ell}(\boldsymbol{\theta},\ \boldsymbol{p})\\
                & &=&\ \ \underset{\boldsymbol{q}\ \in\ \mathbb{R}^N_+}{\textbf{\textit{sup}}}\ \frac{1}{N}\underset{\boldsymbol{c}\ \in\ \mathcal{C}}{\sum}\ \Big\{\ell(\boldsymbol{\theta},\ q_{ij})-\lambda\big[d(p_{ij},\ q_{ij})\big]^2\Big\}\\
                & &=&\ \ \underset{\boldsymbol{q}\ \in\ \mathbb{R}^N_+}{\textbf{\textit{sup}}}\ \frac{1}{N}\underset{\boldsymbol{c}\ \in\ \mathcal{C}}{\sum}\left\{\frac{q_{ij}}{2}\cdot\left[(-\boldsymbol{\theta}^\top,\ 1)\begin{pmatrix}
                    \boldsymbol{a}_{\boldsymbol{c}}\\ 
                    y_{ij}
                \end{pmatrix}\right]^2-\lambda\big|p_{ij}-q_{ij}\big|^2\right\}\\[1.5mm]
                & &=& \ \ \ \frac{1}{N}\underset{\boldsymbol{c}\ \in\ \mathcal{C}}{\sum}\ \underset{\Delta_{ij}\ \in\ \mathbb{R}}{\textbf{\textit{sup}}}\Big(\Delta_{ij} b_{ij}-\lambda \Delta_{ij}^2 + p_{ij}b_{ij}\Big),
            \end{aligned}
        \end{equation}
        where $b_{ij} = (y_{ij}-\boldsymbol{\theta}^\top\boldsymbol{a}_{\boldsymbol{c}})^2/2$, and the third equality holds due to $\psi_{\lambda, \ell}$ is a decomposable function. Expanding \eqref{eq:psi_3}, we can simplify $\psi_{\lambda,\ell}$ as below:
        \begin{equation}
            \begin{aligned}
                & & & \ \ \ \ \psi_{\lambda,\ell}(\boldsymbol{\theta},\ \boldsymbol{p})\\[1.5mm]
                & &=&\ \ \ \frac{1}{N}\langle\boldsymbol{p},\ \boldsymbol{b}\rangle + \frac{1}{N}\underset{\boldsymbol{c}\ \in\ \mathcal{C}}{\sum}\ \underset{\Delta_{ij}\ \in\ \mathbb{R}}{\textbf{\textit{sup}}}(\Delta_{ij} b_{ij}-\lambda \Delta_{ij}^2)\\[1mm]
                & &=&\ \ 
                \left\{
                \begin{array}{cc}
                    \langle\boldsymbol{p},\ \boldsymbol{b}\rangle/N+\|\boldsymbol{b}\|^2_2/(4\lambda N),\ &\ \text{if }\ \lambda>0,\\[2mm]
                    \infty,\ &\ \text{if }\ \lambda=0.
                \end{array}
                \right. 
            \end{aligned}
        \end{equation}
        Next, we investigate the duality of \eqref{opt:robust_game_2} with Proposition \ref{prop:duality}. As $\psi_{\lambda,\ell}(\boldsymbol{\theta},\ \boldsymbol{z}) = \infty$ when $\lambda=0$, the dual formulation of the supremum in \eqref{eq:primal_problem} would be
        \begin{equation}
            \label{eq:supmre_gamma}
            \begin{aligned}
                & & &\ \ \underset{\mathbb{Q}\ \in\ \mathfrak{W}^{\alpha}_p(\mathbb{P}_N)}{\textbf{\textit{sup}}}\ \ \mathbb{E}_{\boldsymbol{z}'\sim\mathbb{Q}}\Big[\ \ell\big(\boldsymbol{\theta},\ \boldsymbol{z}'\big)\ \Big]\\
                & & &=\ \ \underset{\lambda\geq0}{\textbf{\textit{min}}}\ \ \Bigg\{\lambda\alpha+\psi_{\lambda,\ell}(\boldsymbol{\theta},\ \boldsymbol{p})\Bigg\}\\
                & & &=\ \ \underset{\lambda>0}{\textbf{\textit{min}}}\ \left\{\lambda\alpha+\frac{1}{N}\langle\boldsymbol{p},\ \boldsymbol{b}\rangle+\frac{1}{4\lambda N}\|\boldsymbol{b}\|^2_2\right\}.
            \end{aligned}
        \end{equation}
        By the definition of $\boldsymbol{b}$, we know that 
        \begin{equation}
            \ell(\boldsymbol{\theta},\ \boldsymbol{p}) = \frac{1}{N}\ \big\langle\boldsymbol{p},\ \boldsymbol{b}\big\rangle
        \end{equation}
        Moreover, notice that the right hand side of \eqref{eq:supmre_gamma} is a convex function which approaches infinity when $\lambda\rightarrow\infty$, the global optimal of it can be obtained uniquely via the first order optimality condition as
        \begin{equation}
            \frac{\partial}{\partial \lambda} \left\{\ \lambda\alpha+\frac{1}{N}\langle\boldsymbol{p},\ \boldsymbol{b}\rangle+\frac{1}{4\lambda N}\|\boldsymbol{b}\|^2_2\ \right\} = 0,
        \end{equation}
        and the optimal dual variable is
        \begin{equation}
            \lambda^*_{\alpha} = \frac{\|\boldsymbol{b}\|_2}{2\sqrt{\alpha N}}.
        \end{equation}
        Substituting $\lambda^*_{\alpha}$ and $\boldsymbol{b}$ into \eqref{eq:supmre_gamma}, we have
        \begin{equation}
            \begin{aligned}
                & & &\ \ \underset{\mathbb{Q}\ \in\ \mathfrak{W}^{\alpha}_p(\mathbb{P}_N)}{\textbf{\textit{sup}}}\ \ \mathbb{E}_{\boldsymbol{z}'\sim\mathbb{Q}}\Big[\ \ell\big(\boldsymbol{\theta},\ \boldsymbol{z}'\big)\ \Big]\\
                & &=&\ \ \ \sqrt{\frac{\alpha}{N}}\cdot\|\boldsymbol{b}\|_2+\frac{1}{N}\cdot\langle\boldsymbol{p},\ \boldsymbol{b}\rangle\\
                & &=&\ \ \sqrt{\frac{\alpha}{4N}\underset{\boldsymbol{c}\ \in\ \mathcal{C}}{\sum}(y_{ij}-\boldsymbol{\theta}^\top\boldsymbol{a}_{\boldsymbol{c}})^2}+\frac{1}{2N}\underset{\boldsymbol{c}\ \in\ \mathcal{C}}{\sum}p_{ij}(y_{ij}-\boldsymbol{\theta}^\top\boldsymbol{a}_{\boldsymbol{c}})^2.
            \end{aligned}
        \end{equation}
    \end{proof}

    \section{Some Propositions for Generalization Analysis.}
    \label{app:thm:regression}
    \begin{restatable}{proposition}{Lipschitz}
        \label{prop:Lipschitz}
        Suppose that $\ell$ is $L$-Lipschitz function, \textit{i.e.}, $|\ell(\boldsymbol{z})-\ell(\boldsymbol{z}')|\leq L\cdot d_{\mathcal{Z}}(\boldsymbol{z},\boldsymbol{z}')$ for all $\boldsymbol{z},\boldsymbol{z}'\in\mathcal{Z}$. Then, for any $\mathbb{Q}\in\mathfrak{W}^{\alpha}_p(\mathbb{P}_N)$,
        \begin{equation}
            R_\mathbb{Q}(\ell) \leq R_{\mathbb{P},\alpha,p}(\ell) \leq R_\mathbb{Q}(\ell) + 2L\alpha.
        \end{equation}
    \end{restatable}
    \begin{proof}
        For $p=1$, the result follows immediately from the Kantorovich dual representation of $\mathcal{W}_1(\cdot,\ \cdot)$ \cite{villani2008optimal}:
        \begin{align}
            \label{eq:1_Wasserstein}
            \mathcal{W}_1(\mathbb{P},\ \mathbb{Q}) = \textbf{\textit{sup}}\ \Bigg\{\ \bigg|\ \mathbb{E}_{\boldsymbol{z}\sim\mathbb{P}}\big[h(\boldsymbol{z})\big]\ -\ \mathbb{E}_{\boldsymbol{z}\sim\mathbb{Q}}\big[h(\boldsymbol{z})\big]\ \bigg|\ \underset{\boldsymbol{z},\boldsymbol{z}'\in\mathcal{Z},\boldsymbol{z}\neq\boldsymbol{z}'}{\textbf{\textit{sup}}}\frac{\big|\ h(\boldsymbol{z})\ -\ h(\boldsymbol{z}')\ \big|}{d_{\mathcal{Z}}(\boldsymbol{z},\boldsymbol{z}')}\leq 1\Bigg\}
        \end{align}
        with the triangle inequality:
        \begin{equation}
            \mathcal{W}_1(\mathbb{P},\ \mathbb{Q})\leq 2\alpha,\ \ \ \ \forall\ \ \mathbb{P},\ \mathbb{Q}\ \in\ \mathfrak{W}^{\alpha}_1(\mathbb{P}_{N}).
        \end{equation}  

        For $p>1$, the result follows from the fact that
        \begin{equation}
            \mathcal{W}_1(\mathbb{P},\ \mathbb{Q})\ \leq\ \mathcal{W}_p(\mathbb{P},\ \mathbb{Q}),\ \ \ \ \forall\ \ \mathbb{P},\ \mathbb{Q}\ \in\ \mathcal{P}_p(\mathcal{Z}).
        \end{equation}
    \end{proof}
    Next we consider the case when the function $\ell$ is smooth but not Lipschitz-continuous. Since we are working with general metric spaces that may lack an obvious differentiable structure, we need to first introduce some concepts from metric geometry \cite{ambrosio2008gradient}.
    \begin{definition}[Geodesic Space]
        A metric space $(\mathcal{Z}, d_\mathcal{Z})$ is a geodesic space if for every pair of points $\boldsymbol{z},\ \boldsymbol{z}'\in\mathcal{Z}$ there exists a constant-speed geodesic path $\varrho:\big[0,1\big]\rightarrow\mathcal{Z}$, such that $\varrho(0) = \boldsymbol{z}$, $\varrho(1) = \boldsymbol{z}'$, and for all $0\leq s\leq t\leq 1$
        \begin{equation}
            d_{\mathcal{Z}}\Big[\varrho(s),\ \varrho(t)\Big]=(t-s)\cdot d_{\mathcal{Z}}\Big[\varrho(0),\ \varrho(1)\Big].  
        \end{equation}
    \end{definition}    

    \begin{definition}[Geodesic convexity]
        A functional $\ell:\mathcal{Z}\rightarrow\mathbb{R}$ is geodesically convex if for any pair of points $\boldsymbol{z},\ \boldsymbol{z}'\in\mathcal{Z}$ there is a constant-speed geodesic $\varrho$, so that
        \begin{equation}
            \begin{aligned}
                & \ell\Big(\varrho(t)\Big)&\leq&\ \ \ (1-t)\cdot\ell\Big(\varrho(0)\Big)\ +\ t\cdot\ell\Big(\varrho(1)\Big)\\
                & &=&\ \ \ (1-t)\cdot\ell(\boldsymbol{z})\ +\ t\cdot\ell(\boldsymbol{z}').           
            \end{aligned}
        \end{equation}
    \end{definition}    

    \begin{definition}[Upper Gradient]
        Suppose that $\ell:\mathcal{Z}\rightarrow\mathbb{R}$ is a Borel function. The upper gradient of $\ell$ is a functional $G_{\ell}:\mathcal{Z}\rightarrow\mathbb{R}_+$ satisfies that: for any pair of points $\boldsymbol{z},\ \boldsymbol{z}'\in\mathcal{Z}$, there exist a constant-speed geodesic path $\varrho$:
        \begin{equation}
            \label{eq:upper_gradient}
            |\ell(\boldsymbol{z}')-\ell(\boldsymbol{z})|\leq\int^1_0G_{\ell}(\varrho(t))\mathrm{d}t\cdot d_{\mathcal{Z}}(\boldsymbol{z},\ \boldsymbol{z}').
        \end{equation}
    \end{definition}

    
    \begin{restatable}{proposition}{uppergrad}
        \label{eq:pro_upper_grad}
        Suppose that $\ell$ has a geodesically convex upper gradient $G_{\ell}$, we have
        \begin{equation}
            R_\mathbb{Q}(\ell) \leq R_{\mathbb{P},\alpha,p}(\ell) \leq R_\mathbb{Q}(\ell) + 2\alpha\mu,
        \end{equation}
        where 
        \begin{equation}
            \mu = \underset{\mathbb{Q}\ \in\ \mathfrak{W}^{\alpha}_p(\mathbb{P})}{\textbf{\textit{sup}}}\Bigg(\mathbb{E}_{\boldsymbol{z}\sim\mathbb{Q}}\Big[\big|G_{\ell}(\boldsymbol{z})\big|^q\Big]\Bigg)^{\frac{1}{q}}
        \end{equation}
        and $1/p+1/q=1$.
    \end{restatable}
    \begin{proof}
        With fixed $\mathbb{Q},\ \mathbb{Q}'\in\mathfrak{W}^{\alpha}_p(\mathbb{P})$ and let $\gamma\in\Gamma(\mathcal{Z}\times\mathcal{Z})$ achieve the infimum in \eqref{eq:p_Wasserstein} and \eqref{eq:infty_Wasserstein} for $\mathcal{W}_p(\mathbb{Q},\ \mathbb{Q}')$. Then for any $(\boldsymbol{z},\ \boldsymbol{z}')\sim\gamma$, we have
        \begin{equation}
            \begin{aligned}
                & \ell(\boldsymbol{z}')-\ell(\boldsymbol{z})&\leq&\ \ \int^1_0G_{\ell}(\varrho(t))\mathrm{d}t\cdot d_{\mathcal{Z}}(\boldsymbol{z},\ \boldsymbol{z}')\\
                & &\leq&\ \ \frac{1}{2}\Big(G_{\ell}(\boldsymbol{z})+G_{\ell}(\boldsymbol{z}')\Big)\cdot d_{\mathcal{Z}}(\boldsymbol{z},\ \boldsymbol{z}'),
            \end{aligned}
        \end{equation}
        where the first inequality is from the definition of the upper gradient \eqref{eq:upper_gradient} and the second one is by the assumed geodesic convexity of $G_{\ell}$. Taking expectations of both sides with respect to $\gamma$ and using H\"older inequality, we obtain
        \begin{equation}
            \begin{aligned}
                & R_\mathbb{Q}(\ell)-R_{\mathbb{Q}'}(\ell)&\leq&\ \ \ \frac{1}{2}\ \Bigg(\mathbb{E}_{(\boldsymbol{z},\boldsymbol{z}')\sim\gamma}\Big[\big|G_{\ell}(\boldsymbol{z})+G_{\ell}(\boldsymbol{z}')\big|^q\Big]\Bigg)^{\frac{1}{q}}\Bigg(\mathbb{E}_{(\boldsymbol{z},\boldsymbol{z}')\sim\gamma}\big[d_{\mathcal{Z}}(\boldsymbol{z},\ \boldsymbol{z}')\big]^p\Bigg)^\frac{1}{p}\\
                & &=&\ \ \ \frac{1}{2}\ \Bigg(\mathbb{E}_{(\boldsymbol{z},\boldsymbol{z}')\sim\gamma}\Big[\big|G_{\ell}(\boldsymbol{z})+G_{\ell}(\boldsymbol{z}')\big|^q\Big]\Bigg)^{\frac{1}{q}}\cdot\mathcal{W}_p(\mathbb{Q},\ \mathbb{Q}'),\nonumber
            \end{aligned}
        \end{equation}  
        where we adopt the $p$-Wasserstein optimality of $\gamma$ for $\mathbb{Q}$ and $\mathbb{Q}'$. By the triangle inequality, and since $\boldsymbol{z}\sim\mathbb{Q}$ and $\boldsymbol{z}'\sim\mathbb{Q}$,
        \begin{equation}
            \begin{aligned}
                & \Bigg(\mathbb{E}_{(\boldsymbol{z},\boldsymbol{z}')\sim\gamma}\Big[\big|G_{\ell}(\boldsymbol{z})+G_{\ell}(\boldsymbol{z}')\big|^q\Big]\Bigg)^{\frac{1}{q}}&\leq&\ \ \Bigg(\mathbb{E}_{\boldsymbol{z}\sim\mathbb{Q}}\Big[\big|G_{\ell}(\boldsymbol{z})\big|^q\Big]\Bigg)^{\frac{1}{q}}+\ \ \Bigg(\mathbb{E}_{\boldsymbol{z}'\sim\mathbb{Q}'}\Big[\big|G_{\ell}(\boldsymbol{z}')\big|^q\Big]\Bigg)^{\frac{1}{q}}\\
                & &\leq&\ \ 2\underset{\mathbb{Q}\ \in\ \mathfrak{W}^{\alpha}_p(\mathbb{P})}{\textbf{\textit{sup}}}\ \Bigg(\mathbb{E}_{\boldsymbol{z}\sim\mathbb{Q}}\Big[\big|G_{\ell}(\boldsymbol{z})\big|^q\Big]\Bigg)^{\frac{1}{q}}.
            \end{aligned}
        \end{equation}
        Interchanging the roles of $\mathbb{Q}$ and $\mathbb{Q}'$ and proceeding with the same argument, we obtain the following estimation
        \begin{equation}
            \begin{aligned}
                \underset{\mathbb{Q},\ \mathbb{Q}'\ \in\ \mathfrak{W}^{\alpha}_p(\mathbb{P})}{\textbf{\textit{sup}}}\ \Big|\ R_\mathbb{Q}(\ell)\ -\ R_{\mathbb{Q}'}(\ell)\ \Big|\ \ \leq\ \ 2\alpha\underset{\mathbb{Q}\ \in\ \mathfrak{W}^{\alpha}_p(\mathbb{P})}{\textbf{\textit{sup}}}\ \Bigg(\mathbb{E}_{\boldsymbol{z}\sim\mathbb{Q}}\Big[\big|G_{\ell}(\boldsymbol{z})\big|^q\Big]\Bigg)^{\frac{1}{q}}.
            \end{aligned}
        \end{equation}
        Then
        \begin{equation}
            \begin{aligned}
                & & &\ \ R_\mathbb{Q}(\ell)\\
                & &\leq&\ \ R_{\mathbb{P},\alpha,p}(\ell)\\
                & &=&\ \ 2\alpha\underset{\mathbb{Q}'\ \in\ \mathfrak{W}^{\alpha}_p(\mathbb{P})}{\textbf{\textit{sup}}}\ \Big[R_{\mathbb{Q}',\alpha,p}(\ell)-R_\mathbb{Q}(\ell)+R_\mathbb{Q}(\ell)\Big]\\
                & &\leq&\ \ R_\mathbb{Q}(\ell) + 2\alpha\underset{\mathbb{Q}\ \in\ \mathfrak{W}^{\alpha}_p(\mathbb{P})}{\textbf{\textit{sup}}}\ \Bigg(\mathbb{E}_{\boldsymbol{z}\sim\mathbb{Q}}\Big[\big|G_{\ell}(\boldsymbol{z})\big|^q\Big]\Bigg)^{\frac{1}{q}}  
            \end{aligned}            
        \end{equation}

    \end{proof}

    \begin{restatable}{proposition}{regression}
        \label{thm:regression}
        Consider the setting of pairwise ranking problem with the sum-of-squared loss: let $\mathcal{A}$ be a convex subset of $\mathbb{R}^n$, $\mathcal{Y}=[-1,\ 1]$, and equip $\mathcal{Z} = \mathcal{A}\times\mathcal{Y}$ with the Euclidean metric
        \begin{equation}
            d_{\mathcal{Z}}(\boldsymbol{z},\boldsymbol{z}') = \sqrt{\|\boldsymbol{a}-\boldsymbol{a}'\|^2_2+|y-y'|^2},\ \boldsymbol{z}=(\boldsymbol{a},\ y).
        \end{equation}
        It means that we do not aggregate the pairwise comparisons into the same type and the weight. 
        Then, it holds that
        \begin{equation}
            R_{\mathbb{Q}}(\ell) \leq R_{\mathbb{P},\alpha,2}(\ell) \leq R_{\mathbb{Q}}(\ell) + 4\alpha(1+C)\tau,
        \end{equation}
        where
        \begin{equation}
            \tau = \left(1+L\underset{\mathbb{Q}\ \in\ \mathfrak{W}^{\alpha}_2(\mathbb{P}_N)}{\textbf{\textit{sup}}}\mathbb{E}_{\mathbb{Q}}\|\boldsymbol{A}\|_2\right),\ \ \boldsymbol{z}=(\boldsymbol{a},\ y)\sim\mathbb{Q},
        \end{equation}
        and $\boldsymbol{A}=[\boldsymbol{a}^\top_1,\dots,\boldsymbol{a}^\top_N]$.
    \end{restatable}

    \begin{proof}
        As $\mathcal{Z}\subseteq\mathbb{R}^{n+1}$, $\mathcal{Z}$ is a geodesic space as
        \begin{equation}
            \gamma(t) = (1-t)\cdot\boldsymbol{z}+t\cdot\boldsymbol{z}',\ \ \ \forall\ \boldsymbol{z},\ \boldsymbol{z}'\in\mathcal{Z}
        \end{equation}
        is the unique constant-speed geodesic path.
        
        Moreover, the geodesically convex upper gradient of $\ell$ is
        \begin{equation}
            \begin{aligned}
                G_{\ell}(\boldsymbol{z})=G_{\ell}(\boldsymbol{a},\ y)=2(B+C)(1+L\|\nabla h(\boldsymbol{a})\|_2),\ \ \forall\ \boldsymbol{z}\in\mathcal{Z}.    
            \end{aligned}
        \end{equation}
        where $\ell(\boldsymbol{z})=\ell(\boldsymbol{a},\ y)=(y-h(\boldsymbol{a}))^2$. In such a flat Euclidean setting, geodesic convexity coincides with the usual definition of convexity, and the map $\boldsymbol{z}\rightarrow G_{\ell}(\boldsymbol{z})$ is convex evidently: for all pair $\boldsymbol{z},\ \boldsymbol{z}'\in\mathcal{Z}$
        \begin{equation}
            G_{\ell}((1-t)\cdot\boldsymbol{z}+t\cdot\boldsymbol{z}')\leq (1-t)\cdot G_{\ell}(\boldsymbol{z}) + t\cdot G_{\ell}(\boldsymbol{z}').
        \end{equation}
        With the mean-value theorem
        \begin{equation}
            \begin{aligned}
                & & &\ \ \ \ \ell(\boldsymbol{z})-\ell(\boldsymbol{z}')\\
                & &\leq&\ \ \int^1_0\big\langle\boldsymbol{z}-\boldsymbol{z}',\ \nabla \ell\big((1-t)\cdot\boldsymbol{z}+t\cdot\boldsymbol{z}'\big)\big\rangle \mathrm{d}t\\
                & &\leq&\ \ \int^1_0\big\|\nabla \ell\big((1-t)\cdot\boldsymbol{z}+t\cdot\boldsymbol{z}'\big)\big\|_2 \mathrm{d}t\cdot\|\boldsymbol{z}-\boldsymbol{z}'\|_2\\
                & &=&\ \ \int^1_0\big\|\nabla \ell\big((1-t)\cdot\boldsymbol{z}+t\cdot\boldsymbol{z}'\big)\big\|_2 \mathrm{d}t\cdot d_{\mathcal{Z}}(\boldsymbol{z},\ \boldsymbol{z}')
            \end{aligned}
        \end{equation}
        and a simple calculation
        \begin{equation}
            \begin{aligned}
                \|\nabla \ell(\boldsymbol{z})\|^2_2=4\ell(\boldsymbol{z})(1+\|\nabla h(\boldsymbol{a})\|^2_2)\leq4(B+C)^2(1+L^2\|\boldsymbol{a}\|^2_2),
            \end{aligned}
        \end{equation}
        we have $\|\nabla \ell(\boldsymbol{z})\|_2\leq G_{\ell}(\boldsymbol{z})$ for any $\boldsymbol{z}\in\mathcal{Z}$. Thus, by Proposition \ref{prop:Lipschitz}, we have
        \begin{equation}
            \begin{aligned}
                & R_\mathbb{Q}(\ell)&\leq&\ \ \ R_{\mathbb{P},\alpha,2}(\ell)\\
                & &\leq&\ \ \ R_\mathbb{Q}(\ell) + 2\alpha\underset{\mathbb{Q}\ \in\ \mathfrak{W}^{\alpha}_2(\mathbb{P}_N)}{\textbf{\textit{sup}}}\ \Bigg(\mathbb{E}_{\boldsymbol{z}\sim\mathbb{Q}}\Big[\big|G_{\ell}(\boldsymbol{z})\big|^2\Big]\Bigg)^{\frac{1}{2}}\\
                & &=&\ \ \ R_\mathbb{Q}(\ell) + 4\alpha(B+C)\Bigg(1+L\underset{\mathbb{Q}\ \in\ \mathfrak{W}^{\alpha}_2(\mathbb{P}_N)}{\textbf{\textit{sup}}}\mathbb{E}_{\boldsymbol{z}\sim\mathbb{Q}}\|\boldsymbol{A}\|_2\Bigg).
            \end{aligned}            
        \end{equation}
    \end{proof}

    \section{Proof of Theorem \ref{thm:worst_prob}.}
    \label{app:thm:worst_prob}
    \begin{assumption}
        $d:\mathcal{Z}\times\mathcal{Z}\rightarrow\mathbb{R}_+$ in \eqref{eq:p_Wasserstein} and \eqref{eq:infty_Wasserstein} is a nonnegative lower semi-continuous function satisfying $d(\boldsymbol{w}, \boldsymbol{w}')=0$ if and only if $\boldsymbol{w}=\boldsymbol{w}'$.        
    \end{assumption}    

    \begin{assumption}
        The loss function $\ell\in\mathcal{F}\subseteq L^1(\mathrm{d}\mathbb{Q})$ are upper semi-continuous, where $L^1(\mathrm{d}\mathbb{Q})$ denote the collection of Borel measurable functions $\ell: \mathcal{Z}\rightarrow\mathbb{R}$ such that
        \begin{equation*}
            \int|\ell|\ \mathrm{d}\mathbb{Q} < \infty,\ \forall\ \mathbb{Q}\in\mathcal{P}(\mathcal{Z}).
        \end{equation*}
    \end{assumption}   

    \begin{assumption}
        \label{assump:1}
        The instance space $\mathcal{Z}$ is bounded, namely,
        \begin{equation}
            \text{diam}(\mathcal{Z}) = \underset{\boldsymbol{z},\boldsymbol{z}'\in\mathcal{Z}}{\textbf{\textit{sup}}} d_{\mathcal{Z}}(\boldsymbol{z},\ \boldsymbol{z}') < \infty.
        \end{equation}
    \end{assumption}    

    \begin{assumption}
        \label{assump:2}
        $\ell\in\mathcal{F}$ is uniformly bounded as
        \begin{equation}
             0\leq\ell(\boldsymbol{z})\leq B < \infty,\ \ \forall\ \ell\in\mathcal{F},\ \text{ and }\ \boldsymbol{z}\in\mathcal{Z}. 
        \end{equation}
    \end{assumption}

    \begin{definition}
        Let $(\mathcal{Z}, d_{\mathcal{Z}})$ be a metric space. For a function $\ell:\mathcal{Z}\rightarrow\mathbb{R}$ and a point $s\in\mathbb{R}$, the upper contour set defined by $s$ is
        \begin{equation}
            \ell^{-1}\big([s,\ \infty)\big)=\big\{\boldsymbol{z}\in\mathcal{Z}:\ \ell(\boldsymbol{z})\geq s\big\},
        \end{equation}
        and the corresponding lower contour set is
        \begin{equation}
            \ell^{-1}\big((\infty,\ s]\big)=\big\{\boldsymbol{z}\in\mathcal{Z}:\ \ell(\boldsymbol{z})\leq s\big\}.
        \end{equation}
        We call a function $\ell:\mathcal{Z}\rightarrow\mathbb{R}$ is upper semi-continuous if and only if for any $s\in\mathbb{R}$, $\ell^{-1}\big((\infty,\ s]\big)$ is an open set.
    \end{definition}    

    We adopt the Dudley’s entropy integral \cite{DUDLEY1967290} as the complexity measure of the hypothesis class $\mathcal{F}$,
    \begin{equation}
        \label{eq:entropy_integral}
        \mathcal{J}(\mathcal{F})=\int_0^{\infty}\sqrt{\textbf{\textit{log}}\ \mathfrak{N}(\mathcal{F},\ \|\cdot\|_{\infty},\ \upsilon )}\ \mathrm{d}\upsilon,
    \end{equation}
    where $\mathfrak{N}(\mathcal{F},\ \|\cdot\|_{\infty},\ \upsilon)$ is $\upsilon$-covering number of $\mathcal{F}$ with respect to the uniform metric $\|\cdot\|_{\infty}$, defined as the size of the smallest $\upsilon$-cover of $\mathcal{F}$
    \begin{equation}
        \begin{aligned}
        & & &\ \ \mathfrak{N}(\mathcal{F},\ \|\cdot\|_{\infty},\ \upsilon) \\
        & &=&\ \ \underset{m\in\mathbb{N}}{\textbf{\textit{min}}}\ \left\{\exists\ \{\ell_1,\ \dots,\ \ell_m\}\subseteq\mathcal{F}\subseteq\bigcup_{k=1}^m\mathcal{B}^{\|\cdot\|_\infty}_\upsilon(\ell_k)\right\}
        \end{aligned}
    \end{equation}
    and $\bigcup_{k=1}^m\mathcal{B}^{\|\cdot\|_\infty}_\upsilon(\cdot)$ is a $\upsilon$-cover of $\mathcal{F}$ with respect to $\|\cdot\|_{\infty}$
    \begin{equation}
        \|\ \ell-{\ell}'\ \|_{\infty} = \underset{\boldsymbol{z}\ \in\ \mathcal{Z}}{\textbf{\textit{sup}}}\ |\ \ell(\boldsymbol{z})\ -\ {\ell}'(\boldsymbol{z})\ |.
    \end{equation}

    \worstprob*
    \begin{proof}
        This proof is a specialization of data-dependent generalization bounds for margin cost function class \cite{koltchinskii2002}. From the definition of the local minimax risk \eqref{eq:local_minimax_risk} and its duality form,
        \begin{equation}
            \begin{aligned}
                 & R_{\mathbb{P},\alpha,p}(\ell)&=&\ \ \ \underset{\lambda>0}{\textbf{\textit{min}}}\ \Big\{\ \lambda\alpha^p\ +\ \mathbb{E}_{\boldsymbol{z}\sim\mathbb{P}}\big[\psi_{\lambda,\ell}(\boldsymbol{z})\big]\ \Big\}\\
                 & &\leq&\ \ \ \underset{\lambda>0}{\textbf{\textit{min}}}\ \Big\{\ \lambda\alpha^p\ +\ \mathbb{E}_{\boldsymbol{z}\sim\mathbb{P}}\big[\psi_{\lambda,\ell}(\boldsymbol{z})\big]\ +\ \boldsymbol{V}_{\lambda}\Big\}
            \end{aligned}
        \end{equation}
        where
        \begin{equation}
            \boldsymbol{V}_{\lambda}\ \ =\ \ \underset{\ell\ \in\ \mathcal{F}}{\textbf{\textit{sup}}}\ \Big\{\ \mathbb{E}_{\boldsymbol{z}\sim\mathbb{P}}\big[\psi_{\lambda,\ell}(\boldsymbol{z})\big]-\mathbb{E}_{\boldsymbol{z}\sim\mathbb{P}_N}\big[\psi_{\lambda,\ell}(\boldsymbol{z})\big]\ \Big\}
        \end{equation}
        is a data-dependent random variable for any $\lambda\geq 0$. As $\mathcal{F}$ and $\mathbb{P}$ satisfy the Assumption \ref{assump:1} and \ref{assump:2}, we have
        \begin{equation}
             0\leq\psi_{\lambda,\ell}(\boldsymbol{z})\leq B,\ \forall\ \boldsymbol{z}\in\mathcal{Z}.
        \end{equation}
        Furthermore, known from McDiarmid’s inequality that, for any fixed $\lambda\geq 0$
        \begin{equation}
            \Pr\Big(\boldsymbol{V}_{\lambda}\geq\mathbb{E}\boldsymbol{V}_{\lambda} + \frac{Bt}{\sqrt{N\ }\ \ }\Big)\ \leq\ 2e^{-2t^2}.
        \end{equation}
        Using a standard symmetrization argument, we have
        \begin{equation}
            \label{eq:V_bound}
            \mathbb{E}\boldsymbol{V}_{\lambda}\leq 2\cdot\mathbb{E}\left[\underset{\ell\ \in\ \mathcal{F}}{\textbf{\textit{sup}}}\ \frac{1}{N}\sum^{N}_{i=1}\ \epsilon_i\psi_{\lambda,\ell}(\boldsymbol{z_i})\right]
        \end{equation}
        where $\epsilon_1,\dots,\epsilon_N$ are \textit{i.i.d.} Rademacher random variables independent of $\boldsymbol{z}_1,\dots,\boldsymbol{z}_N$.   

        To bound \eqref{eq:V_bound}, we define the $\mathcal{F}$-indexed process $\boldsymbol{\beta}_\mathcal{F}=\{\beta_{\ell}\}_{\ell\in\mathcal{F}}$ as
        \begin{equation}
            \beta_{\ell} = \frac{1}{N}\sum^{N}_{i=1}\ \epsilon_i\psi_{\lambda,\ell}(\boldsymbol{z_i}).
        \end{equation}
        This is a zero-mean and sub-Gaussian process \cite{vanderVaart1996} with respect to the metric $\|\cdot\|_{\infty}$ as
        \begin{equation}
            \begin{aligned}
                & & &\ \ \mathbb{E}\ \Big[\textbf{\textit{exp}}\big(t(\beta_{\ell}-\beta_{{\ell}'})\big)\Big]\\
                & &=&\ \ \mathbb{E}\left[\textbf{\textit{exp}}\left(\frac{t}{\sqrt{N}\ \ }\sum^{N}_{i=1}\ \epsilon_i\big(\psi_{\lambda,\ell}(\boldsymbol{z_i})-\psi_{\lambda,{\ell}'}(\boldsymbol{z_i})\big)\right)\right]\\
                & &=&\ \ \Bigg\{\mathbb{E}\Bigg[\textbf{\textit{exp}}\Bigg(\frac{t}{\sqrt{N\ }\ \ }\epsilon_i\ \underset{\boldsymbol{z}'}{\textbf{\textit{sup}}}\ \underset{\boldsymbol{z}''}{\textbf{\textit{inf}}}\ \Big\{\ell(\boldsymbol{z}')-\lambda[d_{\mathcal{Z}}(\boldsymbol{z}_1,\boldsymbol{z}')]^p-{\ell}'(\boldsymbol{z}'')+\lambda[d_{\mathcal{Z}}(\boldsymbol{z}_1,\boldsymbol{z}'')]^p\Big\}\Bigg)\Bigg]\Bigg\}^N\\
                & &\leq&\ \ \Bigg\{\mathbb{E}\Bigg[\textbf{\textit{exp}}\left(\frac{t}{\sqrt{N\ }\ \ }\epsilon_i\ \underset{\boldsymbol{z}'}{\textbf{\textit{sup}}}\ \Big\{\ell(\boldsymbol{z}')-{\ell}'(\boldsymbol{z}')\Big\}\right)\Bigg]\Bigg\}^N\\
                & &\leq&\ \ \ \textbf{\textit{exp}}\left(\frac{t^2\|\ell-{\ell}'\|^2_{\infty}}{2}\right),
            \end{aligned}
        \end{equation}
        where the second equation comes from the  independence of $\{\boldsymbol{z}_i\}_{i\in[N]}$ and the definition of $\psi_{\lambda,\ell}(\cdot)$. The last inequality follows the Hoeffding’s lemma \cite{10.2307/2282952}.    

        With the $\mathcal{F}$-indexed process $\boldsymbol{\beta}_\mathcal{F}$ and invoking Dudley’s entropy integral \eqref{eq:entropy_integral} \cite{DUDLEY1967290} for the right-hand side of \eqref{eq:V_bound}, we obtain
        \begin{equation}
            \mathbb{E}\boldsymbol{V}_{\lambda}\leq 2\cdot\mathbb{E}\left[\underset{\ell\ \in\ \mathcal{F}}{\textbf{\textit{sup}}}\ \beta_{\ell}\right]\leq\frac{24}{\sqrt{N\ }\ \ }\ \mathcal{J}(\mathcal{F}),\ \ \ \forall\ \lambda\geq 0
        \end{equation}
        and
        \begin{equation}
            \Pr\Big(\boldsymbol{V}_{\lambda}\geq \frac{24\mathcal{J}(\mathcal{F})+Bt}{\sqrt{N\ }}\Big)\ \leq\ 2e^{-2t^2}.
        \end{equation}
        In addition, the first part of the claims holds with ant fixed $\lambda\geq 0$:
        \begin{equation*}
            \Pr\left(\ \exists\ \ell\in\mathcal{F}: R_{\mathbb{P}, \alpha, p}(\ell)>\varsigma_1\ \right)\ \leq\ e^{-2t^2},\ \ \forall\ t>0,
        \end{equation*}
        where
        \begin{equation*}
            \varsigma_1 = \underset{\lambda \geq 0}{\textbf{\textit{min}}}\ \ \Bigg\{\ \lambda\alpha^p\ +\ \mathbb{E}_{\boldsymbol{z}\sim\mathbb{Q}}\ \big[\ \psi_{\lambda,\ell}\ (\boldsymbol{z})\ \big]\Bigg\}+\frac{24\mathcal{J}(\mathcal{F})+Bt}{\sqrt{N\ }}.
        \end{equation*}
        
        For the second part, we start with two sequences: $\{\lambda_k\}$ and $\{t_k\}$
        \begin{equation}
            \lambda_k = k,\ \ t_k = t+\sqrt{\textbf{\textit{log}}(k)},\ k=1,2,\dots
        \end{equation}
        and \eqref{eq:union_bound} also holds as
        \begin{equation}
            \begin{aligned}
                & & &\ \ \Pr\left(\ \exists\ \ell\in\mathcal{F}: R_{\mathbb{P}, \alpha, p}(\ell)>\underset{k}{\textbf{\textit{min}}}\ \ \Bigg\{\ \lambda_k\alpha^p\ +\ \mathbb{E}_{\boldsymbol{z}\sim\mathbb{Q}}\ \big[\ \psi_{\lambda_k,\ell}\ (\boldsymbol{z})\ \big]\Bigg\}+\frac{24\mathcal{J}(\mathcal{F})+Bt_k}{\sqrt{N\ }}\ \right)\\
                & &\leq&\ \ \sum_{k}\ e^{-2t_k^2}\\
                & &\leq&\ \ \sum_{k}\ e^{-2\textbf{\textit{log}}(k)}\ \cdot\ e^{-2t^2}\\
                & &\leq&\ \ 2e^{-2t^2}.
            \end{aligned}
        \end{equation}
        Moreover,
        \begin{equation}
            \begin{aligned}
                & & &\ \ \underset{k}{\textbf{\textit{min}}}\ \ \Bigg\{\ \lambda_k\alpha^p\ +\ \mathbb{E}_{\boldsymbol{z}\sim\mathbb{Q}}\ \big[\ \psi_{\lambda_k,\ell}\ (\boldsymbol{z})\ \big]\Bigg\}+\frac{24\mathcal{J}(\mathcal{F})+Bt_k}{\sqrt{N\ }}\\
                & &=&\ \ \underset{k}{\textbf{\textit{min}}}\ \ \Bigg\{\ \ k\alpha^p\ \ +\ \mathbb{E}_{\boldsymbol{z}\sim\mathbb{Q}}\ \big[\ \psi_{\lambda_k,\ell}\ (\boldsymbol{z})\ \big]\Bigg\}+\frac{24\mathcal{J}(\mathcal{F})+Bt}{\sqrt{N\ }} + \frac{B\sqrt{\textbf{\textit{log}}(k)}}{\sqrt{N\ }}\\
                & &\leq&\ \ \underset{\lambda\geq 0}{\textbf{\textit{min}}}\ \ \Bigg\{\ \ (\lambda+1)\alpha^p\ \ +\ \mathbb{E}_{\boldsymbol{z}\sim\mathbb{Q}}\ \big[\ \psi_{\lambda,\ell}\ (\boldsymbol{z})\ \big]\Bigg\}+\frac{24\mathcal{J}(\mathcal{F})+Bt}{\sqrt{N\ }} + \frac{B\sqrt{\textbf{\textit{log}}(\lambda+1)}}{\sqrt{N\ }}\\
            \end{aligned}
        \end{equation}
        where the last inequity holds since, for any $\lambda\geq 0$, there exists $k\in\mathbb{N}_+$ such that $\lambda\leq k\leq \lambda+1$, and $\psi_{\lambda_1,\ell}\leq\psi_{\lambda_2,\ell}$ holds whenever $\lambda_1\geq\lambda_2$ as \eqref{eq:psi_1}.  

        Notice that
        \begin{equation}
            R_{\mathbb{P}_N, \alpha,p}(\ell)\leq\underset{\lambda>0}{\textbf{\textit{min}}}\ \Big\{\ \lambda\alpha^p\ +\ \mathbb{E}_{\boldsymbol{z}\sim\mathbb{P}_N}\big[\psi_{\lambda,\ell}(\boldsymbol{z})\big]\ +\ \boldsymbol{W}_{\lambda}\Big\},
        \end{equation}
        where
        \begin{equation}
            \boldsymbol{W}_{\lambda}\ \ =\ \ \underset{\ell\in\mathcal{F}}{\textbf{\textit{sup}}}\ \Big\{\ \mathbb{E}_{\boldsymbol{z}\sim\mathbb{P}_N}\big[\psi_{\lambda,\ell}(\boldsymbol{z})\big]\ -\ \mathbb{E}_{\boldsymbol{z}\sim\mathbb{P}}\big[\psi_{\lambda,\ell}(\boldsymbol{z})\big]\ \Big\}.
        \end{equation}
        Following the similar analysis of $R_{\mathbb{P},\alpha,p}(\ell)$, the second part of the claims holds
        \begin{equation*}
            \label{eq:empirical_bound_2}
            \Pr\left(\ \exists\ \ell\in\mathcal{F}: R_{\mathbb{P}_N, \alpha, p}(\ell)>\varsigma_2\ \right)\ \leq\ 2e^{-2t^2},\ \forall\ t>0
        \end{equation*}
        where
        \begin{equation*}
            \begin{aligned}
                \varsigma_2=\underset{\lambda \geq 0}{\textbf{\textit{min}}}\ \ \Bigg\{\ (\lambda+1)\alpha^p\ +\ \mathbb{E}_{\boldsymbol{z}\sim\mathbb{Q}}\ \big[\ \psi_{\lambda,\ell}\ (\boldsymbol{z})\ \big]+\frac{B\sqrt{\textbf{\textit{log}}(\lambda+1)}}{\sqrt{N\ }}\ \Bigg\}+\frac{24\mathcal{J}(\mathcal{F})+Bt}{\sqrt{N\ }}.
            \end{aligned}
        \end{equation*}
    \end{proof}

    \section{Proof of Theorem \ref{thm:excess_risk}.}
    \label{app:thm:excess_risk}
    The common choice of the smoothness assumption is Lipschitz smoothness. Next, we explore the behavior of the dual variable $\lambda$ in \eqref{eq:duality} when the \eqref{eq:duality} archives the minimal. The following lemma enables the control of its upper bound.   

    \begin{assumption}
        \label{assump:Lipschitz}
        The functions in $\mathcal{F}$ are $L$-Lipschitz, if they satisfy
        \begin{equation}
            \underset{\boldsymbol{z},\boldsymbol{z}'\in\mathcal{Z},\boldsymbol{z}\neq\boldsymbol{z}'}{\textbf{\textit{sup}}}\frac{\ \ell(\boldsymbol{z}')-\ell(\boldsymbol{z})\ }{d_{\mathcal{Z}}(\boldsymbol{z}',\ \boldsymbol{z})}\leq L,\ \forall\ \ell\in\mathcal{F}.
        \end{equation}
    \end{assumption}

    \begin{restatable}{lemma}{lambdacontrol}
        \label{lemma:lambda_control}
        Suppose that $\mathbb{Q}\in\mathfrak{W}_p^{\alpha}(\mathbb{P}_n)\subset\mathcal{P}_{m}(\mathcal{Z})$ and $\tilde{\ell}$ is the optimal solution of local worst-case risk with distribution $\mathbb{Q}$
        \begin{equation}
            \tilde{\ell}\in\underset{\ell\ \in\ \mathcal{F}}{\textbf{\textit{arg min}}}\ R_{\mathbb{Q}, \alpha, p}\ (\ell),
        \end{equation}
        $\tilde{\lambda}$ is the infimum-archiving dual variable corresponding to $\tilde{f}$
        \begin{equation}
            \tilde{\lambda}\in\underset{\lambda\geq0}{\textbf{\textit{min}}}\ \bigg\{\ \lambda\alpha^p\ +\ \mathbb{E}_{\boldsymbol{z}\sim\mathbb{Q}}\ \Big[\psi_{\lambda,\ \tilde{\ell}}(\boldsymbol{z})\Big]\ \bigg\}.
        \end{equation}
        Then under Assumption \ref{assump:1}-\ref{assump:Lipschitz}, $\tilde{\lambda}$ satisfies
        \begin{equation}
             \tilde{\lambda}\ \leq\ L\alpha^{-(p-1)}.
        \end{equation}
    \end{restatable}
    \begin{proof}
        With the fixed $\mathbb{Q}$ and the estimator $\tilde{f}$, we have
        \begin{equation}
            \begin{aligned}
                \label{eq:lemma_1_1}
                \tilde{\lambda}\alpha^p\leq\tilde{\lambda}\alpha^p+\mathbb{E}_{\boldsymbol{z}\sim\mathbb{Q}}\left[\underset{\boldsymbol{z}'\in\mathcal{Z}}{\textbf{\textit{sup}}}\bigg\{\tilde{f}(\boldsymbol{z}')-\tilde{f}(\boldsymbol{z})-\tilde{\lambda}\Big[d_{\mathcal{Z}}(\boldsymbol{z},\boldsymbol{z}')\Big]^p\bigg\}\right]\\   
            \end{aligned}
        \end{equation}
        and the equality holds with $\boldsymbol{z}'=\boldsymbol{z}$. Due to the optimality of $\tilde{\lambda}$ and the dual formulation of local worst-case risk \eqref{eq:duality},  \eqref{eq:lemma_1_1} can be further bounder as
        \begin{equation}
            \begin{aligned}
                &\tilde{\lambda}\alpha^p&\leq&\ \ \lambda\alpha^p+\mathbb{E}_{\boldsymbol{z}\sim\mathbb{Q}}\left[\underset{\boldsymbol{z}'\in\mathcal{Z}}{\textbf{\textit{sup}}}\bigg\{\tilde{f}(\boldsymbol{z}')-\tilde{f}(\boldsymbol{z})-\lambda\Big[d_{\mathcal{Z}}(\boldsymbol{z},\boldsymbol{z}')\Big]^p\bigg\}\right]\\
                & &\leq&\ \ \lambda\alpha^p+\mathbb{E}_{\boldsymbol{z}\sim\mathbb{Q}}\left[\underset{\boldsymbol{z}'\in\mathcal{Z}}{\textbf{\textit{sup}}}\bigg\{L\cdot d_{\mathcal{Z}}(\boldsymbol{z},\boldsymbol{z}')-\lambda\Big[d_{\mathcal{Z}}(\boldsymbol{z},\boldsymbol{z}')\Big]^p\bigg\}\right]\\
                & &\leq&\ \ \lambda\alpha^p + \underset{v\geq 0}{\textbf{\textit{sup}}}\ \big\{Lv-\lambda v^p\big\},
            \end{aligned}
        \end{equation}
        where the second line comes from the Lipschitz smoothness of $\ell\in\mathcal{F}$ and the third line holds by substituting $v=d_{\mathcal{Z}}(\boldsymbol{z},\boldsymbol{z}')$. When $p=1$, the result can be obtained by taking $\lambda=L$
        \begin{equation}
            \tilde{\lambda}\alpha\leq L\alpha+ \underset{\boldsymbol{z}'\in\mathcal{Z}}{\textbf{\textit{sup}}}\big\{Lv-Lv\big\}=L\alpha.
        \end{equation}
        If $p > 1$, we can take the $v^*=\left(\frac{L}{\lambda p}\right)^{\frac{1}{p-1}}$ which satisfies first-order optimal condition for 
        \begin{equation}
            \underset{v\geq 0}{\textbf{\textit{sup}}}\ \big\{Lv-\lambda v^p\big\}
        \end{equation}
        and
        \begin{equation}
            \label{eq:lemma_2}
            \tilde{\lambda}\alpha^p\leq\lambda\alpha^p + (p-1)L^{\frac{p}{p-1}}p^{\frac{p}{1-p}}\lambda^{\frac{1}{1-p}}.
        \end{equation}
        Treating $\lambda$ as a variable and minimizing the right-hand side of \eqref{eq:lemma_2} by choosing $\lambda = \frac{L}{p\alpha^{p-1}}$, the claim holds.
    \end{proof}

    \excessrisk*
    \begin{proof}
        Suppose that $\ell^*\in\mathcal{F}$ can archive the local minimax risk $R^*_{\mathbb{P},\alpha,p}(\mathcal{F})$, we decompose the excess risk 
        \begin{equation}
            \label{eq:theorem_3_1}
            \begin{aligned}
                & R_{\mathbb{P},\alpha,p}(\hat{\ell})-R^*_{\mathbb{P},\alpha,p}(\mathcal{F})
                &=&\ \ R_{\mathbb{P},\alpha,p}(\hat{\ell})-R^*_{\mathbb{P},\alpha,p}(\ell^*)\\
                & &\leq&\ \ R_{\mathbb{P},\alpha,p}(\hat{\ell})-R_{\mathbb{P}_N,\alpha,p}(\hat{\ell})+R^*_{\mathbb{P}_N,\alpha,p}(\ell^*)-R^*_{\mathbb{P},\alpha,p}(\ell^*),\\
            \end{aligned}
        \end{equation}
        where the last equality stands by the optimality of $\hat{\ell}$.   

        Next, we introduce $\hat{\lambda}$ and $\lambda^*$ as the corresponding dual variables of $\hat{\ell}$ and $\ell^*$ as
        \begin{equation}
            \hat{\ell}\ \in\ \ \underset{\lambda\geq0}{\textbf{\textit{min}}}\ \bigg\{\ \lambda\alpha^p\ +\ \mathbb{E}_{\boldsymbol{z}\sim\mathbb{P}_N}\ \Big[\psi_{\lambda,\ \hat{\ell}}(\boldsymbol{z})\Big]\ \bigg\},
        \end{equation}
        and
        \begin{equation}
            \ell^*\ \in\ \underset{\lambda\geq0}{\textbf{\textit{min}}}\ \bigg\{\ \lambda\alpha^p\ +\ \mathbb{E}_{\boldsymbol{z}\sim\mathbb{P}}\ \Big[\psi_{\lambda,\ \ell^*}(\boldsymbol{z})\Big]\ \bigg\}.
        \end{equation}
        By the first part of Theorem \ref{thm:main_result}, the right-hand side of \eqref{eq:theorem_3_1} can be further bounded by
        \begin{equation}
            \label{eq:theorem_3_2}
            \begin{aligned}
                & R_{\mathbb{P},\alpha,p}(\hat{\ell})-R_{\mathbb{P}_N,\alpha,p}(\hat{\ell})&=&\ \ \underset{\lambda\geq0}{\textbf{\textit{min}}}\left\{\ \lambda\alpha^p\ +\ \int_{\mathcal{Z}}\psi_{\lambda,\hat{\ell}}(\boldsymbol{z})\mathbb{P}(\mathrm{d}\boldsymbol{z})\ \right\}-\left(\ \hat{\lambda}\alpha^p\ +\ \int_{\mathcal{Z}}\psi_{\hat{\lambda},\hat{\ell}}(\boldsymbol{z})\mathbb{P}_N(\mathrm{d}\boldsymbol{z})\ \right)\\
                & &\leq&\ \ \int_{\mathcal{Z}}\psi_{\hat{\lambda},\hat{\ell}}(\boldsymbol{z})(\mathbb{P}-\mathbb{P}_N)(\mathrm{d}\boldsymbol{z}),
            \end{aligned}
        \end{equation}
        and
        \begin{equation}
            \label{eq:theorem_3_3}
            \begin{aligned}
                R^*_{\mathbb{P}_N,\alpha,p}(\ell^*)-R^*_{\mathbb{P},\alpha,p}(\ell^*)\ \leq\ \int_{\mathcal{Z}}\psi_{{\lambda^*},{\ell^*}}(\boldsymbol{z})(\mathbb{P}_N-\mathbb{P})(\mathrm{d}\boldsymbol{z}).
            \end{aligned}
        \end{equation}
        
        By Lemma \ref{lemma:lambda_control}, we know
        \begin{equation}
            \hat{\lambda}\in\boldsymbol{\Lambda}:=\big[\ 0,\ L\alpha^{-(p-1)}\ \big]
        \end{equation}
        and define the function class
        \begin{equation}
            \boldsymbol{\Psi} = \Big\{\psi_{\lambda,\ \ell}\ \big\vert\ \lambda\in\boldsymbol{\Lambda},\ \ell\in\mathcal{F}\Big\},
        \end{equation}
        \eqref{eq:theorem_3_2} can be written as 
        \begin{equation}
            \begin{aligned}
                R_{\mathbb{P},\alpha,p}(\hat{\ell})\ -\ R_{\mathbb{P}_N,\alpha,p}(\hat{\ell})\ \ \leq\ \ \underset{\psi\in\boldsymbol{\Psi}}{\textbf{\textit{sup}}}\ \left\{\ \int_{\mathcal{Z}}\ \psi\ \mathrm{d}(\ \mathbb{P}-\mathbb{P}_N\ )\ \right\}.
            \end{aligned}
        \end{equation}
        By Assumption \ref{assump:1}, \ref{assump:2} and the definition of $\psi_{\lambda,\ \ell}$ as \eqref{eq:psi_1}, we know that every $\psi\in\boldsymbol{\Psi}$ is bounded and take value in $[0,\ B]$. Employing symmetrization, we have
        \begin{equation}
            \label{eq:theorem_3_4}
            R_{\mathbb{P},\alpha,p}(\hat{\ell})-R_{\mathbb{P}_N,\alpha,p}(\hat{\ell}) \leq 2\mathfrak{R}_N(\boldsymbol{\Psi}) + B\sqrt{\frac{\textbf{\textit{log}}(\frac{2}{\eta})}{N}}
         \end{equation}
         with probability at least $1-\frac{\eta}{2}$, where
         \begin{equation}
             \mathfrak{R}_N(\boldsymbol{\Psi}) = \mathbb{E}\left[\ \underset{\psi\in\boldsymbol{\Psi}}{\textbf{\textit{sup}}}\ \frac{1}{N}\sum^{N}_{i=1}\epsilon_i\psi(\boldsymbol{z})\ \right]
         \end{equation}
         is the expected Rademacher average of $\boldsymbol{\Psi}$, with \textit{i.i.d} Rademacher random variables $\{\epsilon_i\}$ which are independent of $\{\boldsymbol{z}_i\}$, $i\in[N]$. Moreover, from Hoeffding’s inequality, it follows that
         \begin{equation}
            \label{eq:theorem_3_5}
            R^*_{\mathbb{P}_N,\alpha,p}(\ell^*)-R^*_{\mathbb{P},\alpha,p}(\ell^*)\leq B\sqrt{\frac{\textbf{\textit{log}}(\frac{2}{\eta})}{2N}}
         \end{equation}
         with probability at least $1-\frac{\eta}{2}$. Combining \eqref{eq:theorem_3_4} and \eqref{eq:theorem_3_5}, and apply the Lemma from Appendix, we obtain the whole theorem.
    \end{proof}

    \section*{The Stackelberg Game Attack on Pairwise Ranking.}
    We study the poisoning attack on pairwise ranking, which injects the malicious pairwise comparisons into the training set of the ranking algorithm. Meanwhile, the robust ranking algorithm could prune the outlier when leaning a consensus ranking with the noise observation. Such an adversarial interaction between two opponents can is naturally a game. One player will control the ranking algorithm, and the other player will manipulate the distribution of input data, especially the pairwise comparisons. The optimal action for each player generally depends on both players’ strategies.

    We adopt positive integers to index alternatives and users. Henceforth, $\boldsymbol{V}$ always is the set $\{1, \dots, n\}$ and denotes a set of alternatives to be ranked. In our approach to attack pairwise ranking, we represent these candidates as vertices of a graph. $\boldsymbol{U}$ = $\{1, \dots, m\}$ denotes a set of voters or users. For $i,\ j\in\boldsymbol{V}$, we write the pairwise comparison $i\succ j$ or $(i,\ j)$ to mean that alternative $i$ is preferred over alternative $j$. If we hope to emphasize the preference judgment of a particular user $u$, we will write $i\succ^{u} j$ or $(u,\ i,\ j)$. 

    For each user $u\in\boldsymbol{U}$, the pairwise ranking matrix of user $u$ is a skew-symmetric matrix
    \begin{equation}
        \boldsymbol{Y}^u=\{y^u_{ij}\}\in\mathbb{R}^{n\times n},\ i,\ j\in\boldsymbol{V},\ u\in\boldsymbol{U},
    \end{equation}
    \textit{i.e.} for any ordered pair $(i,\ j)\in\boldsymbol{V}\times\boldsymbol{V}$ , we have
    \begin{equation}
        y^u_{ij} = -y^u_{ji}.
    \end{equation}
    Informally, $y^u_{ij}$ measures the ``degree of preference'' of the $i^{\text{th}}$ alternative over the $j^{\text{th}}$ alternative held by the $u^{\text{th}}$ voter. 
    Here we focus on the ``binary'' case of $\boldsymbol{Y}^u\in\{-1,\ 1\}^{n\times n}$. Here $y^u_{ij} = 1$ means there exist a particular preference judgment $(u,\ i,\ j)$ made by user $u$. 

    Define the weight function $w:\boldsymbol{U}\times\boldsymbol{V}\times\boldsymbol{V}\rightarrow[0,\ \infty)$ as the indicator function
    \begin{equation}
        w^u_{ij}=w(u,i,j)=
        \left\{\begin{matrix}
        1, & \text{if}\ y^u_{ij} = 1,\\ 
        0, & \text{otherwise.}
        \end{matrix}\right.
    \end{equation}
    With the weight function $w$, we can aggregate all users' pairwise comparison matrices $\{\boldsymbol{Y}^u\},\ u\in\boldsymbol{U}$ into a single comparison matrix $\boldsymbol{Y}=\left\{y_{ij}\right\}$ with weights matrix $\boldsymbol{W}^0=\{w^0_{ij}\}$, where
    \begin{equation}
        y_{ij} = 1,\ \forall\ (i,\ j)\in\binom{\boldsymbol{V}}{2},
    \end{equation}
    $\binom{\boldsymbol{V}}{2}$ is the set of all ordered pairs of elements of $\boldsymbol{V}$, and
    \begin{equation}
        w^0_{ij} = \underset{u\in\boldsymbol{U}}{\sum}w^u_{ij},\ \forall\ (i,\ j)\in\binom{\boldsymbol{V}}{2}.
    \end{equation}  

    A graph structure arises naturally from ranking data as follows. Let $\boldsymbol{G} = (\boldsymbol{V}, \boldsymbol{E})$ be a directed graph whose vertex set is $\boldsymbol{V}$, the set of candidates to be ranked. The edge set is
    \begin{equation}
        \boldsymbol{E} := \left\{e=(i,\ j)\ \Big\vert\ (i,\ j)\in\binom{\boldsymbol{V}}{2}\right\}.
    \end{equation}
    We call such $\boldsymbol{G}$ a pairwise comparison graph. One can further associate weights on the edges as \eqref{eq:edge_weight}. Different from the general pairwise ranking setting, we do not prune the edges whose weights equal to $0$. As a consequence, the pairwise comparison graph $\boldsymbol{G}$ is a complete graph. The cardinality of the edge set is $$|\boldsymbol{E}|:= N = n(n-1).$$ 

    The comparison between $i$ and $j$ will be labeled by different annotators and their answers to the same question could be inconsistent, \textit{i.e.}, 
    \begin{equation}
        y^{u_1}_{ij} = y^{u_2}_{ji} = 1,\ u_1,\ u_2\in\boldsymbol{U}.
    \end{equation}
    To obtain the true direction between vertex $i$ and $j$, we define an estimator $\hat{y}_{ij}$ of noise label ${y}_{ij}$ on edge $e=(i,\ j)$, 
    \begin{equation}
        \hat{y}_{e} = \langle\boldsymbol{z}_{e}, \boldsymbol{\theta}\rangle + \gamma_{e} + \varepsilon_{e},\ \forall\ e\in\boldsymbol{E}, 
    \end{equation}
    where $\boldsymbol{Z}=\{\boldsymbol{z}_{e}\}\in\{-1, 0, 1\}^{N\times n}$, $e\in\boldsymbol{E}$ is the incident matrix of $\boldsymbol{G}$, $\boldsymbol{\theta}\in\mathbb{R}^n$ is some true scaling scores on $\boldsymbol{V}$, $\varepsilon_{e}\sim\mathcal{N}(0,\sigma^2)$ is the Gaussian noise with zero mean and variance $\sigma$, and the outlier indicator variable $\gamma_{e}\in\mathbb{R}$ is assumed to have a higher magnitude than $\sigma$. Here the outliers are the aggregated edges whose directions conflict with the true ranking. In order to estimate the $N + n$ unknown parameters ($N$ for $\boldsymbol{\gamma}$ and $n$ for $\boldsymbol{\theta}$), we aim to minimize the discrepancy between the annotation $\boldsymbol{y}$ and the prediction $\boldsymbol{Z}\boldsymbol{\theta} + \boldsymbol{\gamma}$, as well as holding the outlier indicator $\boldsymbol{\gamma}$ sparse. It gives us the following optimization problem:
    \begin{equation}
        \label{opt:hodgerank}
        \begin{aligned}
            & &\underset{\boldsymbol{\theta},\ \boldsymbol{\gamma}}{\textbf{\textit{minimize}}}&\ \ \ell_{\boldsymbol{w}_0}(\boldsymbol{\theta},\boldsymbol{\gamma}) +\lambda\cdot\mathcal{R}_{\boldsymbol{w}_0}(\boldsymbol{\gamma}),
        \end{aligned}
    \end{equation}
    where
    \begin{equation}
        \begin{aligned}
            & \ell_{\boldsymbol{w}_0}(\boldsymbol{\theta},\boldsymbol{\gamma}) &=& \ \ \frac{1}{2}\ \|\boldsymbol{y} - \boldsymbol{Z}\boldsymbol{\theta} - \boldsymbol{\gamma}\|^2_{2,\boldsymbol{w}_0}\\
            & &=&\ \ \frac{1}{2}\underset{e\in\boldsymbol{E}}{\sum}w^0_{ij}(y_{ij}-\gamma_{ij}-\theta_i+\theta_j)^2,
        \end{aligned}
    \end{equation}
    $\boldsymbol{y} = \text{ver}(\boldsymbol{Y}),\ \boldsymbol{w}^0 = \text{ver}(\boldsymbol{W}^0)$ is the vector form of $\boldsymbol{Y}$ and $\boldsymbol{W}^0$, and the weighted regularization term $\mathcal{R}_{\boldsymbol{w}_0}$ is 
    \begin{equation}
        \label{eq:weight_l1_regular}
        \mathcal{R}_{\boldsymbol{w}_0}(\boldsymbol{\gamma})=\|\boldsymbol{\gamma}\|_{1,\boldsymbol{w}_0}=\underset{e\in\boldsymbol{E}}{\sum}w^0_{ij}|\gamma_e|.
    \end{equation}
    In this situation, the weight $\boldsymbol{w}^0$ and the label $\boldsymbol{y}$ would be treated as the input data of the ranking problem \eqref{opt:hodgerank}. Moreover, we introduce the variable $\boldsymbol{\beta} = (\boldsymbol{\theta},\ \boldsymbol{\gamma})^\top$ to define the action space of the ranking algorithm. We rewrite \eqref{opt:hodgerank} as
    \begin{equation}
        \label{opt:hodgerank_2}
        \begin{aligned}
            \underset{\boldsymbol{\beta}\in\mathcal{B}_{\lambda}}{\textbf{\textit{minimize}}}\ \ \ell_{\boldsymbol{w}_0}(\boldsymbol{\beta}),
        \end{aligned}
    \end{equation}
    where
    \begin{equation}
        \ell_{\boldsymbol{w}_0}(\boldsymbol{\beta}) = \frac{1}{2}\left\|\boldsymbol{y}-\begin{bmatrix}
        \boldsymbol{Z} & \\ 
         & \boldsymbol{1}
        \end{bmatrix}\binom{\boldsymbol{\theta}}{\boldsymbol{\gamma}}\right\|^2_{2,\boldsymbol{w}_0}
    \end{equation}
    and
    \begin{equation}
        \mathcal{B}_{\lambda} = \left\{\boldsymbol{\beta}\ \left|\ \left\langle\left(\boldsymbol{0},\ \boldsymbol{w}_0\right),\ \binom{\boldsymbol{\theta}}{\boldsymbol{\gamma}}\right\rangle\right.\leq\varepsilon(\lambda)\right\}
    \end{equation}
    is the feasible set of \eqref{opt:hodgerank} and the ranker's action space.

    We model poisoning attack as a game between two players, the ranker and an attacker, where the latter wants to mislead its opponent into picking parameters to generate a difference order against the true ranking. To disguise himself, the adversary needs to coordinate a poisoned $\boldsymbol{w}$ associate with $\boldsymbol{y}$. Intuitively, the adversary could not obtain $\boldsymbol{w}$ through drastic changes, neither on each $w_{ij}$ nor $\sum w_{ij}$. Such limitations lead to the following constraints for adversary's action. First, the total difference between $\boldsymbol{w}_0$ and $\boldsymbol{w}$ would be smaller than $b$, namely,
    \begin{equation}
        \|\boldsymbol{w}-\boldsymbol{w}_0\|_1\leq b,\ \ b\in\mathbb{Z}_+.
    \end{equation}
    Furthermore, the adversary could not alter the number of votes on any  pairwise comparison $e\in\boldsymbol{E}$ obviously,
    \begin{equation}
        \|\boldsymbol{w}-\boldsymbol{w}_0\|_{\infty}\leq l, \ \ l\in\mathbb{Z}_+,\ \ l\leq\textbf{\textit{min}}\{\textbf{\textit{max}}(\boldsymbol{w}_0), b\},
    \end{equation}
    and the adversary‘s action space $\mathcal{W}_{\boldsymbol{w}_0}$ is
    \begin{equation}
        \begin{aligned}
            \mathcal{W}_{\boldsymbol{w}_0}=\ \ \left\{\boldsymbol{w}\ \left|\  
        \begin{matrix}
            \boldsymbol{w}\in\mathbb{Z}^{N}_+,\ l,\ b\in\mathbb{Z}_+,\\
            \ \|\boldsymbol{w}-\boldsymbol{w}_0\|_1\ \leq b,\\
            \ \|\boldsymbol{w}-\boldsymbol{w}_0\|_{\infty}\leq l,\\
            l\leq\textbf{\textit{min}}\{\textbf{\textit{max}}(\boldsymbol{w}_0), b\}\ 
        \end{matrix}\right.\right\}.
        \end{aligned}
    \end{equation}
    The robust ranking algorithm \eqref{opt:hodgerank} observes the poisoned training set sampling from $\boldsymbol{G}$, prunes the outlier and learns a ranking from the remaining data simultaneously. Against the robust ranking algorithm that employs the defense described above, we can formulate the attacker’s goal as the following bi-level optimization problem:
    \begin{align*}
        & &\underset{\boldsymbol{w}}{\textbf{\textit{maximize}}}&\ \ \ell_{\boldsymbol{w}}(\hat{\boldsymbol{\theta}},\hat{\boldsymbol{\gamma}}) +\lambda\cdot\mathcal{R}_{\boldsymbol{w}}(\hat{\boldsymbol{\gamma}}), \numberthis \label{opt:integer_opt}\\
        & &\textbf{\textit{subject to}}&\ \ \hat{\boldsymbol{\theta}},\ \hat{\boldsymbol{\gamma}}\in\underset{\boldsymbol{\theta},\boldsymbol{\gamma}}{\textbf{\textit{arg min}}}\ \ \ell_{\boldsymbol{w}}(\boldsymbol{\theta},\ \boldsymbol{\gamma}) +\lambda\cdot\mathcal{R}_{\boldsymbol{w}}(\boldsymbol{\gamma}),\\
        & & &\ \ \boldsymbol{w}\in\mathcal{W}_{\boldsymbol{w}_0}.
    \end{align*}    

    \subsection*{Distributional Perspective}
    In \eqref{eq:adversary_action_space_1}, the $\ell_1$ and $\ell_{\infty}$ distance constraints on $\boldsymbol{w}$ correspond to the attacker only being able to find the perturbation in the neighborhood of $\boldsymbol{w}^0$. The lower problem in \eqref{opt:integer_opt} corresponds to the robust pairwise ranking algorithm. With input data $\{\boldsymbol{w}, \boldsymbol{y}\}$, the ranker obtain the relative score $\boldsymbol{\theta}$ by minimizing the discrepancy between the annotation $\boldsymbol{y}$ and the prediction $\boldsymbol{Z}\boldsymbol{\theta} + \boldsymbol{\gamma}$ while keeping $\boldsymbol{\gamma}$ to be sparse. Unfortunately, the bilevel nature of \eqref{opt:integer_opt} \cite{Bard1991,bard2013practical}—maximizing the outer loss involves an inner minimization to find the parameters $\boldsymbol{\theta}, \boldsymbol{\gamma}$—makes it difficult to solve, even less the discrete property of \eqref{eq:adversary_constraint_1} and \eqref{eq:adversary_constraint_2}.
    Next, we will discuss the poisoning attack on pairwise ranking in a different way. Generally, we can look at the poison attack \eqref{opt:integer_opt} from a distributional perspective. The attacker and the ranker both access the weighted comparison graph $\boldsymbol{G}$ to play a game as \eqref{opt:integer_opt}. Actually, the non-toxic training data $\{\boldsymbol{w}_0,\ \boldsymbol{y}\}$ are drawn according to a probability distribution $P$
    \begin{equation}
        p(\boldsymbol{w}_0,\ \boldsymbol{y})=\underset{e\in\boldsymbol{E}}{\sum}\ p(w^0_{ij},\ y_{ij}). 
    \end{equation}
    The attacker chooses a perturbation function $\psi:\mathbb{Z}^{N}_+\rightarrow\mathbb{Z}^{N}_+$ that change the weight $\boldsymbol{w}_0$ to $\boldsymbol{w}$. The attacker constructs the perturbation $\psi$ with the limitation as \eqref{eq:adversary_action_space_1}. Such a perturbation $\psi$ induces a transition from empirical distribution $P$ to a poisoned distribution $Q$
    \begin{equation}
        \label{eq:poison_distribution}
        q(\boldsymbol{w}, \boldsymbol{y}) = q(\psi(\boldsymbol{w}_0), \boldsymbol{y})=\underset{e\in\boldsymbol{E}}{\sum}q(w_{ij}, y_{ij}).
    \end{equation}
    The attacker can only alter $b$ pairwise comparisons at most, increase or decrease the number of vote on any comparison less than $l$, and formulate the poisoned training set $\{\boldsymbol{w},\ \boldsymbol{y}\}$.
    If the attacker selects $Q$ in a small enough \textit{neighborhood} of $P$, namely, the ``distance'' between the poisoned distribution $Q$ and the empirical distribution $P$ would be small, the attacker could obtain a good approximation of $P$ in the sense of such a ``distance'' and the poisoned sample $\{\boldsymbol{w},\ \boldsymbol{y}\}$ would be satisfied the constraints \eqref{eq:adversary_constraint_1} and \eqref{eq:adversary_constraint_2}.   

    Let $\phi: \mathbb{R}_+\rightarrow\mathbb{R}$ be a convex function with $\phi(1) = 0$. Then the $\phi$-divergence between distributions $Q$ and $P$ defined on a space $\mathcal{X}$ is
    \begin{equation}
        \begin{aligned}
            & d_{\phi}(Q||P)&=&\ \ \ \int\phi\left(\frac{dQ}{dP}\right)dP\\
            & &=&\ \ \int_{\mathcal{X}}\phi\left(\frac{q(x)}{p(x)}\right)p(x)d\mu(x),
        \end{aligned}
    \end{equation}
    where $\mu$ is a $\sigma$-finite measure with $Q, P \ll \mu$, and $q=\frac{dQ}{d\mu}$, $p=\frac{dP}{d\mu}$. Given $\phi$ and sample $\boldsymbol{w}_0$, we reformulate
    the adversary's action space, the local neighborhood of the empirical distribution $P$ with radius $\rho$ as
    \begin{equation}
        \mathcal{Q}_{P}=\{\text{distribution } Q \text{ satisfies }d_{\phi}(Q||P)\leq \rho\},
    \end{equation}
    where $P$ is the empirical distribution of the pairwise comparisons, and $Q$ is the toxic distribution for poisoning attack. Throughout this paper, we adopt
    $$
    \phi(t) = \frac{1}{2}(t-1)^2,
    $$
    which gives the $\chi^2$-divergence \cite{Tsybakov:2008:INE:1522486,DBLP:conf/nips/NamkoongD17,DBLP:journals/jmlr/DuchiN19}. It means that $Q$ consists of discrete distributions supported on the observation $\{(\boldsymbol{w}^0,\ \boldsymbol{y})\}$. With opportunely chosen $\rho$, the adversary could obtain $\boldsymbol{w}$ which satisfies the neighborhood constraints as \eqref{eq:adversary_constraint_1} and \eqref{eq:adversary_constraint_2}.  

    The possible actions of two players $\boldsymbol{a} = [\boldsymbol{\beta}_{\lambda},\ \boldsymbol{w}]$ constitute the joint action space $\mathcal{A}=\mathcal{B}_{\lambda}\times\mathcal{Q}_{P}$ which is assumed to be nonempty, compact, and convex. Action spaces $\mathcal{A}$ are parameters of the game \eqref{opt:integer_opt}. Then the bi-level integer programming \eqref{opt:integer_opt} can be written as a min-max optimization problem:
    \begin{equation}
        \underset{q\in\mathcal{Q}_{P}}{\boldsymbol{\sup}}\ \underset{\boldsymbol{\beta}\in\mathcal{B}_{\lambda}}{\vphantom{Q\in\mathcal{Q}_{P}}\boldsymbol{\inf}}\ \mathbb{E}_{Q}[\ell(\boldsymbol{\beta},\ q(\boldsymbol{w},\ \boldsymbol{y}))]\label{opt:minmax_opt_1}=\underset{d_{\phi}(Q||P)\leq \rho}{\boldsymbol{\sup}}\ \ \underset{\|q(\boldsymbol{w})\circ\boldsymbol{\beta}\|_1\leq\varepsilon(\lambda)\vphantom{Q\in\mathcal{Q}_{P}}}{\boldsymbol{\inf}}\ \ \mathbb{E}_{Q}[\ell(\boldsymbol{\beta},\ q(\boldsymbol{w},\ \boldsymbol{y}))]\nonumber
    \end{equation}
    where 
    \begin{equation}
        q(\boldsymbol{w})\circ\boldsymbol{\beta}=
        \begin{bmatrix}
        \boldsymbol{0} & \\
         & q(\boldsymbol{W})
        \end{bmatrix}
        \begin{bmatrix}
        \boldsymbol{\theta}\\
        \boldsymbol{\gamma}
        \end{bmatrix},
    \end{equation}
    $\boldsymbol{W} = \textbf{\textit{diag}}(\boldsymbol{w})$ is a diagonal matrix. Due to the $\ell_1$ norm is decomposable, we can define a new set of loss function $f_{ij}:\mathcal{B}_{\lambda}\times\mathbb{Z}_+\times\{-1,\ 1\}^{N}\rightarrow\mathbb{R}_+,\ \ \forall\ e\in\boldsymbol{E}$
    \begin{align}
        & & &\ \ f_{ij}(\boldsymbol{\beta},\ q(\boldsymbol{w},\ \boldsymbol{y}))\\
        & &=&\ \ q(w_{ij})\cdot\frac{1}{2}(y_{ij}-\gamma_{ij}-\theta_i+\theta_j)^2+\lambda\cdot q(w_{ij})|\gamma_{ij}|\nonumber\\
        & &=&\ \ q(w_{ij})\cdot\left[\frac{1}{2}(y_{ij}-\gamma_{ij}-\theta_i+\theta_j)^2+\lambda\cdot|\gamma_{ij}|\right]\nonumber
    \end{align}
    and the finite sum of $\{f_{ij}\}$
    \begin{equation}
        f(\boldsymbol{\beta},\ q(\boldsymbol{w},\ \boldsymbol{y})) = \underset{e\in\boldsymbol{E}}{\sum}\ f_{ij}(\boldsymbol{\beta},\ q(\boldsymbol{w},\ \boldsymbol{y}))
    \end{equation}
    With fixed $\lambda$ and some special form of $q$, \eqref{opt:minmax_opt_1} could be a convex problem. We swap the order of minimization and maximization in the min-max optimization problem \eqref{opt:minmax_opt_1} as 
    \begin{equation}
        \label{opt:minmax_opt_2}
        \underset{\boldsymbol{\beta}\in\mathcal{B}_{\lambda}}{\vphantom{Q\in\mathcal{Q}_{P}}\boldsymbol{\inf}}\ \ \underset{q\in\mathcal{Q}_{P}}{\boldsymbol{\sup}}\ \ \ \ \mathbb{E}_{Q}[f(\boldsymbol{\beta},\ q(\boldsymbol{w},\ \boldsymbol{y}))]=\underset{\boldsymbol{\beta}\in\mathcal{B}_{\lambda}}{\vphantom{Q}\boldsymbol{\inf}}\ \ \ \underset{q}{\boldsymbol{\sup}}\ \left\{\mathbb{E}_{Q}[f(\boldsymbol{\beta},\ q(\boldsymbol{w},\ \boldsymbol{y}))],\ \textbf{\textit{s.t.}}\ d_{\phi}(Q||P)\leq \rho\vphantom{\frac{1}{2}}\right\}.
        \end{equation}
    In fact, the minimization of \eqref{opt:minmax_opt_1} and \eqref{opt:minmax_opt_2} correspond to the residual method and the Tikhonov regularization with discrepancy principle of the LASSO. Indeed, it can be shown that the constrained minimization problem is equivalent to Tikhonov regularization, when the regularization parameter $\lambda$ is chosen according to Morozov’s discrepancy principle \cite{doi:10.1002/cpa.20350}. {Note that the objective function of \eqref{opt:minmax_opt_2} is a strictly convex function with respect to its arguments, then by \cite[Theorem 4.3]{Basar-Olsder1999}, at least one Nash equilibrium exists.}  

    \subsection*{Optimization}
    From a game-theoretic viewpoint, \eqref{opt:minmax_opt_2} can be seen as a zero-sum game between two agents: the agent ranker (the infimum) seeks to incur the least possible loss, while the agent adversary (the supremum) seeks to obtain the worst possible objective function value – both given by $f(\boldsymbol{\beta}_{\lambda}, \boldsymbol{w}, \boldsymbol{y})$.

    For the supremum part of \eqref{opt:minmax_opt_2}, the integer characteristic of $\boldsymbol{w}$ obstructs obtaining a probability density function $q$ of the toxic distribution $Q$. Thanks to distributionally robust optimization, we reformulate the supremum part of \eqref{opt:minmax_opt_2} as a quadratically constrained linear maximization problem. This tractable formulation can be solved by the probability simplex projection method. Suppose the total number of pairwise comparison without toxic is 
    \begin{equation}
        M^0 = \underset{e\in\boldsymbol{E}}{\sum}\ w^0_{ij},
    \end{equation}
    and the frequencies of each comparison are 
    \begin{equation}
        \boldsymbol{p} = \frac{1}{M^0}\ \boldsymbol{w}^0.
    \end{equation}
    Let the maximum toxic dosage be $\kappa$. It suggests that the number of toxic pairwise comparisons $M$ satisfies 
    \begin{equation}
        M = \underset{e\in\boldsymbol{E}}{\sum}\ w_{ij} \leq(1+\kappa)\cdot M^0,
    \end{equation}
    Furthermore, we replace the toxic weight $\boldsymbol{w}$ with its frequency $\boldsymbol{q} = \frac{\boldsymbol{w}}{M}$. We relax the integer programming problem into a general optimization by such a variable substitution. We note
    \begin{equation}
        \label{eq:objective_value}
        z_{e} = \frac{1}{2}(y_{ij}-\gamma^{\lambda}_{ij}-\theta^{\lambda}_i+\theta^{\lambda}_j)^2+\lambda\cdot|\gamma^{\lambda}_{ij}|,\ e\in\boldsymbol{E}
    \end{equation}
    and $\boldsymbol{z}=[z_1,\dots,z_N]\in\mathbb{R}^N_+$. The objective function with fixed $\boldsymbol{\beta}_\lambda$, maximizing the expectation $\mathbb{E}_{Q}[f(\boldsymbol{\beta}_{\lambda},\ q(\boldsymbol{w},\ \boldsymbol{y}))]$ equals to compute the worst-case linear combination of $\{z_{ij}\},\ e\in\boldsymbol{E}$ as
    \begin{equation}
        \underset{\boldsymbol{q}}{\textbf{\textit{maximize}}}\ \ \left\langle\boldsymbol{q},\ \boldsymbol{z}\right\rangle,\ \textbf{s.t.}\ d_{\phi}(Q||P)\leq \rho.
    \end{equation}
    As $Q$ is a distribution, it requires that the combination coefficients $\boldsymbol{q}$ should satisfy $$\sum_{e\in\boldsymbol{E}} q_e = 1\ \ \text{or}\ \ \langle\boldsymbol{1},\boldsymbol{q}\rangle=1.$$
    It means that the distribution of $\boldsymbol{q}$ is a probability simplex. Furthermore, as $P$ and $Q$ are the discrete distributions and we choose $\phi(t) = \frac{1}{2}(t-1)^2$ in \eqref{eq:phi_divergence}, the neighborhood constraint $d_{\phi}(Q||P)\leq \rho$
    can be transformed as
    \begin{equation}
        \label{eq:quad_constraint}
        \frac{1}{2}\|\boldsymbol{q}-\boldsymbol{p}\|^2_2\leq\rho\|\boldsymbol{p}\|^2_2.
    \end{equation}
    Now we obtain the following quadratically constrained linear maximization problem which could be used to compute the supremum problem in \eqref{opt:minmax_opt_2}:
    \begin{equation}
        \label{opt:quad_linear_uniform_1}
        \begin{aligned}
            \underset{\boldsymbol{q}}{\textbf{\textit{maximize}}}&\ \ \left\langle\boldsymbol{q},\boldsymbol{z}\right\rangle\ \ \textbf{\textit{s.t.}}\ \ \boldsymbol{q}\in\mathcal{Q}_{\boldsymbol{p}}
        \end{aligned}
    \end{equation}
    where
    \begin{equation}
        \label{eq:q_constraint_uniform}
        \mathcal{Q}_{\boldsymbol{p}}=\left\{\boldsymbol{q}\ \left|\ \frac{1}{2}\|\boldsymbol{q}-\boldsymbol{p}\|^2_2\leq\rho\|\boldsymbol{p}\|^2_2,\ \left\langle\boldsymbol{1},\ \boldsymbol{q}\right\rangle=1\right.\right\}.
    \end{equation}
    We reformulate the concave optimization \eqref{opt:quad_linear_uniform_1} as a minimization problem for simplicity:
    \begin{equation}
        \label{opt:quad_linear_uniform_2}
        \begin{aligned}
            \underset{\boldsymbol{q}}{\textbf{\textit{minimize}}}&\ \ \left\langle\boldsymbol{q},\boldsymbol{z}\right\rangle\ \ \textbf{\textit{s.t.}}\ \ \boldsymbol{q}\in\mathcal{Q}_{\boldsymbol{p}},
        \end{aligned}
    \end{equation}
    and take a partial dual problem of this minimization, then maximize this dual problem to find the optimal $\boldsymbol{q}$. 

    First, we introduce the dual variable $\mu\geq 0$ for the quadratical constraint \eqref{eq:quad_constraint}. Notice that the strong duality exists for \eqref{opt:quad_linear_uniform_2} because the Slater condition is satisfied by
    $$
    \boldsymbol{q} = \boldsymbol{p}\ \ \text{ and }\ \ \boldsymbol{1}^\top\boldsymbol{p}=1.
    $$
    Performing the standard min-max swap \cite{boyd2004convex}, it yields the following problem
    \begin{gather}
        \underset{\mu\geq 0}{\textbf{\textit{minimize}}}\ g(\mu) = \label{opt:dual_max}\\
        \underset{\boldsymbol{q}}{\boldsymbol{\inf}}\left\{\frac{\mu}{2}\|\boldsymbol{q}-\boldsymbol{p}\|^2_2-\mu\rho\|\boldsymbol{p}\|^2_2+\boldsymbol{q}^\top\boldsymbol{z}\ |\ \boldsymbol{q}\in\mathbb{R}^N_+, \boldsymbol{q}^\top\boldsymbol{1}=1\right\}\nonumber.
    \end{gather}    

    \begin{algorithm}[tbh!]
        \SetAlgoLined
        \SetKwInOut{Input}{Input}
        \SetKwInOut{Output}{Output}
        \Input{the original data $\{\boldsymbol{w}_0,\boldsymbol{y}\}$, maximum toxic dosage $\kappa$, parameter $\rho$, solution accuracy $\varepsilon$, and an outlier pruning rate $\tau$.}
        {Initialize} the frequency of $\boldsymbol{w}_0$, $\boldsymbol{p}$ by \eqref{eq:fre_p}, 

        Obtain the ranking parameters on the original data: 
        $$\boldsymbol{\beta}_{\lambda}\leftarrow\textbf{\textit{HodgeRank}}(\boldsymbol{w}_0,\boldsymbol{y}, \tau),$$   

        Calculate the objective function value $\boldsymbol{z}$ by \eqref{eq:objective_value}:

        \While{$\boldsymbol{w}$ not converged}
        {
            Update the frequency: $\boldsymbol{q} = \textbf{\textit{WorstCase}}(\boldsymbol{z}, \boldsymbol{p}, \rho, \varepsilon)$,

            Assign the weight with $\boldsymbol{q}$,
            \begin{equation}
                {\boldsymbol{w}}' = \left[(1+\kappa)\cdot M^0\right]\boldsymbol{q},
            \end{equation}  

            Round ${\boldsymbol{w}}'$ to obtain the $\boldsymbol{w}$ as integer vector
            $$
                \boldsymbol{w}=\textbf{\textit{rounding}}({\boldsymbol{w}}'),
            $$  

            Update the ranking parameters: 
            $$\boldsymbol{\beta}_{\lambda}\leftarrow\textbf{\textit{HodgeRank}}(\boldsymbol{w},\boldsymbol{y}, \tau),
            $$  

            Update the objective function value: $\boldsymbol{z}$ via \eqref{eq:objective_value},
        }   

        \Output{the poisoned data $\{\boldsymbol{w},\boldsymbol{y}\}$, the ranking parameters $\boldsymbol{\beta}_{\lambda}=\{\boldsymbol{\theta}_{\lambda}, \boldsymbol{\gamma}_{\lambda}\}$.}
        \caption{Poisoning Attack on Pairwise Ranking}
        \label{alg:main_2}
    \end{algorithm} 

    Given a collection of concave functions $\{g_q\}_{q\in\mathcal{Q}_{\boldsymbol{p}}}$, if it attains 
    $$
    \inf g\ =\ \underset{q\in\mathcal{Q}_{\boldsymbol{p}}}{\inf}\ \ g_q
    $$
    at some $q_0\in\mathcal{Q}_{\boldsymbol{p}}$, we know that $\nabla g_{q_0}$ is the super-gradient of $g$ \cite[Chapter VI.4.4]{hiriart2013convex}. Suppose $\boldsymbol{q}(\mu)$ is the unique minimizer of the right hand side of \eqref{opt:dual_max}, the dual function $g$ will be
    \begin{equation}
         g(\mu) = \frac{\mu}{2}\|\boldsymbol{q}(\mu)-\boldsymbol{p}\|^2_2-\mu\rho\|\boldsymbol{p}\|^2_2+\boldsymbol{q}(\mu)^\top\boldsymbol{z}
    \end{equation}
    and the derivative with respect to $\mu$ (keeping $\boldsymbol{q}(\mu)$ fixed) is
    \begin{equation}
        g'(\mu) = \frac{1}{2}\|\boldsymbol{q}(\mu)-\boldsymbol{p}\|^2_2-\rho\|\boldsymbol{p}\|^2_2.
    \end{equation}
    As the constraints $\boldsymbol{q}\geq 0$ and $\boldsymbol{q}^\top\boldsymbol{1}=1$ require $\boldsymbol{q}$ is on the probability simplex, we adopt the Euclidean projection of a vector to the probability simplex \cite{DBLP:conf/icml/DuchiSSC08}. Such a projection provides an efficient solver of the infimum \eqref{opt:quad_linear_uniform_2}. With no loss of generality, we assume that $\boldsymbol{z}$ is an increasing sequence and the mean of $\boldsymbol{z}$ is zero,
    \begin{equation}
        z_1\leq z_2\leq \dots \leq z_N,\ \ \ \langle\boldsymbol{z}, \boldsymbol{1}\rangle = 0.
    \end{equation}
    Then we use $\boldsymbol{a},\boldsymbol{\sigma}\in\mathbb{R}^N_+$, the cumulative summation of $\boldsymbol{z}$ and  $\boldsymbol{z}^2$ as
    \begin{equation}
        a_i = \underset{j\leq i}{\sum}z_i,\ \ \ \sigma_i = \underset{j\leq i}{\sum}z_i^2,\ \ \ i\in[N].
    \end{equation}
    The infimum in \eqref{opt:dual_max} is equivalent to projecting the vector $\boldsymbol{v}(\mu)\in\mathbb{R}^N$ onto the probability simplex,
    \begin{equation}
        v_i = p_i - \frac{1}{\mu}z_i,\ \ \ i\in[N]
    \end{equation}
    According to \cite{DBLP:conf/icml/DuchiSSC08}, $\boldsymbol{q}(\mu)$ has the form as $q_i(\mu)=(v_i - \eta)_+$ for some $\eta\in\mathbb{R}$, where $\eta$ is selected such that $\sum q_i(\mu) = 1$. Finding such a value $\eta$ is equivalent to finding the unique index $i$ such that
    \begin{equation}
        \label{eq:eta_constraint}
        \sum^i_{j=1}(v_j-v_i)<1\ \ \text{and}\ \ \sum^{i+1}_{j=1}(v_j-v_{i+1})>1.
    \end{equation}
    If no such index exists, we set $i=n$ as the sum  $\sum^i_{j=1}(v_j-v_i)$ is increasing in $i$ and $v_1-v_1=0$. Given the index $i$,
    \begin{equation}
        \label{eq:eta}
        \eta = p_i-\frac{1}{i}-\frac{1}{i\mu}\sum^{i}_{j=1}z_i=p_i-\frac{1}{i}-\frac{1}{i\mu}a_i
    \end{equation}
    satisfies $\sum(v_i-\eta)_+ = 1$ and $v_j-\eta\geq 0$ for any $j\leq i$ while $v_j-\eta\leq 0$ for $j>i$. Meanwhile, the derivative $\frac{\partial}{\partial \mu}g(\mu)$ (where $\boldsymbol{q}(\mu)$ is fixed) has a explicit form
    \begin{equation}
        \label{eq:dual_derivative}
        \begin{aligned}
            & g'(\mu) &=&\ \ \frac{\partial}{\partial \mu}\left\{\frac{\mu}{2}\|\boldsymbol{q}(\mu)-\boldsymbol{p}\|^2-\mu\rho\|\boldsymbol{p}\|^2_2+\boldsymbol{q}^\top(\mu)\boldsymbol{z}\right\}\\
            & &=&\ \ \frac{1}{2}\|\boldsymbol{q}(\mu)-\boldsymbol{p}\|^2-\rho\|\boldsymbol{p}\|^2\\
            & &=&\ \ \frac{1}{2}\sum^{i}_{j=1}(v_j-\eta-p_j)^2+\frac{1}{2}\sum^{N}_{j=i+1}p^2_j-\rho\|\boldsymbol{p}\|^2\\
            & &=&\ \ \frac{1}{2}\sum^{i}_{j=1}\left(\frac{z_j}{\mu}+\eta\right)^2+\frac{1}{2}\sum^{N}_{j=i+1}p^2_j-\rho\|\boldsymbol{p}\|^2\\
            & &=&\ \ \frac{\sigma_i}{2\mu^2}+\frac{\eta^2i}{2}+\frac{a_i\eta}{\mu}+\sum_{j=i+1}^{N}p^2_j-\rho\|\boldsymbol{p}\|^2
        \end{aligned}
    \end{equation}
    The derivative $g'(\mu)$ only needs $\mathcal{O}(1)$ when $\boldsymbol{a}$ and $\boldsymbol{\sigma}$ are known. Binary search can calculate the optimal index $i$ and $\boldsymbol{q}$ efficiently, which requires $\mathcal{O}(\log\frac{1}{\varepsilon}\log N)$ to find $\mu$ with accuracy $\varepsilon$. We can get $\eta$  through \eqref{eq:eta} if \eqref{eq:eta_constraint} are satisfied. The solution $\boldsymbol{q}(\mu)$ is
    \begin{equation}
        \label{eq:final_solution}
        q_i = \left(p_i - \frac{z_i}{\mu}-\eta\right)_+, i\in[N].
    \end{equation}
    Specifically, the computational complexity to obtain the sorted vector $\boldsymbol{z}$ is ${\cal O}(N\log N)$, and that of the estimate of the frequency $\boldsymbol{q}$ is $\mathcal{O}(\log \frac{1}{\varepsilon}\log N)$. The overall time computational complexity is $\mathcal{O}(N\log N + \log \frac{1}{\varepsilon}\log N)$.  

    At last, we describe the whole optimization of the poison attack on pairwise ranking with \textbf{Algorithm \ref{alg:main}}. We summarize the complete optimization procedure of the supremum in \eqref{opt:minmax_opt_2} as \textbf{Algorithm \ref{alg:worstcase}} and \textbf{Algorithm 3}. For the infimum part of \eqref{opt:minmax_opt_2}, the agent HodgeRank finds $\boldsymbol{\beta}_{\lambda}$ that minimizes the regularized loss on $\{\boldsymbol{w},\ \boldsymbol{y}\}$ where the hyper-parameter $\lambda$ controls the regularization strength. We include the solving process of HodgeRank as \textbf{Algorithm\ \ref{alg:hodgerank}} for completeness. 

        \begin{algorithm}[h!]
        \SetAlgoLined
        \SetKwInOut{Input}{Input}
        \SetKwInOut{Output}{Output}
        \Input{the objective function value $\boldsymbol{z}\in\mathbb{R}^N$,\ the frequency of true comparisons $\boldsymbol{p}\in\mathbb{R}^N_+$,\ parameter $\rho$,\ solution accuracy $\varepsilon$.}
        Make $\boldsymbol{z}$ have the zero mean: $\boldsymbol{z} \leftarrow \boldsymbol{z} - \bar{\boldsymbol{z}}$, and sort $\boldsymbol{z}$. 

        Initialize $\mu_{\min} = 0$, $a_i = \sum_{j\leq i}z_{j}$, and $\sigma_i = \sum_{j\leq i}z^2_{j}$ for all $i\in[N]$,
        $$
        \mu_{\max} = \mu_{\infty} = \max\left\{\|\boldsymbol{z}\|_{\infty},\ \sqrt{\frac{1}{\rho\|\boldsymbol{p}\|^2_2}}\|\boldsymbol{z}\|_2\right\}
        $$  

        \While{$|\mu_{\max}-\mu_{\min}|>\varepsilon\mu_{\infty}$}
        {
            Set $\mu = \frac{1}{2}(\mu_{\min}+\mu_{\max})$, and $$(\eta,\ i) = \textbf{\textit{FindShift}}(\boldsymbol{z},\ \boldsymbol{p},\ \boldsymbol{a},\ \mu),$$   

            Obtain the partial derivative ${g}'(\mu)$ by \eqref{eq:dual_derivative}

            \uIf{${g}'(\mu)>0$}
            {
                $\mu_{\min} \leftarrow \mu$
            }
            \Else
            {
                $\mu_{\max} \leftarrow \mu$
            }
        }   

        {Set} $\mu = \frac{1}{2}(\mu_{\min}+\mu_{\max})$, and
        $$
        (\eta,\ i) = \textbf{\textit{FindShift}}(\boldsymbol{z},\ \boldsymbol{p},\ \boldsymbol{a},\ \mu),
        $$

        \Output{$\boldsymbol{q}$ by \eqref{eq:final_solution}.}
        \caption{$\textbf{\textit{WorstCase}}(\boldsymbol{z},\ \boldsymbol{p},\ \rho,\ \varepsilon)$}
        \label{alg:worstcase}
    \end{algorithm} 

    \begin{algorithm}[bht!]
        \caption{$\textbf{\textit{FindShift}}(\boldsymbol{z}, \boldsymbol{p}, \boldsymbol{a}, \mu)$}
        \SetAlgoLined
        \SetKwInOut{Input}{Input}
        \SetKwInOut{Output}{Output}
        \Input{the sorted and zero mean vector $\boldsymbol{z}\in\mathbb{R}^N$, the frequency $\boldsymbol{p}$, the cumulative sum $\boldsymbol{a}$, and the dual variable $\mu$.}
        \textbf{Initialize} $i_{\text{low}} = 1$ and $i_{\text{high}} = N$, 

        \uIf{$p_N - \frac{1}{\mu}z_N\geq 0$}
        {
            $\eta = 0$,\ $i=N$, 

            \textbf{Break.}
        }
        \Else
        {
            \While{$i_{\text{low}}\neq i_{\text{high}}$}
            {
                $i = \frac{1}{2}(i_{\text{low}}+i_{\text{high}})$,  

                $a_{\text{left}} = \sum^i_{j=1}(v_j-v_i) = \frac{1}{\mu}(iz_i-a_i)$,    

                $a_{\text{right}} = \sum^{i+1}_{j=1}(v_j-v_{i+1}) = \frac{1}{\mu}[(i+1)z_{i+1}-a_{i+1}]$,   

                \uIf{$a_{\text{right}} \geq 1\wedge a_{\text{left}} < 1$}
                {
                    $\eta = p_i-\frac{1}{i}-\frac{1}{i\mu}a_i$, 

                    \textbf{Break.}
                }
                \uElseIf{$a_{\text{left}}\geq 1$}
                {
                    $i_{\text{high}} = i-1$
                }
                \Else
                {
                    $i_{\text{low}} = i+1$
                }
            }
        }
        \Output{$i=i_{\text{low}}$, $\eta = p_i-\frac{1}{i}-\frac{1}{i\mu}a_i$.}
        \label{alg:findshift}
    \end{algorithm} 

    \begin{algorithm}[h!]
        \SetAlgoLined
        \SetKwInOut{Input}{Input}
        \SetKwInOut{Output}{Output}
        \Input{the weight $\boldsymbol{w}$, the comparison matrix $\boldsymbol{A}$ and the corresponding label $\boldsymbol{y}$.}

        Calculate the relative ranking score $\hat{\boldsymbol{\theta}}$
        \begin{equation*}
            \hat{\boldsymbol{\theta}} = (\boldsymbol{X}^{\top}\boldsymbol{X}+\delta\boldsymbol{I})^{-1}\boldsymbol{X}^{\top}\sqrt{\boldsymbol{W}}\boldsymbol{y}
        \end{equation*}
        {\scriptsize where $\boldsymbol{X} = \sqrt{\boldsymbol{W}}\boldsymbol{A}$, $\sqrt{\boldsymbol{W}}= \textbf{\textit{diag}}(\sqrt{\boldsymbol{w}})$.}    

        \Output{the corresponding ranking parameter $\hat{\boldsymbol{\theta}}$.}
        \caption{$\textbf{\textit{HodgeRank}}(\boldsymbol{w}, \boldsymbol{A}, \boldsymbol{y})$}
        \label{alg:hodgerank}
    \end{algorithm}
}

\end{document}